\documentclass[journal]{IEEEtran}

\usepackage{cite}

\ifCLASSINFOpdf
  \usepackage[pdftex]{graphicx}
  \graphicspath{{pdfs/}{images/}}
  \DeclareGraphicsExtensions{.pdf,.jpeg,.png}
\else
\fi

\usepackage{amsmath}
\usepackage{amsfonts}
\usepackage{url}

\usepackage{fp}
\usepackage{overpic}
\usepackage{import}
\usepackage{caption}
\usepackage{mathtools}
\usepackage{caption}
\usepackage{subcaption}
\usepackage{orcidlink}
\usepackage{multirow}
\usepackage{colortbl}
\usepackage{booktabs}
\usepackage{makecell}
\usepackage{cleveref}
\usetikzlibrary{spy,
                arrows.meta,
                calc, chains,
                quotes,
                positioning,
                shapes.geometric}

\usepackage{pgfplots}
\usepgfplotslibrary{groupplots}
\usepackage{pgfplotstable}
\pgfplotsset{compat=1.17}

\definecolor{lightred}{HTML}{FF9999}
\definecolor{lightyellow}{HTML}{FFFF99}
\definecolor{lightorange}{HTML}{FFCC99}

\DeclareMathOperator*{\argmin}{arg\,min}

\newcommand{\sh}{spherical harmonic}
\newcommand{\shs}{spherical harmonics}

\newcommand{\vect}[1]{\boldsymbol{\mathbf{#1}}}
\newcommand{\matr}[1]{\boldsymbol{\mathbf{#1}}}

\usepackage{xcolor}
\newcommand{\added}[1]{\textcolor{blue}{#1}}
\newcommand{\deleted}[1]{\textcolor{red}{\sout{#1}}}
\newcommand{\changed}[2]{\deleted{#1} \added{#2}}
\usepackage[normalem]{ulem} 

\renewcommand{\added}[1]{#1}
\renewcommand{\deleted}[1]{}
\renewcommand{\changed}[2]{\added{#2}}

\begin{document}

\title{POTR: Post-Training 3DGS Compression}

\author{
Bert~Ramlot~\orcidlink{0009-0006-7787-4218},
Martijn~Courteaux~\orcidlink{0000-0002-9971-3128},
Peter~Lambert~\orcidlink{0000-0001-5313-4158}, \IEEEmembership{Senior Member, IEEE}, \\
Glenn~Van~Wallendael~\orcidlink{0000-0001-9530-3466}, \IEEEmembership{Member, IEEE}

\thanks{This work was funded in part by the Research Foundation—Flanders (FWO) under Grant \added{1SA0B26N, the imec.prospect project SitSens}, IDLab (Ghent University—imec), Flanders Innovation and Entrepreneurship (VLAIO), and the European Union.}%
\thanks{
The authors are with the IDLab-MEDIA research group, part of Ghent University and imec, located at AA Tower, Technologiepark-Zwijnaarde 122, B-9052 Zwijnaarde, Belgium. Corresponding author: Bert Ramlot. (e-mail: bert.ramlot@ugent.be; martijn.courteaux@ugent.be; peter.lambert@ugent.be; glenn.vanwallendael@ugent.be).}
\thanks{This work has been submitted to the IEEE for possible publication. Copyright may be transferred without notice, after which this version may no longer be accessible.}
}

%


\maketitle
\begin{abstract}
3D Gaussian Splatting (3DGS) has recently emerged as a promising contender to Neural Radiance Fields (NeRF) in 3D scene reconstruction and real-time novel view synthesis. 3DGS outperforms NeRF in training and inference speed but has substantially higher storage requirements.
To remedy this downside, we propose POTR, a post-training 3DGS codec built on two novel techniques.
First, POTR introduces a novel \changed{culling}{pruning} approach that uses a modified 3DGS rasterizer to efficiently calculate every splat's individual removal effect simultaneously. This technique results in 2-4× fewer splats than other post-training \changed{culling}{pruning} techniques and as a result also significantly accelerates inference with experiments demonstrating 1.5-2× faster inference than other compressed models.
Second, we propose a novel method to recompute lighting coefficients, significantly reducing their entropy without using any form of training. Our fast and highly parallel approach especially increases AC lighting coefficient sparsity, with experiments demonstrating increases from 70\% to 97\%, with minimal loss in quality.
Finally, we extend POTR with a simple fine-tuning scheme to further enhance \changed{culling}{pruning}, inference, and rate-distortion performance.
Experiments demonstrate that POTR, even without fine-tuning, consistently outperforms all other post-training compression techniques in both rate-distortion performance and inference speed.
\end{abstract}

\begin{IEEEkeywords}
3DGS, compression, spherical harmonics, energy compaction, pruning.
\end{IEEEkeywords}

%
\IEEEpeerreviewmaketitle

\section{Introduction}
\label{section:intro}

Synthesizing new views from a limited number of camera-captured images is a long-standing problem in computer graphics~\cite{ru2000multiview, lu2009streamcentric, song2024fewarnet}. Various techniques have been proposed to address this challenge, focusing on multiview video  coding~\cite{yamamoto2007multiview, purica2016multiview}, cross-view image matching~\cite{tain2022uav}, sparse compact representations~\cite{SMoE}, and real-time VR experiences~\cite{opendibr2023, courteaux2024dimreduction}.
Among the most promising solutions are Neural Radiance Fields (NeRF)\cite{mildenhall2020nerf} and 3D Gaussian Splatting (3DGS)\cite{kerbl3Dgaussians}. While NeRF has seen substantial year-over-year improvements\cite{barron2021mipnerf,barron2022mipnerf360,barron2023zipnerf}, 3DGS achieves better training and inference speeds through a scene representation based on feature-rich volumetric points called \emph{splats}.
This alternate representation is also the culprit behind 3DGS's biggest comparative downside to NeRF, namely, substantially higher storage requirements. For example, a simple unbounded scene with a central object typically results in a model size of 0.3-1.5 GB, often making transfer and storage challenging, particularly for on-demand applications. Furthermore, with 3DGS literature expanding toward larger scenes~\cite{hierarchicalgaussians24,liu2024citygaussian} and immersive video~\cite{wu20234dgaussians,luiten2023dynamic}, 3DGS compression is becoming increasingly important.

The literature has recognized the need for compression through numerous 3DGS-specific compression techniques~\cite{zhang2025gaussianspa, wu2024implicitgaussiansplattingefficient, lee2025compression, morgenstern2023compact, liu2024compgs, liu2025compgs++, hac2024, chen2025hac++, navaneet2023compact3d, wang2024endtoendratedistortionoptimized3d, girish2024eaglesefficientaccelerated3d, scaffoldgs, Lee_2024_CVPR, sun2024f, papantonakis2024reducing, wang2024contextgs, shin2025locality, cao2024lightweight, fang2024mini, zhang2024lp, hanson2025speedy, niemeyer2025radsplat, fang2024mini2, zhan2025cat, chen2025pcgs, chen2025megs, fan2023lightgaussian, xie2024mesongs, Niedermayr_2024_CVPR, lee2024safeguardgs, hanson2025pup, huang2025entropygs, liu2025efficientgs}, most of whom are \emph{in-training} compression techniques~\cite{zhang2025gaussianspa, chen2025megs, hac2024,wu2024implicitgaussiansplattingefficient,morgenstern2023compact, lee2025compression, navaneet2023compact3d, wang2024endtoendratedistortionoptimized3d, girish2024eaglesefficientaccelerated3d, scaffoldgs, Lee_2024_CVPR, sun2024f, papantonakis2024reducing, wang2024contextgs, shin2025locality, cao2024lightweight, fang2024mini, zhan2025cat, chen2025hac++, chen2025pcgs, liu2024compgs, liu2025compgs++, zhang2024lp, niemeyer2025radsplat, fang2024mini2, hanson2025speedy} that alter the training process to achieve smaller models.
While this works well, this level of control over the training process is not guaranteed as one might want to re-encode an existing 3DGS model, similarly to how videos and images are re-encoded. To this end, a fraction of the 3DGS compression literature focuses on \emph{post-training} compression~\cite{lee2024safeguardgs, Niedermayr_2024_CVPR,fan2023lightgaussian, xie2024mesongs, hanson2025pup, huang2025entropygs, liu2025efficientgs} as they start from an existing model.
Lack of control over the training process generally results in poorer rate-distortion (RD) performance. As a result, post-training compression approaches commonly incorporate a fine-tuning step to somewhat level the playing field. This step further trains the already trained model to recover some of the quality lost during compression.
Although this is a powerful way for a post-training codec to utilize the extra flexibility offered by the in-training compression paradigm, it side-steps the issue of \emph{post}-training compression.
Nevertheless, current post-training methods primarily focus on RD performance after fine-tuning.

To this end, we propose \textbf{POTR}, a \textbf{po}st-\textbf{tr}aining 3DGS codec that focuses predominantly on achieving strong RD performance without any fine-tuning. POTR significantly outperforms existing post-training methods, achieving up to a fourfold reduction in model size compared to the previous state-of-the-art. Additionally, our compressed models use far fewer splats and therefore render faster. This markedly better performance is achieved through two novel compression techniques. 

Our first novel technique introduces a new approach to splat removal. Where existing post-training \changed{culling}{pruning} methods rely on heuristics involving metrics such as a splat's size, opacity, and importance, we propose a \changed{culling}{pruning} technique that directly evaluates the impact of removing each splat on an objective quality metric. 
This approach allows our high-quality compressed models to use 2–4× fewer splats than other post-training methods across four common datasets.  Additionally, our \changed{culling}{pruning} method significantly accelerates inference, often achieving at least 50\% higher frames per second. At higher distortion levels, the performance gap widens further, with experiments showing examples of 2× higher frame rates than other post-training compressed models.

Our second novel technique focuses on spherical harmonics coefficients --- commonly referred to as lighting coefficients --- which represent a splat's view-dependent color. Lighting coefficients constitute over 80\% of the uncompressed model's size, making them vital to compression. To address this, we first represent a splat's colors for all training views using a single linear system. Next, we introduce a spherical harmonics energy compaction method that uses a heavily modified version of ridge regression to compute an alternate set of lighting coefficients. These new coefficients exhibit significantly lower entropy while producing nearly identical colors for relevant training views and generalizing better to novel views. Combined with quantization and entropy compression, we demonstrate that lighting coefficients are no longer the largest contributor to model size using our spherical harmonics energy compaction method. To our knowledge, this is the first post-training 3DGS compression method to non-trivially recompute lighting coefficients without any form of training.

In summary, the main contributions of this paper are:
\begin{itemize}
    \item Proposing an efficient method to evaluate the impact of removing each splat on an objective quality metric. Use this to design a \changed{culling}{pruning} strategy that significantly outperforms other post-training methods in both RD performance and inference speed, particularly at higher distortion levels.

    \item Introducing a fast, systematic approach to transform high-entropy spherical harmonics coefficients into low-entropy ones while preserving the splat's colors for relevant training views. Our method requires no training, is embarrassingly parallel, and improves generalization.
    
    \item Developing POTR, a fine-tuneless post-training codec that leverages the above two techniques to achieve state-of-the-art RD performance and inference speed, surpassing all other (fine-tuning-based) post-training methods. Additionally, we demonstrate that extending POTR  with a simple fine-tuning scheme further enhances RD performance and inference speed.
\end{itemize}

The remainder of this work is organized as follows. \Cref{section:related_work} briefly overviews the relevant literature. Next, \Cref{section:3dgs} discusses the necessary background on 3DGS which will be used extensively in \Cref{section:method} which provides a detailed description of our proposed compression methods and codec. \Cref{section:results} presents our results, experiments, and associated discussions. Finally, \Cref{section:conclusion} summarizes this work.

\section{Related work}
\label{section:related_work}
\subsection{NeRF compression}
Voxel-based techniques are a popular solution for enhancing the training and rendering speeds of NeRFs~\cite{mueller2022instant, Reiser2021ICCV}. However, these methods often lead to large storage overhead, for example, KiloNeRF~\cite{Reiser2021ICCV} necessitates the storage of thousands of neural networks. As a result, a large fraction of the NeRF compression literature focuses on voxel-based techniques. Voxel-based techniques are comparable to 3DGS in that their representation is more localized. This localization generally eases compression as it improves existing, or allows for new, compression techniques such as voxel pruning~\cite{Deng_2023_WACV, Li_2023_CVPR, zhao2023tinynerf}, transform coding~\cite{zhao2023tinynerf, rho2023masked, li2024nerfcodec, pham2024neural, lee2024ecrf, wang2023neural}, various forms of quantization~\cite{Li_2023_CVPR, takikawa2022variable, shin2024binary}, and specialized context models for entropy compression~\cite{Chen_2024_CVPR}.

\subsection{Point-cloud compression}
Point-cloud geometry is commonly compressed using octrees~\cite{wang2023inraprediction, Que_2021_CVPR, huang2020octsqueeze}. The octree structure is serialized by encoding the occupancy bits of the octree's nodes, which are subsequently compressed using entropy coding. 
To achieve higher compression ratios, several custom entropy models have been proposed. Examples include a tree-structured entropy model utilizing MLPs~\cite{huang2020octsqueeze}, an intra-prediction-based entropy model~\cite{wang2023inraprediction}, and a deep entropy model~\cite{Que_2021_CVPR}.

The compression of point-cloud attributes, such as color values and normal vectors, is another key area of research~\cite{cao2019pccsurvey}. While fully 3D methods often use octrees to encode attributes~\cite{ricardo2016raht, huang2006octree, zhang2018hierarchical}, alternative approaches where 3D data is mapped onto 2D planes, allowing the use of standard image and video compression techniques~\cite{schwarz2019emerging}, do occur.

\subsection{3D Gaussian Splatting compression}
To address 3DGS's stringent storage requirements, several compression techniques have been proposed~\cite{zhang2025gaussianspa, wu2024implicitgaussiansplattingefficient, lee2025compression, morgenstern2023compact, liu2024compgs, liu2025compgs++, hac2024, chen2025hac++, navaneet2023compact3d, wang2024endtoendratedistortionoptimized3d, girish2024eaglesefficientaccelerated3d, scaffoldgs, Lee_2024_CVPR, sun2024f, papantonakis2024reducing, wang2024contextgs, shin2025locality, cao2024lightweight, fang2024mini, zhang2024lp, hanson2025speedy, niemeyer2025radsplat, fang2024mini2, zhan2025cat, chen2025pcgs, chen2025megs, fan2023lightgaussian, xie2024mesongs, Niedermayr_2024_CVPR, lee2024safeguardgs, hanson2025pup, huang2025entropygs, liu2025efficientgs}. \changed{These}{Early} approaches are comprehensively surveyed in 3DGS.zip~\cite{3DGSzip2024}, which also introduces a methodology to compare different 3DGS compression techniques. Notably, the 3DGS literature bears close resemblance to point-cloud compression literature by, for example, compressing splat geometry using octrees~\cite{wang2024endtoendratedistortionoptimized3d, xie2024mesongs} and projecting 3D attributes onto 2D planes to allow for image-based compression~\cite{wu2024implicitgaussiansplattingefficient, morgenstern2023compact, lee2025compression}. Some works even directly use point cloud codecs, for example, G-PCC~\cite{chen2025pcgs, chen2025hac++, liu2024compgs, liu2025compgs++, huang2025entropygs}. While point-cloud-inspired techniques result in substantial RD gains, the largest gains of the 3DGS literature originate from 3DGS-specific techniques.

\changed{The}{A} first group of 3DGS-specific compression techniques focuses on reducing the number of splats. 
\added{Most approaches assign each splat a heuristic score, computed using factors such as sensitivity~\cite{Niedermayr_2024_CVPR}, opacity~\cite{chen2025megs, morgenstern2023compact, fan2023lightgaussian, scaffoldgs, hac2024, chen2025hac++, liu2024compgs, liu2025compgs++}, importance~\cite{girish2024eaglesefficientaccelerated3d, xie2024mesongs, fang2024mini2, zhang2024lp}, intersection~\cite{fang2024mini, fang2024mini2}, maximum contribution~\cite{niemeyer2025radsplat, zhang2024lp}, uncertainty~\cite{hanson2025pup, hanson2025speedy} or volume~\cite{fan2023lightgaussian, xie2024mesongs, huang2025entropygs}.
Splats with scores below a chosen threshold are then removed~\cite{fan2023lightgaussian, Niedermayr_2024_CVPR, xie2024mesongs, girish2024eaglesefficientaccelerated3d, morgenstern2023compact, papantonakis2024reducing, scaffoldgs, hac2024, chen2025hac++, liu2024compgs, liu2025compgs++}. Another more indirect approach is to apply a downward pressure on the number of splats during training by altering the loss function~\cite{zhang2025gaussianspa, zhan2025cat, hac2024, chen2025hac++, wang2024endtoendratedistortionoptimized3d, Lee_2024_CVPR, sun2024f, navaneet2023compact3d, shin2025locality, wang2024contextgs, niemeyer2025radsplat}.
}
Both methods have been quite successful at reducing the number of splats, so much so that the largest inference time improvements have thus far originated from the compression literature, highlighting its contribution to Gaussian Splatting beyond size reduction.

A second group of 3DGS-specific compression techniques addresses the substantial storage requirements of lighting information. In 3DGS's original formulation, lighting information is modeled using spherical harmonics (SH) and their coefficients. The latter are stored and account for over 80\% of all attributes, leading some works to avoid explicitly saving lighting coefficients altogether.
Instead, latent attributes are stored and processed \changed{through}{using} an MLP to generate either direct color outputs~\cite{zhan2025cat, scaffoldgs, hac2024, Lee_2024_CVPR, wang2024contextgs, cao2024lightweight, chen2025hac++, liu2024compgs, liu2025compgs++}, or lighting coefficients~\cite{wu2024implicitgaussiansplattingefficient, girish2024eaglesefficientaccelerated3d, sun2024f, lee2025compression}.
For methods that store the lighting coefficients explicitly, additional techniques are applied to lower the lighting coefficients' entropy. Examples include trainable masks~\cite{wang2024endtoendratedistortionoptimized3d, shin2025locality}, SH band pruning after densification~\cite{papantonakis2024reducing}, distillation of higher-degree SH terms to lower degrees~\cite{fan2023lightgaussian}, and transformations such as Region-Adaptive Hierarchical Transform~\cite{xie2024mesongs, fang2024mini} and JPEG XL~\cite{morgenstern2023compact}.

To summarize, a range of techniques are used to compress 3DGS models, but those that focus on reducing the number of splats or compressing lighting information are the most crucial to achieving strong RD performance. Compressing other properties --- such as position, scale, rotation, and opacity --- is comparatively less important, as these contribute only a small portion to the overall uncompressed size.
Moreover, reducing the number of splats not only enhances RD performance but also boosts inference speed, underscoring the importance of compression techniques beyond just storage efficiency.

\section{3D Gaussian splatting}
\label{section:3dgs}
3DGS models consist of a collection of splats, each with a center $\vect{\mu} \in \mathbb{R}^{3}$ in space. For a point in space $\vect{p} \in \mathbb{R}^{3}$, the splat's density is proportional to a 3D Gaussian with a covariance matrix $\matr{\Sigma} \in \mathbb{R}^{3 \times 3}$:
\begin{equation}
    \mathcal{G}(\vect{p}) = \exp{(-\frac{1}{2}(\vect{p}-\vect{\mu})^T \matr{\Sigma}^{-1} (\vect{p}-\vect{\mu}))} \, .
\end{equation}
To render a 3DGS model for a given camera, a splat's density is first projected into screen space. This is done efficiently using the affine invariance of multivariate Gaussians by approximating the projective transformation as an affine transformation. The resulting 2D density is therefore again proportional to a 2D Gaussian and denoted as $\mathcal{G}^{\text{2D}}$. Next, the splats are alpha-composited by associating each splat with a layer and ordering these layers according to the splats' approximate depth.
Upon rasterizing a splat's layer, a pixel at position $\vect{x} \in \mathbb{R}^{2}$ has an opacity
\begin{equation}
    \alpha = o \cdot \mathcal{G}^{2D}(\vect{x}) \, ,
\end{equation}
where $o$ represents the location-independent base opacity of the splat. The yet undetermined fraction of a pixel's color after alpha compositing the first $k$ layers is termed the transmittance and is defined as
\begin{equation}
    T_k = \prod_{j=0}^{k-1} (1-\alpha_j) \, .
\end{equation}
After alpha-compositing all $K$ visible splats over a black background, the final color of a pixel located at $\vect{x} \in \mathbb{R}^{2}$ is given by
\begin{equation}
\label{eq:alpha_blending}
\vect{c} (\vect{x}) = \sum_{k = 0}^{K-1} T_k \alpha_k \vect{c}_k 
\end{equation}
where $\vect{c}_k$ is the color of the $k$-th splat in the alpha compositing process.

\begin{figure}
    \centering
     \begin{overpic}[width=\linewidth]{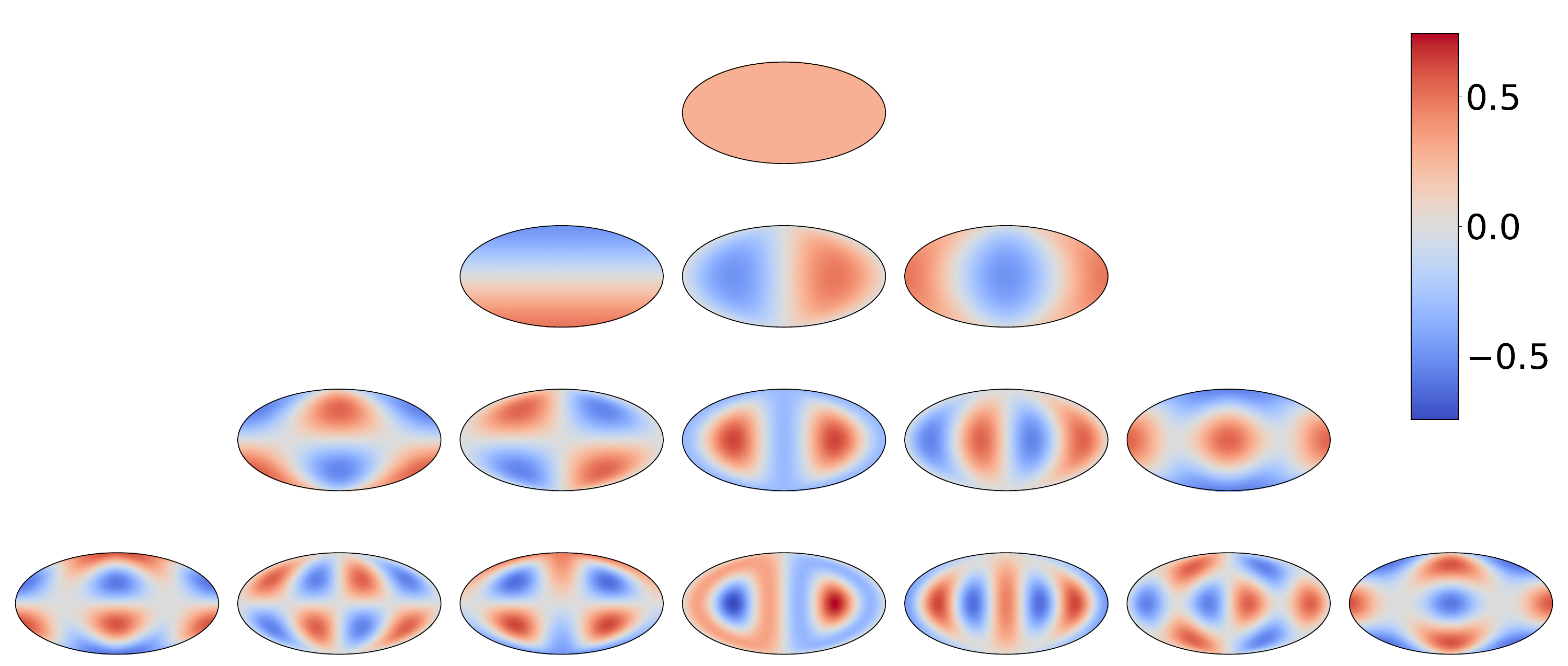}
        \newcommand{\xoffset}{48.6}
        \newcommand{\yoffset}{39.2}
        \newcommand{\xstep}{14.2}
        \newcommand{\ystep}{10.42}

        \FPadd\xpos{\xoffset}{0}
        \FPadd\ypos{\yoffset}{0}
        \put(\xpos,\ypos){\footnotesize $Y_1$}

        \FPadd\xpos{\xoffset}{0}
        \FPsub\xpos{\xpos}{\xstep}
        \FPsub\ypos{\ypos}{\ystep}
        \put(\xpos,\ypos){\footnotesize $Y_2$}
        \FPadd\xpos{\xpos}{\xstep}
        \put(\xpos,\ypos){\footnotesize $Y_3$}
        \FPadd\xpos{\xpos}{\xstep}
        \put(\xpos,\ypos){\footnotesize $Y_4$}

        \FPadd\xpos{\xoffset}{0}
        \FPsub\xpos{\xpos}{\xstep}
        \FPsub\xpos{\xpos}{\xstep}
        \FPsub\ypos{\ypos}{\ystep}
        \put(\xpos,\ypos){\footnotesize $Y_5$}
        \FPadd\xpos{\xpos}{\xstep}
        \put(\xpos,\ypos){\footnotesize $Y_6$}
        \FPadd\xpos{\xpos}{\xstep}
        \put(\xpos,\ypos){\footnotesize $Y_7$}
        \FPadd\xpos{\xpos}{\xstep}
        \put(\xpos,\ypos){\footnotesize $Y_8$}
        \FPadd\xpos{\xpos}{\xstep}
        \put(\xpos,\ypos){\footnotesize $Y_9$}

        \FPadd\xpos{\xoffset}{0}
        \FPsub\xpos{\xpos}{\xstep}
        \FPsub\xpos{\xpos}{\xstep}
        \FPsub\xpos{\xpos}{\xstep}
        \FPsub\ypos{\ypos}{\ystep}
        \put(\xpos,\ypos){\footnotesize $Y_{10}$}
        \FPadd\xpos{\xpos}{\xstep}
        \put(\xpos,\ypos){\footnotesize $Y_{11}$}
        \FPadd\xpos{\xpos}{\xstep}
        \put(\xpos,\ypos){\footnotesize $Y_{12}$}
        \FPadd\xpos{\xpos}{\xstep}
        \put(\xpos,\ypos){\footnotesize $Y_{13}$}
        \FPadd\xpos{\xpos}{\xstep}
        \put(\xpos,\ypos){\footnotesize $Y_{14}$}
        \FPadd\xpos{\xpos}{\xstep}
        \put(\xpos,\ypos){\footnotesize $Y_{15}$}
        \FPadd\xpos{\xpos}{\xstep}
        \put(\xpos,\ypos){\footnotesize $Y_{16}$}
    \end{overpic}
    \caption{Mollweide projection of the first 16 real spherical harmonics.}
    \label{fig:all_spherical_harmonics}
\end{figure}

While a splat's color is monochromatic across the pixels of a still frame, it is view-dependent based on the camera's position through a spherical function defined using real \shs{}. 3DGS's original formulation uses all real \shs{} with a degree $l \leq 3$, all of whom are shown in \Cref{fig:all_spherical_harmonics}. To ease iterating these functions, each degree-order pair $(l,m)$ is mapped to a unique index using $(l,m) \rightarrow l(l+1) + m + 1$.
The first \sh{} $Y_1$ is often singled out and called the DC \sh{} as it is the only basis that is view-independent and the only basis that is not zero-meaned. All other \shs{} are called AC \shs{}.
The value of a splat's color channel is determined by the $M$ lighting coefficients associated with that splat and channel. It is evaluated based on the direction $\vect{d}$ extending from the camera to the splat center, as follows: 
\begin{equation}
\label{eq:energy_compac_decomp_one_camera}
    C(\vect{d}) = \sum_{i=1}^M L_i Y_i(\vect{d}),
\end{equation}
where $L_i$ represents the lighting coefficient corresponding to $Y_i$.

\begin{figure*}
    \centering
    \includegraphics[width=\linewidth]{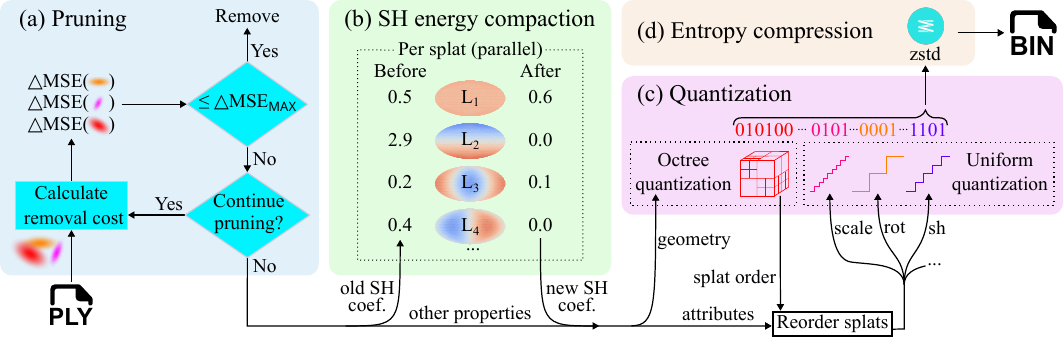}
    \caption{
\label{fig:encoder_design}
Simplified overview of POTR.
(a) Splats are removed across multiple \changed{culling}{pruning} iterations based on the change in the model's mean square error (MSE) upon their removal. (b) Spherical harmonic coefficients are energy compacted, yielding a new set of lighting coefficients with a lower entropy. (c) Splat geometry is quantized and serialized using an octree, then attributes are uniformly quantized and serialized using the spatial order implied by the depth-first traversal of the octree. (d) The serialized data is entropy compressed using zstd, resulting in the final compressed bitstream.
    }
\end{figure*}

\section{Method}
\label{section:method}

This section discusses the different steps of our proposed encoder, for which \Cref{fig:encoder_design} provides an overview. First, \Cref{section:splat_importance} introduces a preliminary metric that is used throughout POTR. Next, we focus on splat removal, the first step of our encoder, with \Cref{section:splat_removal_effect} detailing how our proposed method can efficiently and accurately calculate the effect of a splat's removal. This information is used in \Cref{section:culling_controller} as the basis for our proposed splat-removal method. After \changed{culling}{pruning}, we use a novel spherical harmonics energy compaction method to reduce the lighting coefficients' entropy in \Cref{section:sh_compaction}. To complete our codec, \Cref{section:encoder} discusses tangential topics such as quantization, serialization, entropy compression, and how these are all combined. Finally, \Cref{section:fine-tuning} discusses how our codec can be adapted to incorporate fine-tuning.

\subsection{Splat importance}
\label{section:splat_importance}
POTR utilizes lossy compression, during which a splat may be removed or its properties altered. The impact of modifying or removing a splat on the quality of the model varies heavily based on the splat in question. To help upcoming lossy steps gauge this impact, we define the importance $I_k (\vect{s})$ of the $k$-th splat to a camera $\vect{s}$ as the fraction of the camera's synthesized image that originates from the $k$-th splat, i.e. 
\begin{equation}
I_k (\vect{s}) = \frac{1}{P} \sum_{\vect{x}} T_{i_k} \alpha_{i_k} \,
\end{equation}
where $P$ is the number of pixels and $i_k$ is the rank of the $k$-th splat in the alpha-compositing process for the pixel $\vect{x}$. The importance of a splat can be generalized to account for the entire scene, encompassing all $N$ cameras, as follows: 
\begin{equation}
I_k = \frac{1}{N} \sum_{\vect{s}} I_k (\vect{s}) \, .
\end{equation}
\Cref{fig:cumulative_contribution_splats} shows that a small fraction of splats draws the vast majority of a scene, highlighting the immense potential of selective loss introduction in a compression context.

\input{plots/cumulative_contribution}

\subsection{Effect of a splat's removal}
\label{section:splat_removal_effect}
Removing splats is a common way to reduce the final file size. To help determine if a splat is to be kept or removed, we desire a score that describes the impact of a splat's removal on the quality of the scene.
Other post-training compression methods rely solely on heuristics to set this score, for example, using a splat's size or importance $I_k$. However, these approaches tend to be sub-optimal as they overlook crucial factors such as color information and the intricate geometry of the model. For instance, the effect of removing a red splat varies depending on the splats that lie behind it. If this is another red splat, the change in the rendered image will be relatively minor. Conversely, the visual difference will be much more pronounced if a blue splat lies behind the red splat. Metrics such as a splat's importance $I_k$ fail to capture such nuances.

To address this limitation, we propose to modify the original rasterizer's forward rendering pass to accurately evaluate the effect of removing a splat, for all splats simultaneously. Our proposed modification utilizes readily available data, specifically the opacities of the contributing splats for each pixel and the partial colors $\{\vect{P}_i\}$. The latter is a pixel's color after $i$ splats have been considered in the alpha compositing process and can be expressed as
\begin{equation}
\label{eq:partial_color}
\vect{P}_i = \sum_{k = 0}^{i-1} T_k \alpha_k \vect{c}_k \, .
\end{equation}
\added{
Removing the $k$-th splat with rank $i_k$ in the alpha-compositing process of a given pixel renormalizes the transmittance of all subsequent splats by a factor of $\tfrac{1}{1 - \alpha_{i_k}}$.
Using this observation together with \Cref{eq:alpha_blending,eq:partial_color}, we derive a compact expression for the pruning difference $\vect{PD}_k(\vect{x})$, which quantifies the change in a pixel’s color resulting from the removal of the $k$-th splat:}
\begin{equation}
\begin{aligned}    
\label{Eq:cull_difference_partial_color}
    \added{\vect{PD}_k} (\vect{x})
    &= \Tilde{\vect{c}}_k(\vect{x}) - \vect{c}(\vect{x}) \\
    &= (\sum_{j = 0}^{i_k-1} T_j \alpha_j \vect{c}_j  + \frac{1}{1 - \alpha_{i_k}} \sum_{j=i_k+1}^{K-1}  T_j \alpha_j \vect{c}_j) - \vect{c}(\vect{x}) \\
    &= (\vect{P}_{i_k} + \frac{1}{1 - \alpha_{i_k}} (\vect{P}_K - \vect{P}_{i_k+1})) - \vect{P}_K \\
    &= \vect{P}_{i_k} - \frac{1}{1 - \alpha_{i_k}} \vect{P}_{i_k+1} + \frac{\alpha_{i_k}}{1 - \alpha_{i_k}} \vect{P}_K \, ,
\end{aligned}
\end{equation}
\added{
where $\vect{c}(\vect{x})$ and $\Tilde{\vect{c}}_k(\vect{x})$ denote a pixel’s color before and after removing the $k$-th splat, respectively.}

The \changed{cull}{pruning} difference is crucial for determining the impact of each splat's removal on an objective quality metric. In this initial work, the squared error is used due to its simplicity and relation to the PSNR. The difference in squared error, per pixel and color channel, as a result of removing the $k$-th splat is given by
\begin{equation}
\begin{aligned}
\label{Eq:pixel_sq_error_def}
    \vect{\triangle{SE}}_k (\vect{x})
    &= (\Tilde{\vect{c}}_k(\vect{x}) - \vect{c}_{\text{t}}(\vect{x}))^2 - (\vect{c}(\vect{x}) - \vect{c}_{\text{t}}(\vect{x}))^2 \\
    &= (\Tilde{\vect{c}}_k(\vect{x}) - \vect{c}(\vect{x}))(\Tilde{\vect{c}}_k(\vect{x}) + \vect{c}(\vect{x}) - 2 \vect{c}_{\text{t}}(\vect{x})) \\
    &= (\added{\vect{PD}_k} (\vect{x}))^2 + 2~\added{\vect{PD}_k} (\vect{x})~(\vect{c}(\vect{x})-\vect{c}_{\text{t}}(\vect{x})) \, ,
\end{aligned}
\end{equation}
\added{where $\vect{c}_{\text{t}}(\vect{x})$ is a fixed reference target color, defined as the initial synthesized color of the uncompressed model}\footnote{
Another valid choice for the target color $\vect{c}_{\text{t}}(\vect{x})$ is the ground truth color. However, our limited experiments indicate that its usage does not necessarily yield better results. Furthermore, since this work focuses on \emph{post}-training compression, ground truth images are not always available.
} \added{such that initially $\vect{c}_{\text{t}}(\vect{x}) = \vect{c}(\vect{x})$. In \Cref{section:culling_controller}, we will recompute $\vect{\triangle SE}_k(\vect{x})$ after removing splats, which changes $\vect{c}(\vect{x})$, $\Tilde{\vect{c}}_k(\vect{x})$, and consequently $\vect{PD}_k(\vect{x})$, but not $\vect{c}_{\text{t}}(\vect{x})$.}
By averaging the change in squared error across all $N$ cameras, $P$ pixels, and 3 color channels, we arrive at the overall effect of the $k$-th splat's removal on the mean squared error of the model:
\begin{equation}
    \triangle{\text{MSE}_k} = \frac{1}{N} \sum_{\vect{s}} \frac{1}{P} \sum_{\vect{x}} \frac{1}{3} \vect{\triangle{SE}}_k (\vect{x}) \cdot \vect{1} \, .
\end{equation}

Finally, it should be noted that our CUDA implementation of the above method is deliberately designed to be interoperable with the standard forward-backward render pass used during training. The forward pass comes at no extra cost if our altered render pass is already executed to calculate the splats' removal effects. Future work could explore how our implementation can be properly integrated into the training process.

\subsection{Iterative \changed{culling}{pruning}}
\label{section:culling_controller}
\begin{figure}
    \centering
    \begin{tikzpicture}
\begin{axis}[
    axis lines=middle,
    width=\linewidth,
    height=.45\linewidth,
    xlabel={$x$},
    ylabel={$m(x)$},
    xmin=-1, xmax=1,
    ymin=0, ymax=1,
    ytick={0, 0.5, 1.0},
    samples=100,
    domain=-1:1,
    smooth,
    grid=both,
    major grid style={line width=0.8pt, draw=gray!20},
]
\pgfmathsetmacro{\a}{10}
\addplot[
    thick,
    domain=-1:0,
    color=blue
]
{(1/\a) * (sqrt(1 - 2*\a*x) - 1)};
\addplot[
    thick,
    domain=0:1,
    color=blue
]
{x};

\end{axis}
    \end{tikzpicture}
    \caption{
        \label{fig:culling_mapping_function}
        The mapping function $m(x)$ for $a=10$.
    }
\end{figure}
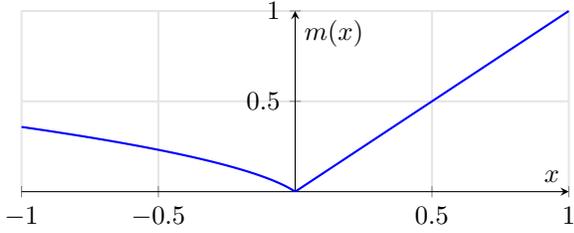

The \emph{\changed{culling}{pruning} controller} manages the removal of splats based on the $\triangle{\text{MSE}_k}$ values. It operates under the assumption that the cumulative effect of removing a group of splats $\mathcal{S}$ is approximately the sum of the individual effects, i.e. 
\begin{equation}
    \triangle{\text{MSE}_\mathcal{S}} \approx \sum_{k \in \mathcal{S}} \triangle{\text{MSE}_k} \, .
\end{equation}
This assumption, which we call the \textit{\changed{culling}{pruning} approximation}, tends to grossly underestimate $\triangle{\text{MSE}_\mathcal{S}}$ when removing too many splats at once. To counteract this limitation, we propose an iterative approach with multiple \changed{culling}{pruning} iterations where all $\triangle{\text{MSE}_k}$ values are recalculated in between \changed{culling}{pruning} iterations. In each iteration, the set of splats $\mathcal{S}$ to be removed is constructed by iteratively adding splats to an empty set until the \changed{culling}{pruning} budget $\mathcal{B}_{\text{\changed{cull}{prune}}}$ is reached. We use the importance of a splat $I_k$ as a simple estimate for the splat's ability to distort the \changed{culling}{pruning} approximation, defining the \changed{culling}{pruning} budget to be reached when
\begin{equation}
    \mathcal{B}_{\text{\changed{cull}{prune}}} \leq \sum_{k \in \mathcal{S}} I_k \, .
\end{equation}
The \changed{culling}{pruning} controller aims to remove all splats where $\triangle{\text{MSE}_k} < \triangle{\text{MSE}_{\texttt{MAX}}}$, and adjusts its \changed{culling}{pruning} budget accordingly every iteration to achieve this:
\begin{equation}
    \mathcal{B}_{\text{\changed{cull}{prune}}} = \frac{1}{\text{\# remaining iterations}} \sum_{k \in \{\,j \,|\, \triangle{\text{MSE}_j} < \triangle{\text{MSE}_{\texttt{MAX}}}\,\}} I_k \,.
\end{equation}
The \changed{culling}{pruning} controller adds splats to $\mathcal{S}$ in ascending order of their $m(\frac{\triangle{\text{MSE}_k}}{\triangle{\text{MSE}_{\texttt{MAX}}}\,})$ value where
\begin{equation}
m(x) = 
\begin{cases} 
  x & x\geq 0 \\
  \frac{1}{a} (\sqrt{1 - 2a \cdot x} - 1) & x < 0
\end{cases}
\end{equation}
is a mapping function (see \Cref{fig:culling_mapping_function}). This function is designed to prioritize the removal of splats with a small or negative  $\triangle{\text{MSE}_k}$ value, with the parameter $a$ controlling the balance between these two objectives. By prioritizing splats with small $\triangle{\text{MSE}_k}$, often more than 50\% of the splats can be removed in just one iteration, accelerating subsequent \changed{culling}{pruning}. On the other hand, focusing on splats with a negative $\triangle{\text{MSE}_k}$ value helps minimize quality loss during the \changed{culling}{pruning} process.

\subsection{Spherical harmonics energy compaction}
\label{section:sh_compaction}
Spherical harmonics, and their coefficients, define a splat's color for every direction. However, in practice, most splats are sampled only from a small subset of directions. This phenomenon arises for various reasons, including occlusion between splats, and the observer's restricted range of motion. This section's key idea is to use this phenomenon to find alternate, easier-to-compress, lighting coefficients that retain the color within the sampled subset while being agnostic to color changes outside the sampled subset. Since color changes only occur for unsampled directions, the model's overall quality is unaffected.

To find an alternate set of lighting coefficients, we first observe that \Cref{eq:energy_compac_decomp_one_camera} holds for all $N$ \added{training} cameras resulting in $N$ equations per splat that can be rewritten as follows:
\begin{equation}
\underbrace{
    \begin{psmallmatrix}
    C(\vect{d}_1) \\ C(\vect{d}_2) \\  \vdots \\ C(\vect{d}_N) 
    \end{psmallmatrix}
}_{\vect{C}}
=
\underbrace{
    \begin{psmallmatrix}
    Y_{1}(\vect{d}_1) & Y_{2}(\vect{d}_1) & \cdots & Y_{M}(\vect{d}_1) \\
    Y_{1}(\vect{d}_2) & Y_{2}(\vect{d}_2) & \cdots & Y_{M}(\vect{d}_2) \\
    \vdots & \vdots & \ddots & \vdots \\
    Y_{1}(\vect{d}_N) & Y_{2}(\vect{d}_N) & \cdots & Y_{M}(\vect{d}_N) \\
    \end{psmallmatrix}
}_{\matr{Y}}
\times 
\underbrace{
    \begin{psmallmatrix}
    L_{1} \\ L_{2} \\ \vdots \\ L_{M}
    \end{psmallmatrix}
}_{\vect{L}}
\label{Eq:energy_comp_lin_sys}
\end{equation}
We propose to use conventional energy compaction to find a new $\vect{L}$, denoted as $\vect{L}'$, located on the RD Pareto front \added{by using all $N$ training cameras}. This work uses ridge regression to define the optimal trade-off:
\begin{equation}
    \vect{L}' = \argmin_{\vect{x} \in \mathbb{R}^{M\times1}} (\lVert \matr{Y} \vect{x} - \vect{C} \rVert^2_2 + \lVert \matr{\Gamma} \vect{x} \rVert^2_2)
    \label{Eq:energy_comp_L2_reg_def}
\end{equation}
where $\lVert \cdot \rVert_2$ is the Euclidean norm of a vector, and $\matr{\Gamma}$ denotes the Tikhonov matrix which defines how the lighting coefficients are regularized. The Tikhonov matrix is chosen as follows:
\begin{equation}
\matr{\Gamma} = \sqrt{\lambda}
\left(
\begin{array}{c|c}
    0 & \matr{0}_{1 \times M-1} \\
    \hline
    \matr{0}_{M-1 \times 1} & \matr{I}_{M-1 \times M-1}
\end{array}
\right)
\end{equation}
where $\lambda$ is the regularization coefficient and $\matr{I}$ denotes the identity matrix. The top-left element is zero as we do not wish to regularize the DC lighting coefficient. The benefit of ridge regression, and the reason it was chosen for this work as opposed to other regularization techniques, is its closed-form solution:
\begin{equation}
\label{Eq:energy_comp_L2_reg_sol}
    \vect{L}' = (\matr{Y}^T \matr{Y} + \matr{\Gamma}^T \matr{\Gamma})^{-1} \matr{Y}^T \vect{C}
\end{equation}
A closed-form solution is essential as $\vect{L}'$ has to be calculated for each splat, with models having hundreds of thousands to millions of splats.

With the core idea in place, we propose four additional modifications to improve RD performance further:
\subsubsection{Importance}
\Cref{Eq:energy_comp_L2_reg_def} punishes each deviation from $\vect{C}$ equally, regardless of the importance of the splat to the camera. However, we expect a deviation in a splat's color (as perceived by a given camera) to affect the quality of a model approximately in accordance with the importance of that splat to that camera. We can take this into account by doing the following substitutions in \Cref{Eq:energy_comp_L2_reg_def} (and \Cref{Eq:energy_comp_L2_reg_sol}):
\begin{equation}
    \vect{C} \gets \vect{C} \odot_{\text{col}} (I_k(\vect{s}_1), ..., I_k(\vect{s}_N))^T
\end{equation}
\begin{equation}
    \matr{Y} \gets \matr{Y} \odot_{\text{col}} (I_k(\vect{s}_1), ..., I_k(\vect{s}_N))^T
\end{equation}
where $\odot_{\text{col}}$ represents a column element-wise multiplication.

\subsubsection{Color model}
RGB lighting coefficients correlate highly across color channels. By switching to a luminance-chrominance color model, we decorrelate the color channels and as a result, reduce the total energy. Additionally, chrominance channels can be more aggressively regularized, as human vision is less sensitive to chrominance changes. For this work, we use the YCoCg color model and regularize the chrominance channels thrice as hard by setting $\lambda=\lambda_{\text{Y}}=\frac{1}{3}\lambda_{\text{Co}}=\frac{1}{3}\lambda_{\text{Cg}}$. Surprisingly, to the best of our knowledge, this is the first post-training 3DGS compression work to use a non-RGB color model.

\subsubsection{Sparsity}
Due to the limited set of sampled directions, column vectors of $\matr{Y}$ often exhibit quasi-parallelism among themselves, suggesting that lighting coefficients can be sparsified. Unfortunately, ridge regression spreads energy across parallel column vectors, leading to a non-sparse and under-regularization solution. We address this by zeroing out the $i$-th column of $\matr{Y}$, which leads to $L'_i = 0$, if a preceding column is 'sufficiently parallel' to it. Two column vectors are said to be sufficiently parallel if the absolute value of their cosine similarity is larger than some threshold $\alpha$.

\subsubsection{Two-pass regularization}
Ridge regression forces a large number of AC lighting coefficients to be \emph{almost} zero such that these lighting coefficients will be quantized to zero in \Cref{section:encoder}. To reduce the quantization loss and to present ridge regression with a more true representation of its flexibility, a two-pass regularization system is used. The first pass identifies all almost zero AC coefficients and subsequently forces them to zero. The second pass finds the optimal values for the remaining non-zero lighting coefficients.

In summary, we demonstrate that a splat's color for all \added{$N$ training} cameras can be expressed as one linear system of lighting coefficients. We then propose to use ridge regression to find a new set of lighting coefficients with a lower entropy. Finally, we improve upon this basic energy compaction method through importance weighting, a luminance-chrominance model, the removal of parallel column vectors, and a two-pass regularization system.

\subsection{Codec}
\label{section:encoder}
We use the proposed \changed{culling}{pruning} and spherical harmonics energy compaction methods to create a codec (see \Cref{fig:encoder_design}). 
The encoder first removes redundant splats before spherical harmonics energy compaction reduces the entropy of the remaining splats' lighting coefficients. Subsequently, the geometry and attributes are quantized and serialized contiguously into a bitstream, iterating over all splats before serializing the next property. Finally, the bitstream is compressed using zstd\footnote{\url{https://github.com/facebook/zstd}}, an out-of-the-box lossless compression algorithm,  yielding the final compressed bitstream. The decoder follows the opposite process by decompressing using zstd, deserializing, and dequantizing the bitstream, in that order.

Quantization is performed using multiple strategies. The splat attributes \added{(scale, rotation, opacity, and SH coef.)} are uniformly quantized \added{and dequantized} using the transformations
\begin{equation}
\label{eq:quant_dequant}
    \hat{x}= \left \lfloor \frac{1}{2} + x \cdot \texttt{SF} \right \rfloor \,
    \added{
    \quad 
    \text{and}
    \quad 
    \tilde{x} = \frac{\hat{x}}{\texttt{SF}} \, ,}
\end{equation}
where $\texttt{SF}$ is the scale factor, a hyperparameter tailored to each attribute. Unlike attribute quantization, the geometry of the splats is quantized using an octree structure. Each splat's position $\vect{\mu}$ is quantized to the center of an octree leaf $\vect{\hat{\mu}}$. The octree is constructed by repeatedly splitting leaves containing multiple splats or failing to meet the precision criterion
\begin{equation}
    \lVert \vect{\mu} - \vect{\hat{\mu}} \rVert_2 < \added{\max\Big(\gamma,} \; \beta \cdot \min_{\vect{s}} \lVert \vect{\mu} - \vect{\mu}_{\vect{s}} \rVert_2 \added{\Big)} \, ,
\end{equation}
where $\lVert \cdot \rVert_2$ denotes the Euclidean distance, $\vect{\mu}_{\vect{s}}$ is the position of the eye of camera $\vect{s}$, and $\beta$ \changed{is}{and $\gamma$ are hyperparameters}. This criterion leverages the observation that positional changes of distant splats have minimal impact on their screen-space positions. \added{
We serialize the octree using a depth-first traversal. Upon visiting a node for the first time, we append its occupancy to the bitstream. The occupancy of an octree's node is a single byte in which each bit corresponds to one of the eight octants and indicates whether that octant contains a child node.
}

Our codec also employs some additional optimizations. For instance, DC lighting coefficients are differentially encoded, \changed{one}{the real } quaternion component of the rotation is omitted, and opacity reconstruction values are shifted off-center during dequantization \added{by using $\tilde{x} = \frac{\hat{x}+0.25}{\texttt{SF}}$ (see \Cref{eq:quant_dequant})}. We also note that the order of the steps shown in \Cref{fig:encoder_design} is flexible. For example, spherical harmonics energy compaction can be applied earlier, e.g., after half of the \changed{culling}{pruning} iterations, to allow the remaining iterations to account for the changes introduced by energy compaction.

\begin{table*}
    \newcommand{\first}[1]{\cellcolor{lightred}#1}
    \newcommand{\second}[1]{\cellcolor{lightorange}#1}
    \newcommand{\third}[1]{\cellcolor{lightyellow}#1}
    \addtolength{\tabcolsep}{-0.55em}
\begin{tabular}{lcc|ccccc|ccccc|ccccc|ccccc}
\toprule
\multirow{2}{*}{Method} & \multirow{2}{*}{\makecell{\added{Lossless}\\\added{Format}}} & \multirow{2}{*}{FT} & \multicolumn{5}{c|}{Tanks And Temples} & \multicolumn{5}{c|}{Mip-NeRF 360} & \multicolumn{5}{c|}{Deep Blending} & \multicolumn{5}{c}{NeRF-Synthetic} \\
 & &
 & \tiny PSNR$\uparrow$ & \tiny SSIM$\uparrow$ & \tiny LPIPS$\downarrow$ & \tiny \makecell{Size \\ MB}$\downarrow$ & \tiny \makecell{\# Splats \\ x1,000}$\downarrow$
 & \tiny PSNR$\uparrow$ & \tiny SSIM$\uparrow$ & \tiny LPIPS$\downarrow$ & \tiny \makecell{Size \\ MB}$\downarrow$ & \tiny \makecell{\# Splats \\ x1,000}$\downarrow$
 & \tiny PSNR$\uparrow$ & \tiny SSIM$\uparrow$ & \tiny LPIPS$\downarrow$ & \tiny \makecell{Size \\ MB}$\downarrow$ & \tiny \makecell{\# Splats \\ x1,000}$\downarrow$
 & \tiny PSNR$\uparrow$ & \tiny SSIM$\uparrow$ & \tiny LPIPS$\downarrow$ & \tiny \makecell{Size \\ MB}$\downarrow$ & \tiny \makecell{\# Splats \\ x1,000}$\downarrow$ \\
\midrule
Baseline~\cite{kerbl3Dgaussians} & \added{-} & -
    & 23.36 & .838 & .186 & 442 & 1,784
    & 27.47 & .821 & .206 & 834 & 3,362
    & 29.43 & .898 & .246 & 738 & 2,976
    & 33.79 & .970 & .030 & 71.8 & 289 \\
\midrule
MesonGS~\cite{xie2024mesongs} & \added{npz} & No
    & 22.84 & .820 & .211 & \third{17.3} & 1,163
    & 26.22 & .785 & .249 & \third{29.7} & 2,147
    & 28.70 & .890 & .271 & 29.0 & 2,023
    & 32.49 & .962 & .039 & 3.52 & 210 \\
MesonGS-FT~\cite{xie2024mesongs} & \added{npz} & Yes
    & \third{23.16} & .832 & .200 & \third{17.3} & 1,163
    & 27.03 & .805 & .231 & \third{29.7} & 2,147
    & \first{29.54} & \second{.901} & \third{.255} & 29.0 & 2,023 
    & 33.25 & .966 & .035 & \third{3.51}  & 210 \\
C3DGS~\cite{Niedermayr_2024_CVPR} & \added{npz} & Yes
    & 23.13 & \second{.834} & \third{.195} & 18.3   & 1,483 
    & \second{27.16} & \first{.811} & \second{.226} & 30.4  & 2,973 
    & \third{29.35} & \second{.901} & .256 & \third{26.7}  & 2,613
    & \third{33.31} & \first{.968} & \third{.033} & 3.99 & 270 \\
LightGaussian~\cite{Lee_2024_CVPR} & \added{npz} & Yes
    & 22.86 & .817 & .215 & 29.1 & \second{607}
    & 26.75	& .805 & .244 & 54.5 & \first{1,143}
    & 29.16 & .894 & .261 & 47.9 & \third{1,012}
    & 31.33	& .956 & .047 & 4.92 & \first{99} \\
\midrule
\multirow{2}{*}{POTR (ours)} & \added{npz} & \multirow{2}{*}{No}
    & \multirowcell{2}{\second{23.27}} & \multirowcell{2}{\second{.834}} & \multirowcell{2}{\second{.191}} & \added{12.8} & \multirowcell{2}{\third{690}}
    & \multirowcell{2}{\third{27.08}} & \multirowcell{2}{\third{.806}} & \multirowcell{2}{\second{.226}} & \added{29.3} & \multirowcell{2}{\third{1,500}}
    & \multirowcell{2}{29.31} & \multirowcell{2}{.897} & \multirowcell{2}{\second{.253}} & \added{18.6} & \multirowcell{2}{\second{785}}
    & \multirowcell{2}{\second{33.34}} & \multirowcell{2}{\first{.968}} & \multirowcell{2}{\first{.032}} & \added{3.94} & \multirowcell{2}{\third{154}} \\
& \added{zstd} & & & & & \second{11.3} & & & & & \second{26.0} & & & & & \second{16.5} & & & & & \second{3.50} & \\
\multirow{2}{*}{POTR-FT (ours)} & \added{npz} & \multirow{2}{*}{Yes}
    & \multirowcell{2}{\first{23.34}} & \multirowcell{2}{\first{.837}} & \multirowcell{2}{\first{.189}} & \added{10.6} & \multirowcell{2}{\first{594}}
    & \multirowcell{2}{\first{27.20}} & \multirowcell{2}{\second{.808}} & \multirowcell{2}{\first{.223}} & \added{24.1} & \multirowcell{2}{\second{1,285}}
    & \multirowcell{2}{\second{29.44}} & \multirowcell{2}{\first{.902}} & \multirowcell{2}{\first{.250}} & \added{13.2} & \multirowcell{2}{\first{585}}
    & \multirowcell{2}{\first{33.37}} & \multirowcell{2}{\first{.968}} & \multirowcell{2}{\first{.032}} & \added{3.12} & \multirowcell{2}{\second{126}} \\
& \added{zstd} & & & & & \first{9.36} & & & & & \first{21.3} & & & & & \first{11.8} & & & & & \first{2.82} & \\

\bottomrule
\end{tabular}
    \caption{
        \label{tab:quantitative_results}
        Quantitative comparison of our proposed codec with the baseline (uncompressed starting scene) and other compression methods. The \colorbox{lightred}{first}, \colorbox{lightorange}{second}, and \colorbox{lightyellow}{third} best compression results are highlighted for each metric across different datasets. The 'FT' column indicates whether a method utilizes fine-tuning. All fine-tuning methods use 1,000 training iterations. \added{The 'Lossless Format' column indicates the lossless compression method used: NumPy's (npz) or Meta's Zstandard (zstd).}
    }
\end{table*}

\subsection{Fine-tuning}
\label{section:fine-tuning}

\begin{figure}
    \centering
\begin{tikzpicture}[
    node distance = 6mm and 16mm,
    start chain = A going below,
    base/.style = {draw, minimum width=32mm, minimum height=8mm, align=center, on chain=A},
    start/.style = {base, rectangle, rounded corners, fill=red!30},
    end/.style = {base, rectangle, rounded corners, fill=green!30},
    process/.style = {base, rectangle, fill=orange!30},
    decision/.style = {base, diamond, fill=blue!20, aspect=1.3, inner sep=0pt, minimum width=30mm, minimum height=17mm},
    every edge quotes/.style = {auto=right}
]
\node [start] (start) {Uncompressed model};
\node [process, yshift=-2mm] (enc) {Encode model\\(using POTR)};
\node [decision] (check) {$R_{\text{FT}} > 0$};
\node [end] (end) {Compressed model};

\node [process, right=12mm of check] (decode) {Decode model\\(using POTR)};
\node [process, above=of decode] (ft) {Fine-tune};
\node [process, above=of ft] (update_r) {$R_{\text{FT}} \gets R_{\text{FT}} - 1$};
\node[inner sep=0, right=of check, xshift=-15mm, yshift=2.5mm] {Yes};

\draw [arrows=-latex] 
    (start) edge[] (enc)
    (enc) edge[] (check)
    (check) edge["No"] (end)
    (check.east) edge[] (decode.west)
    (decode) edge[] (ft)
    (ft) edge[] (update_r)
    (update_r.west) -| (enc.north);
    ;
    
\end{tikzpicture}    
    \caption{
        \label{fig:ft_encoder_design}
        Schematic representation of the POTR-FT encoder. \\ \added{$R_{\text{FT}}$ denotes the remaining number of fine-tuning cycles}.
    }
\end{figure}
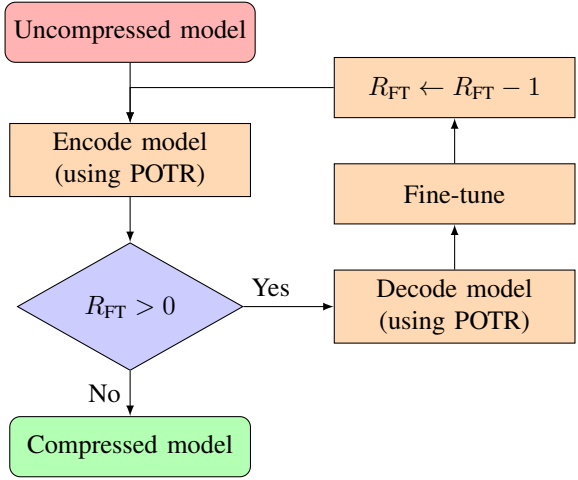

To enable a fair comparison of our proposed codec with others that incorporate fine-tuning, we introduce a variant of POTR, called POTR-FT, which includes a fine-tuning mechanism. Since fine-tuning is not the focus of this work, we deliberately keep POTR-FT's design simple. Specifically, its encoder employs the original (fine-tuneless) encoder and decoder as an internal black-box component, \added{performing $R_{\text{FT}}$ cycles of decode $\rightarrow$ fine-tune $\rightarrow$ encode, following the initial encoding.} The fine-tuning step trains the model using the original 3DGS training routine. The encoder design of POTR-FT is depicted in \Cref{fig:ft_encoder_design}.

\section{Experiments and discussion}
\label{section:results}
We present and discuss our codec's performance using the methodology laid out by 3DGS.zip~\cite{3DGSzip2024}, a 3DGS compression survey focusing on low distortion. If 3DGS.zip's methodology provides no guidance, we follow the initial 3DGS implementation's~\cite{kerbl3Dgaussians} approach where possible.

The remainder of this section is structured as follows. First, we discuss our experimental settings in \Cref{section:configuration}. Afterward, we present POTR's quantitative and qualitative results in \Cref{section:quant_and_qual_results}. In \Cref{section:culling_results} and \Cref{section:sh_results} we take a closer look at our proposed \changed{culling}{pruning} and spherical harmonics energy compaction approach respectively. \added{In \Cref{section:ablation} we present two ablation studies, and in \Cref{section:speed} we discuss our codec’s speed.}

\subsection{Experimental settings}
\label{section:configuration}
\subsubsection{Datasets}
We evaluate our codec across four datasets: three COLMAP~\cite{schoenberger2016sfm} datasets (Mip-NeRF 360~\cite{barron2022mipnerf360}, Deep Blending~\cite{hedman2018deep}, and Tanks And Temples~\cite{Knapitsch2017}) and one Blender dataset (NeRF-Synthetic~\cite{mildenhall2020nerf}). For the COLMAP datasets, every 8th image is designated for testing, while the Blender dataset comes with a predefined train-test split. The additional Mip-NeRF 360 scenes (\textit{flowers} and \textit{treehill}) are included and outdoor scenes (\textit{bicycle}, \textit{flowers}, \textit{stump}, \textit{treehill}, \textit{garden}) are downscaled 4× while indoor scenes (\textit{counter}, \textit{kitchen}, \textit{room}, \textit{bonsai}) are downscaled 2x.

\subsubsection{Models}
As POTR is a \emph{post}-training compression codec, a trained model is required to use our proposed codec. We use the models published alongside the initial 3DGS implementation\footnote{\url{https://repo-sam.inria.fr/fungraph/3d-gaussian-splatting/datasets/pretrained/models.zip}} where possible, however, NeRF-Synthetic models are absent from this collection. We train these models ourselves using the configuration provided by 3DGS's initial implementation.

\subsubsection{Hyperparameters}
\begin{table}
\centering
\begin{tabular}{l l|l l}
\hline
Parameter & Value & Parameter & Value \\
\hline
$\lambda$ & $10^{-q}$ & $\texttt{SF}_\text{SH}$ & $1 + 100q$ \\
$\alpha$ & $\operatorname{sigmoid}(3q)$ & $\texttt{SF}_\text{opacity}$ & $1 + 200 q$ \\
$\triangle{\text{MSE}_{\texttt{MAX}}}$ & $10^{-8.8 - 2q}$ & $\texttt{SF}_\text{rotation}$ & $1 + 400 q$ \\
$\beta$ & $1.4 \cdot 10^{-4} \cdot q$ & $\texttt{SF}_\text{scale}$ & $1 + 4000 q$ \\
$\gamma$ & $5.0 \cdot 10^{-3-5q}$ & \# pruning iterations & 48 \\
\hline
\end{tabular}

\caption{
\label{tab:hyperparameters}
Hyperparameters as functions of $q \in [0,1]$.}
\end{table}
All hyperparameters are governed by a single quality parameter $q$ \added{as specified by \Cref{tab:hyperparameters}. We set $q$ to $0.5$ in spirit, but to match the results of other works, we alter it slightly per dataset} such that POTR's objective quality slightly exceeds those of other compression works. In comparison, MesonGS~\cite{xie2024mesongs} sets their hyperparameters on a per-scene basis. This yields further non-negligible RD gains but is currently done manually or at great computational cost. Finally, for POTR-FT, we use two \added{fine-tuning cycles ($R_{\text{FT}}=2$)} of 500 training iterations each. To ensure fairness, comparisons to other fine-tuning methods match the total number of training iterations.

\subsubsection{Metrics}
Our evaluation focuses predominately on RD performance, with visual quality being objectively assessed using PSNR, SSIM~\cite{zhou2004ssim}, and LPIPS~\cite{zhang2018perceptual}. \added{We also report inference time, though it is inherently hardware-dependent. For a hardware-agnostic indicator of computational cost, we additionally consider the number of splats.}

\subsection{Codec results}
\label{section:quant_and_qual_results}

\newcommand{\zoominTruck}[3]{ %
\begin{tikzpicture}[spy using outlines={rectangle,magnification=4,size=1cm}] 
    \node[anchor=south west,inner sep=0]{\includegraphics[width=\mytmplen]{#3}};
    \spy[draw=lime] on (.63\mytmplen, .51\mytmplen) in node at (.88\mytmplen,0\mytmplen);
    \node[anchor=west] at (0.01\mytmplen,-.06\mytmplen) {
        #1
    };
    \node[anchor=east] at (0.76\mytmplen,-.06\mytmplen) {
        #2
    };
\end{tikzpicture}
}

\newcommand{\zoominGarden}[3]{ %
\begin{tikzpicture}[spy using outlines={rectangle,magnification=4,size=1cm}] 
    \node[anchor=south west,inner sep=0]{\includegraphics[width=\mytmplen,trim=0cm 2.1cm 0cm 2cm,clip]{#3}};
    \spy[draw=lime] on (.515\mytmplen, .33\mytmplen) in node at (.88\mytmplen,0\mytmplen);
    \node[anchor=west] at (0.01\mytmplen,-.06\mytmplen) {
        #1
    };
    \node[anchor=east] at (0.76\mytmplen,-.06\mytmplen) {
        #2
    };
\end{tikzpicture}
}

\newcommand{\zoominPlayroom}[3]{ %
\begin{tikzpicture}[spy using outlines={rectangle,magnification=4,size=1cm}] 
    \node[anchor=south west,inner sep=0]{\includegraphics[width=\mytmplen,trim=0cm 2.2cm 0cm 2.25cm,clip]{#3}};
    \spy[draw=lime] on (.72\mytmplen, .46\mytmplen) in node at (.88\mytmplen,0\mytmplen);
    \node[anchor=west] at (0.01\mytmplen,-.06\mytmplen) {
        #1
    };
    \node[anchor=east] at (0.76\mytmplen,-.06\mytmplen) {
        #2
    };
\end{tikzpicture}
}

\newcommand{\zoominLego}[3]{ %
\begin{tikzpicture}[spy using outlines={rectangle,blue,magnification=4,size=1cm}] 
    \node[anchor=south west,inner sep=0]{\includegraphics[width=\mytmplen,trim=0cm 8.6cm 0cm 3.85cm,clip]{#3}};
    \spy[draw=lime] on (.48\mytmplen, .12\mytmplen) in node at (.88\mytmplen,0\mytmplen);
    \node[anchor=west] at (0.01\mytmplen,-.06\mytmplen) {
        #1
    };
    \node[anchor=east] at (0.76\mytmplen,-.06\mytmplen) {
        #2
    };
\end{tikzpicture}
}

\begin{figure*}[!p]
    \newlength\mytmplen
    \setlength\mytmplen{.23\linewidth}
	\centering
\begin{tabular}{c@{\hskip 0.1cm}|@{\hskip 0.1cm}c@{\hskip 0.2cm}c@{\hskip -0.11cm}c@{\hskip -0.11cm}c@{\hskip -0.11cm}p{4cm}}

\multicolumn{2}{c}{\raisebox{0.25\height}{\rotatebox{90}{Ground-truth}}} &
\zoominTruck{Size}{\# of splats}{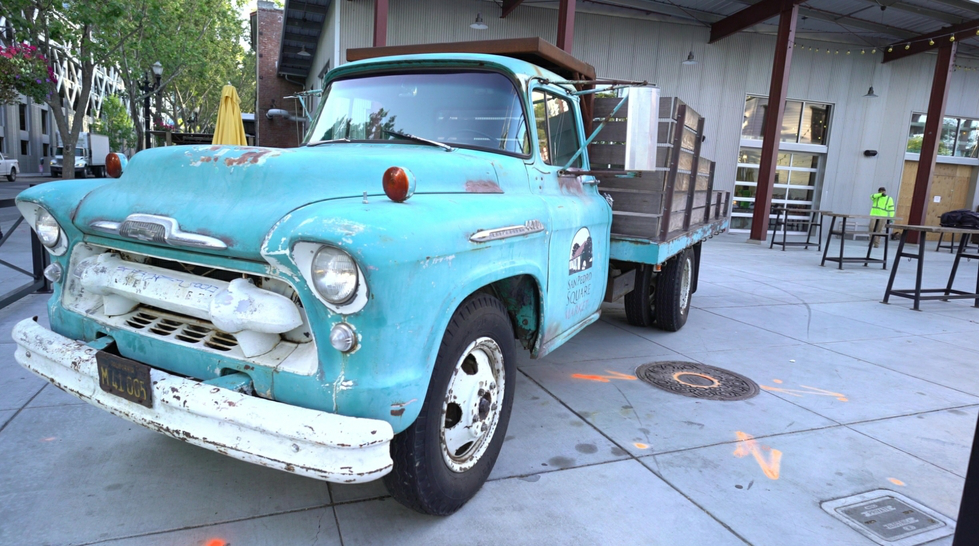} &
\zoominGarden{Size}{\# of splats}{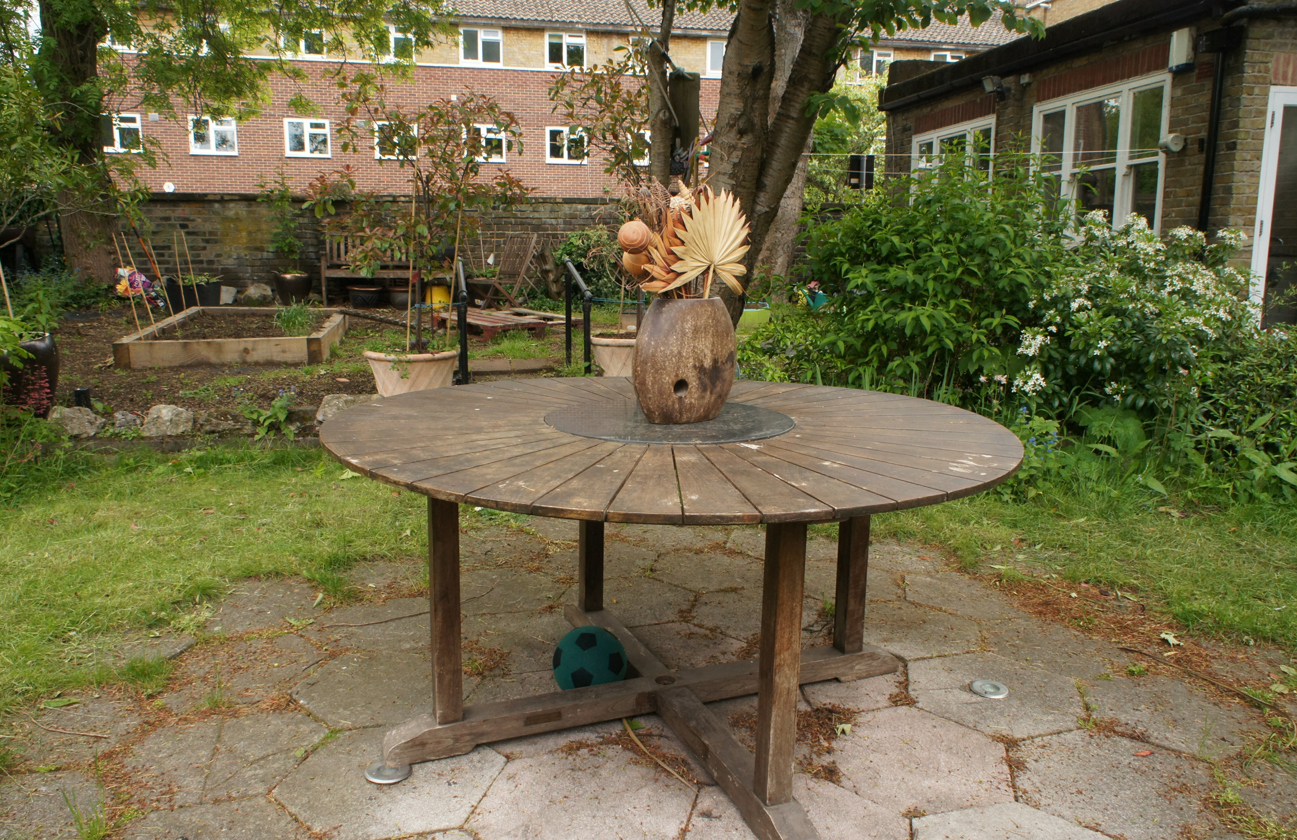} &
\zoominPlayroom{Size}{\# of splats}{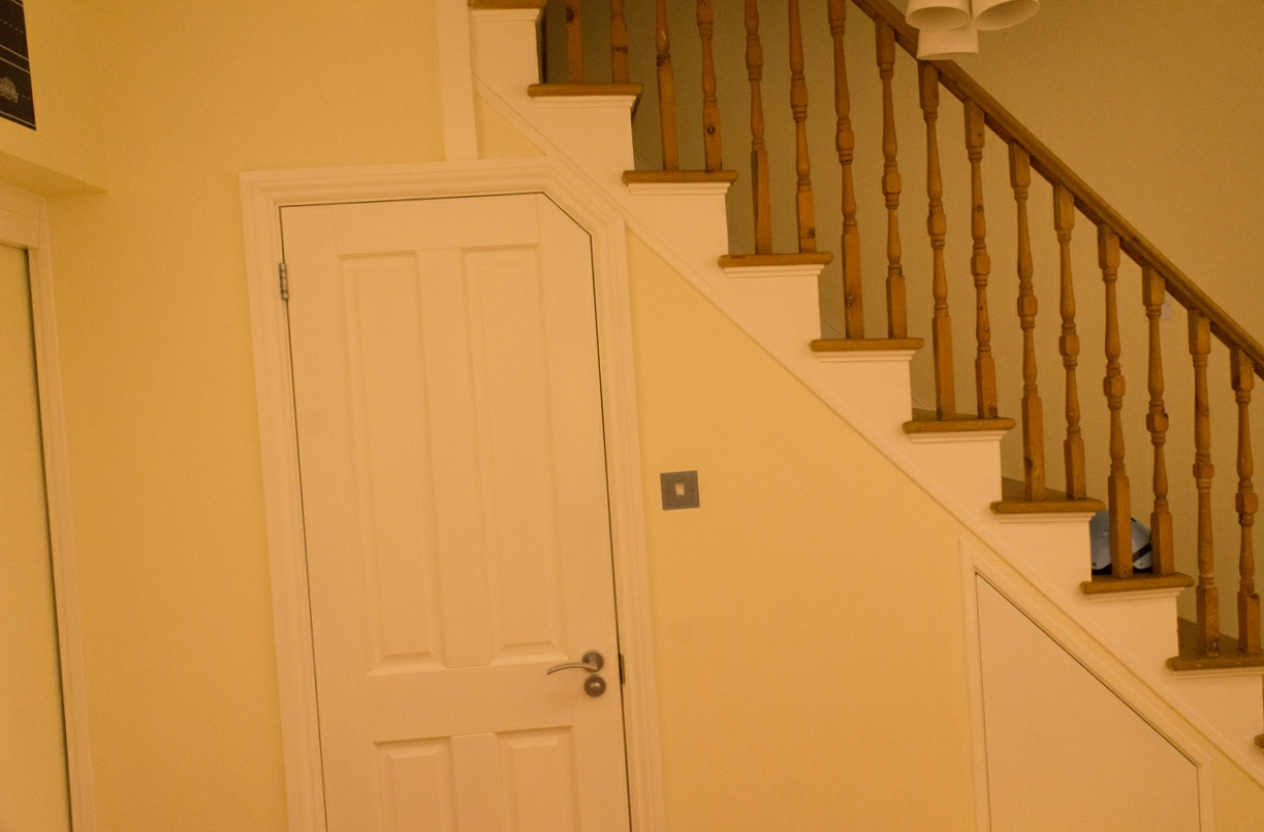} &
\zoominLego{Size}{\# of splats}{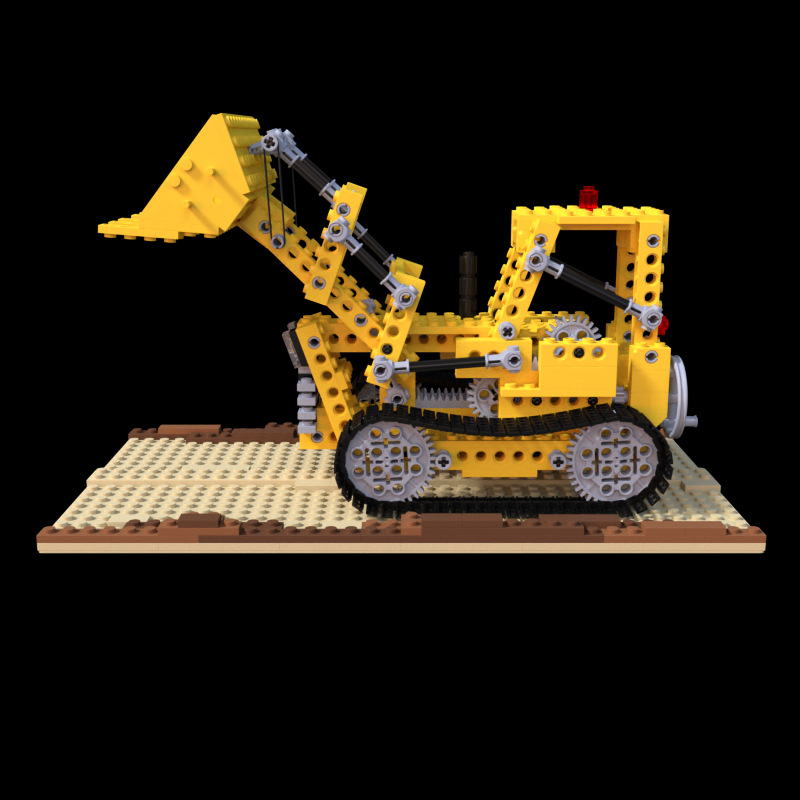} \\

\multicolumn{2}{c}{\raisebox{0.7\height}{\rotatebox{90}{Baseline}}} &
\zoominTruck{630 MB}{2.54 M}{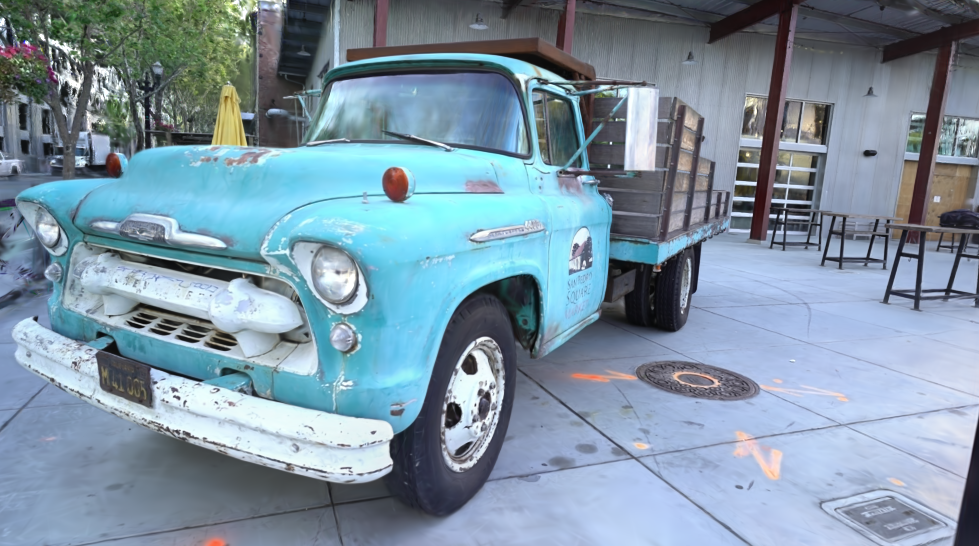} &
\zoominGarden{1.45 GB}{5.83 M}{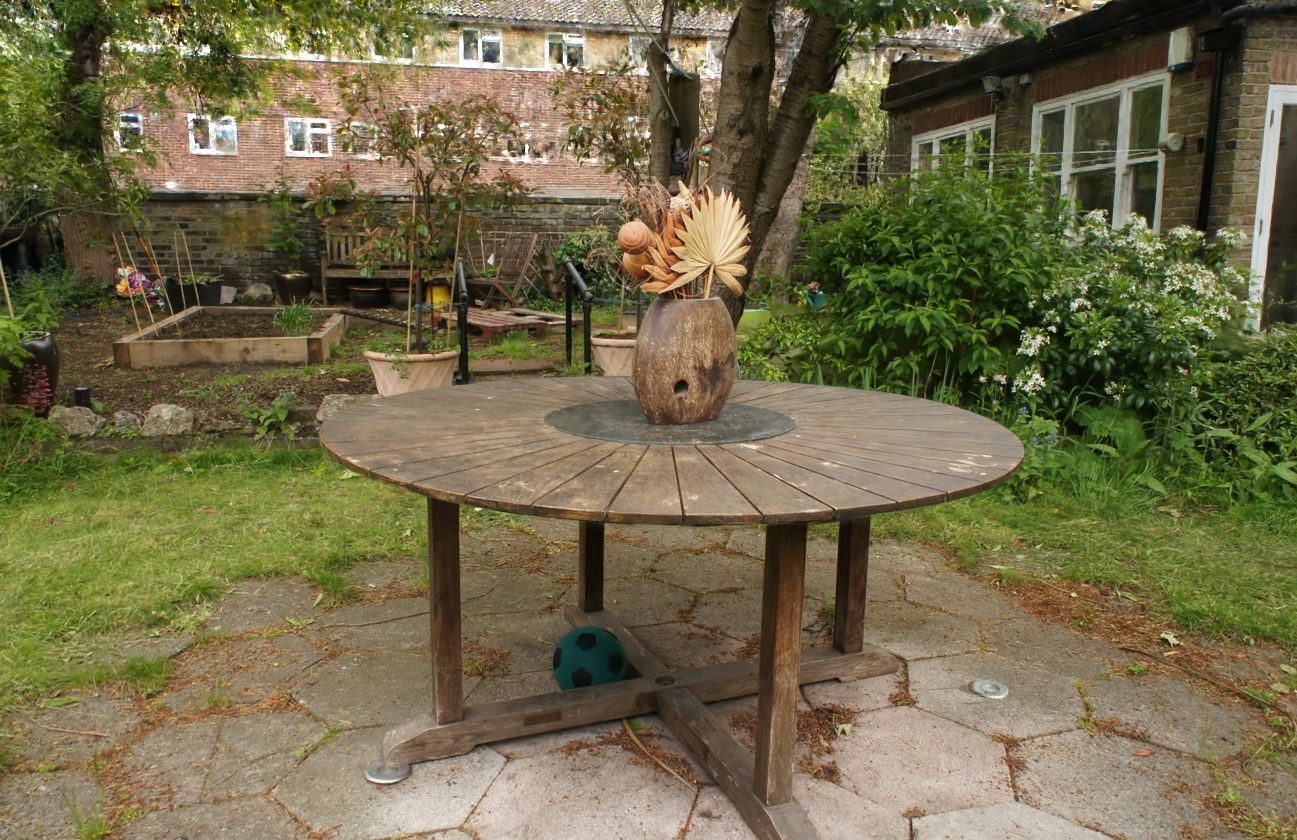} &
\zoominPlayroom{631 MB}{2.55 M}{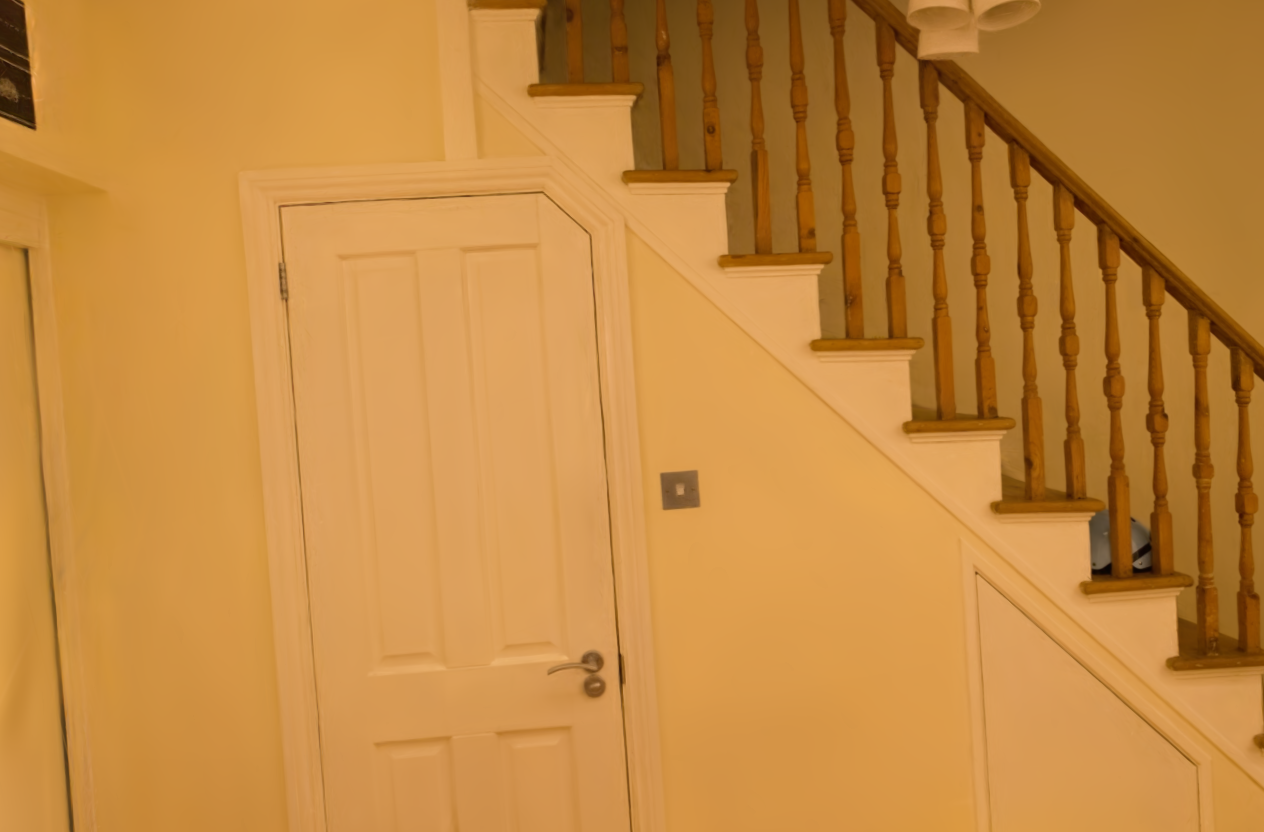} &
\zoominLego{85.0 MB}{343 k}{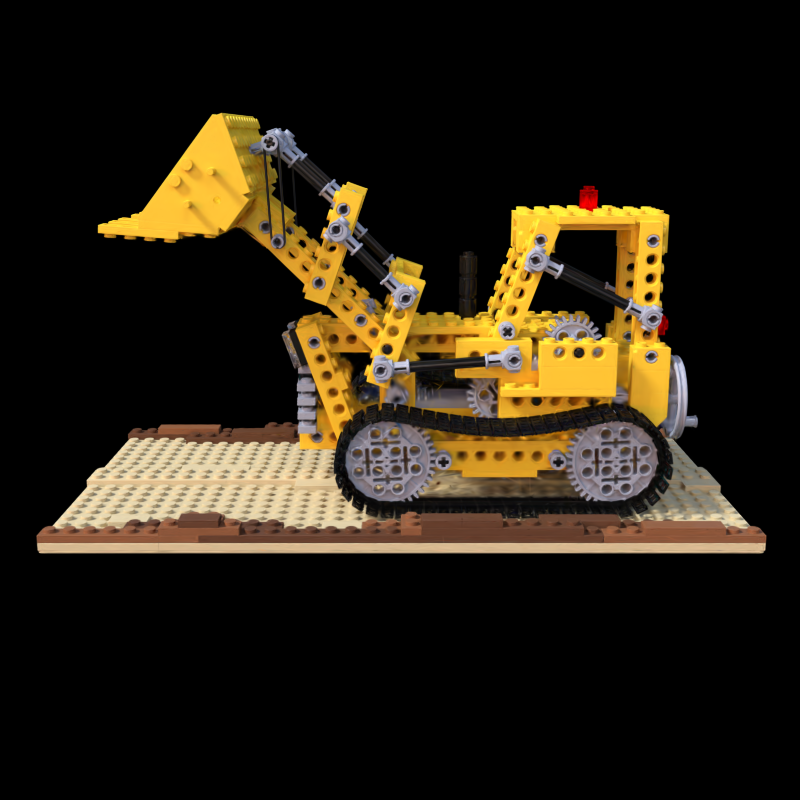} \\

\cline{1-6} \vspace{-.35cm} \\
\multirow{2}{*}{\parbox[c][1em][c]{\ht\strutbox}{\centering\rotatebox{90}{No fine-tuning}}} & 
\raisebox{0.27\height}{\rotatebox{90}{POTR (ours)}} &
\zoominTruck{12.5 MB}{822 k}{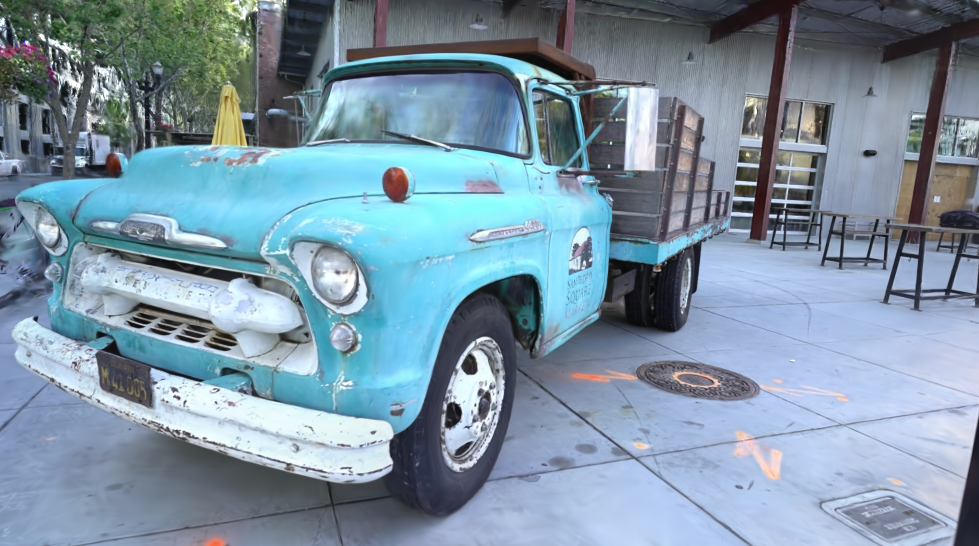} &
\zoominGarden{40.7 MB}{2.36 M}{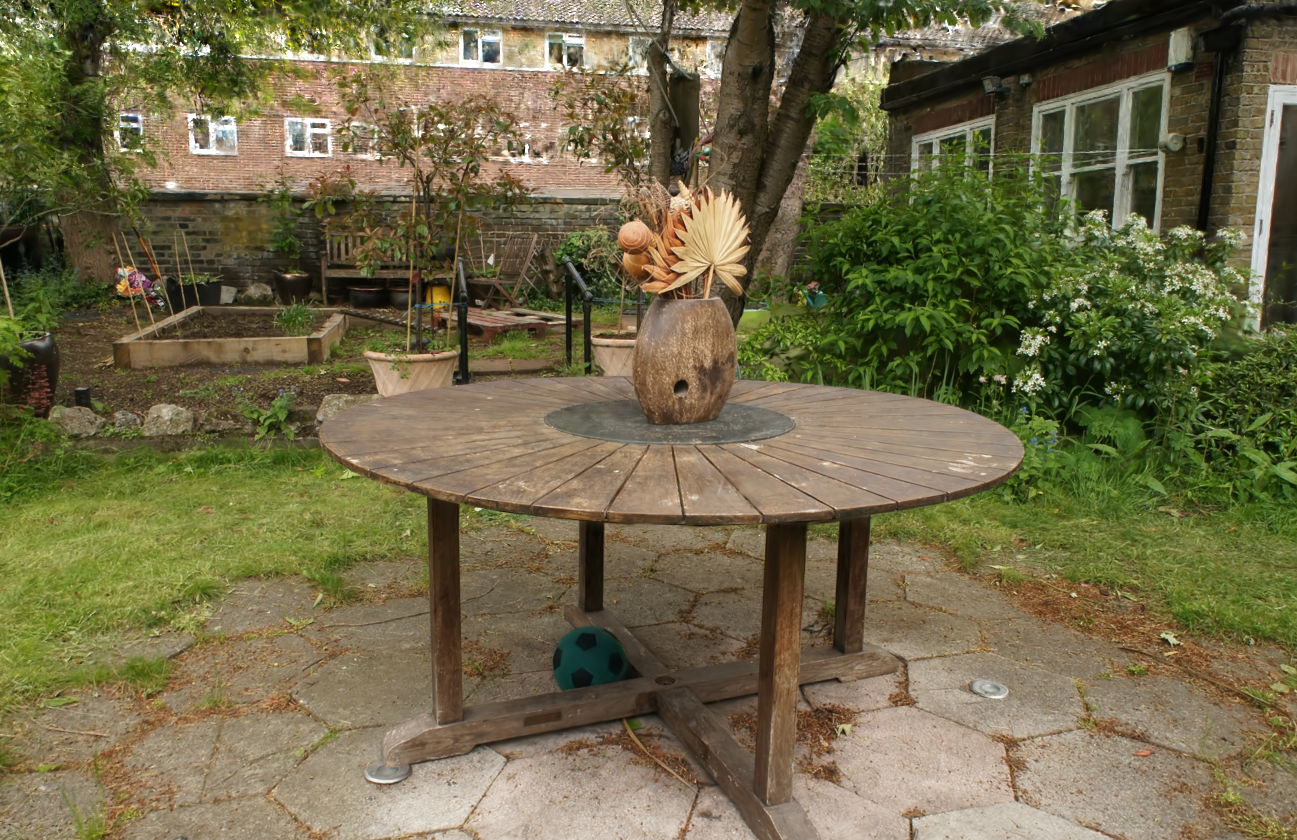} &
\zoominPlayroom{13.7 MB}{635 k}{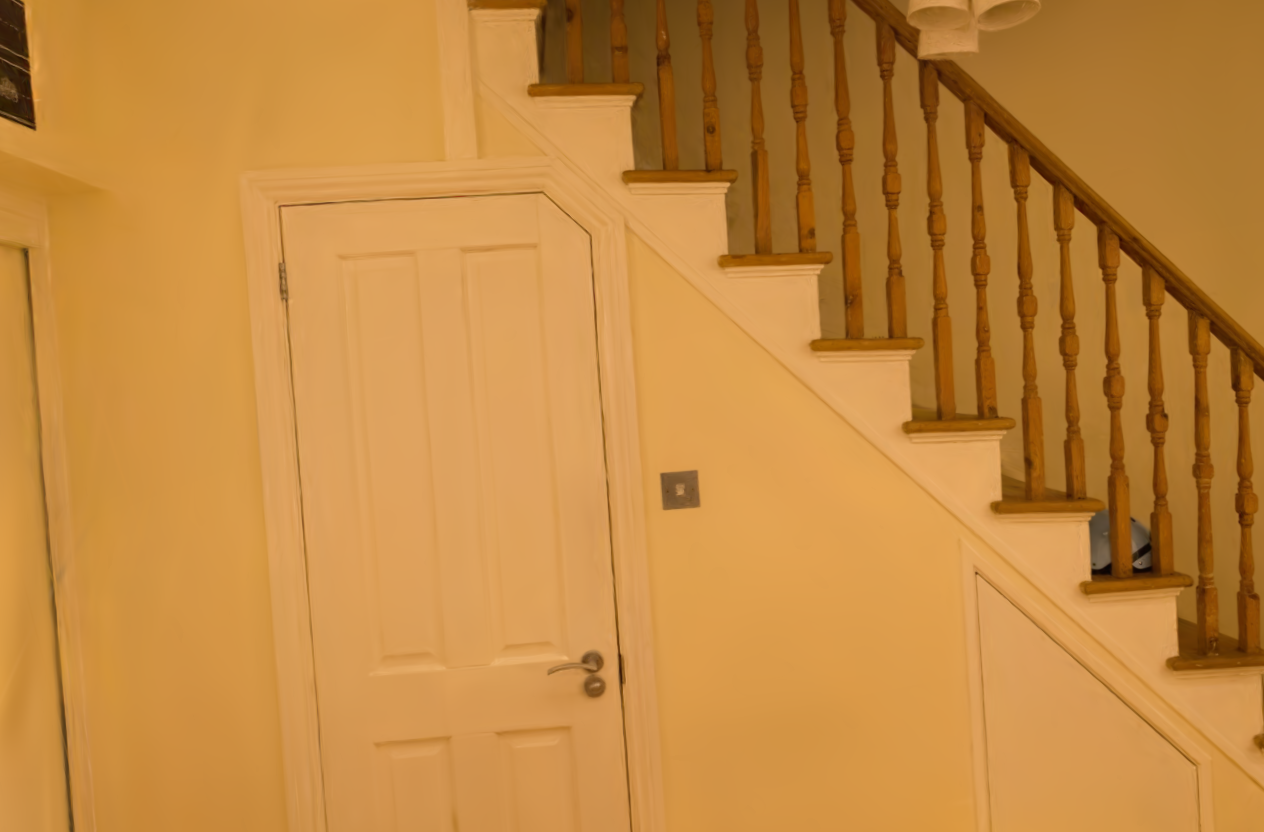} &
\zoominLego{4.00 MB}{192 k}{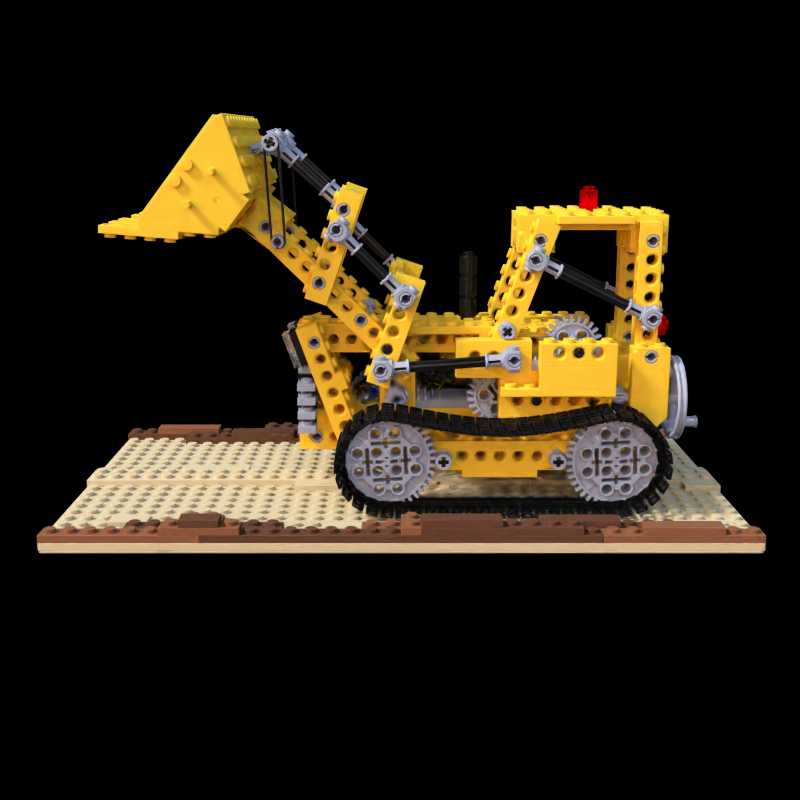} \\

&
\raisebox{0.55\height}{\rotatebox{90}{MesonGS}} &
\zoominTruck{22.5 MB}{1.52 M}{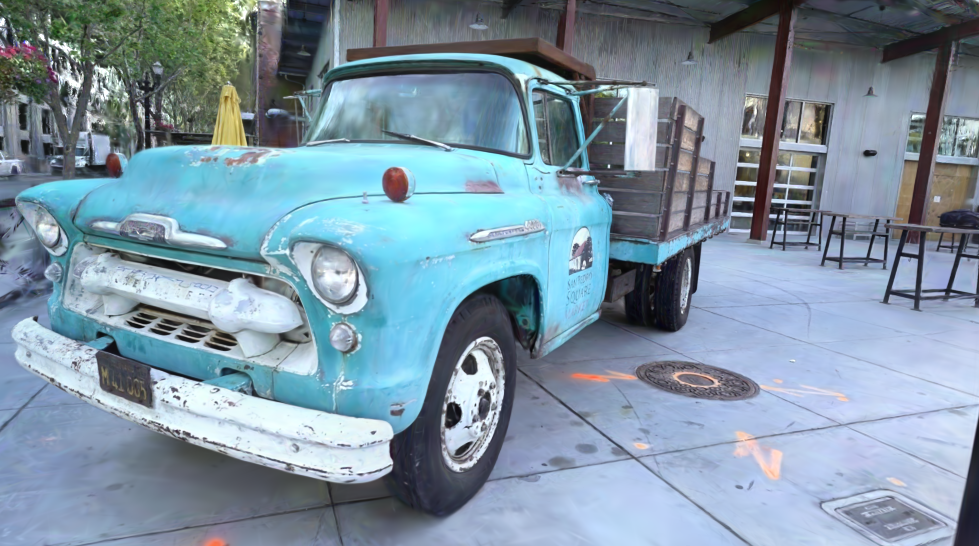} &
\zoominGarden{57.6 MB}{4.20 M}{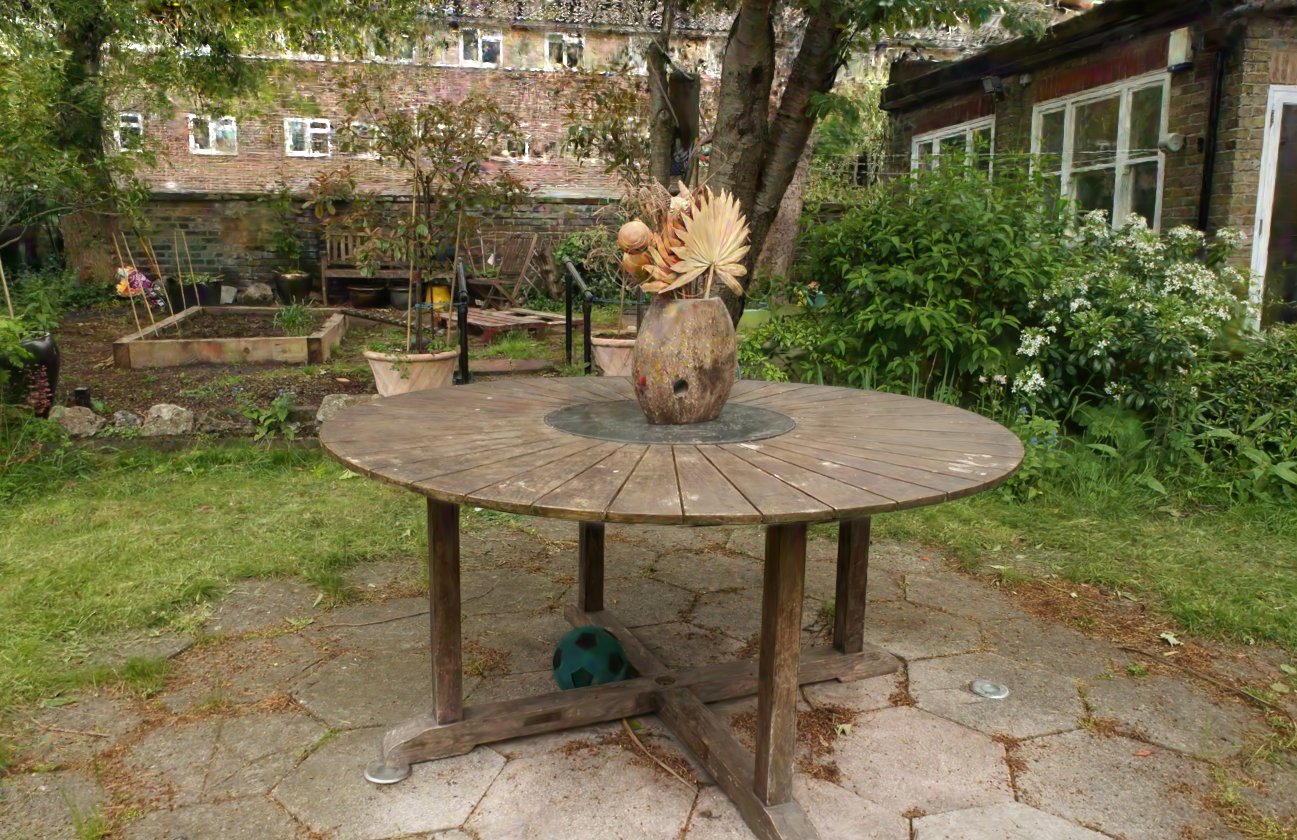} &
\zoominPlayroom{29.0 MB}{2.04 M}{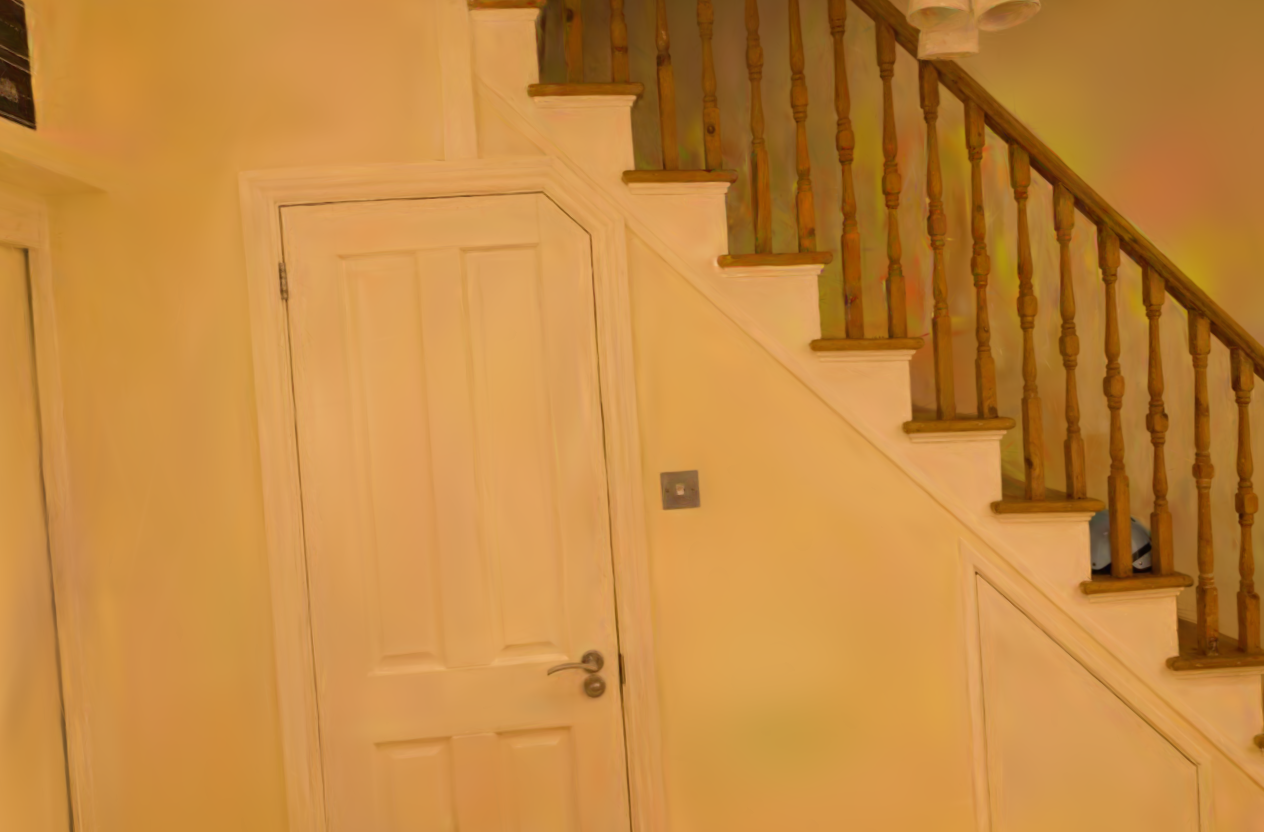} &
\zoominLego{4.42 MB}{274 k}{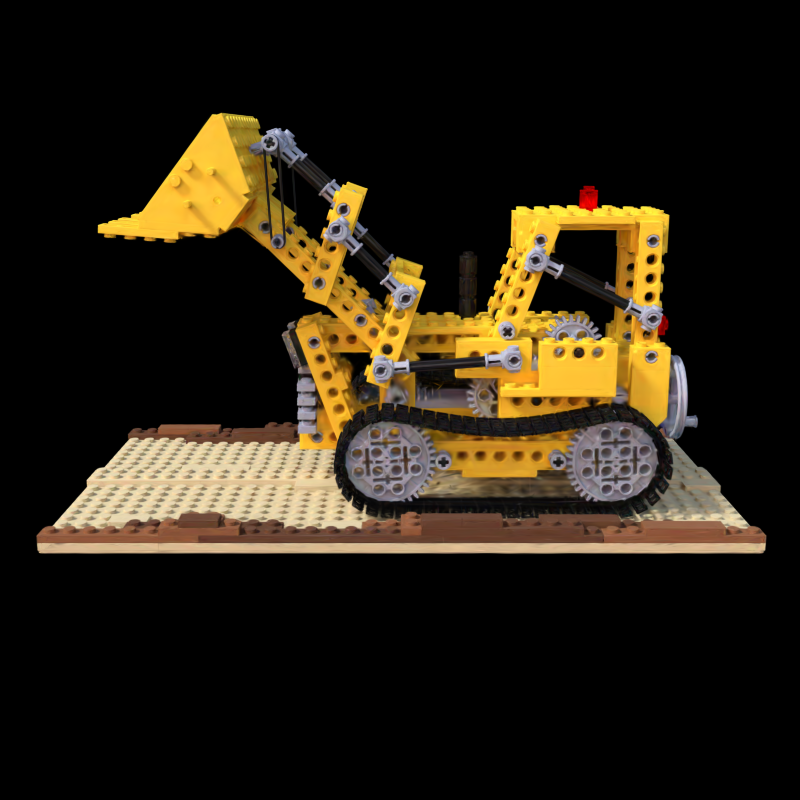} \\

\cline{1-6} \vspace{-.35cm} \\
\multirow{3}{*}{\raisebox{-1.43\height}{\rotatebox{90}{Fine-tuning}}} &
\raisebox{0.08\height}{\rotatebox{90}{POTR-FT (ours)}} &
\zoominTruck{10.2 MB}{691 k}{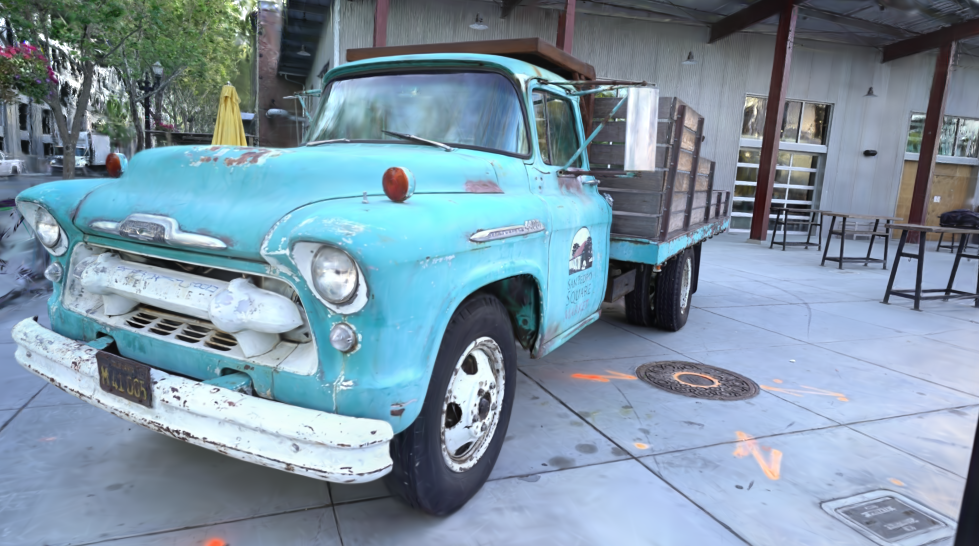} &
\zoominGarden{31.7 MB}{1.94 M}{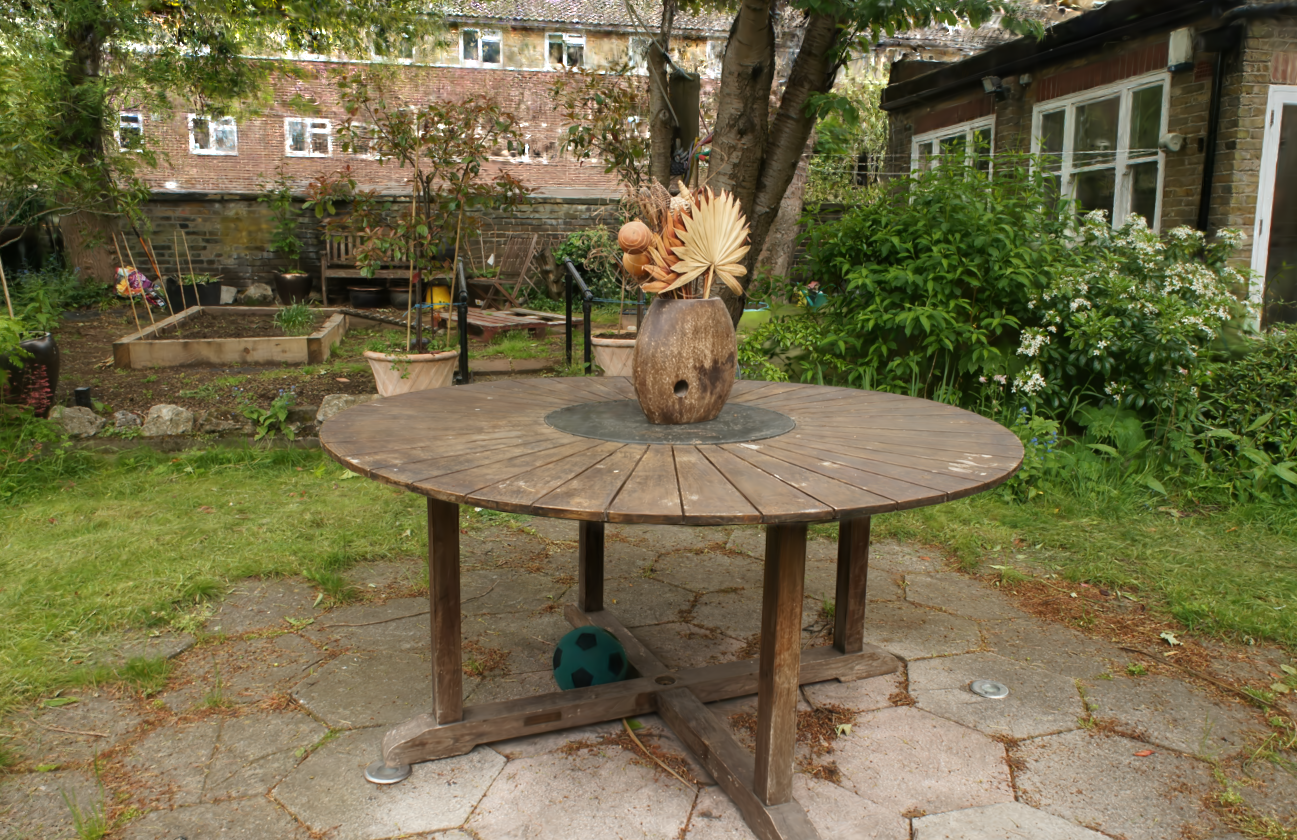} &
\zoominPlayroom{9.64 MB}{466 k}{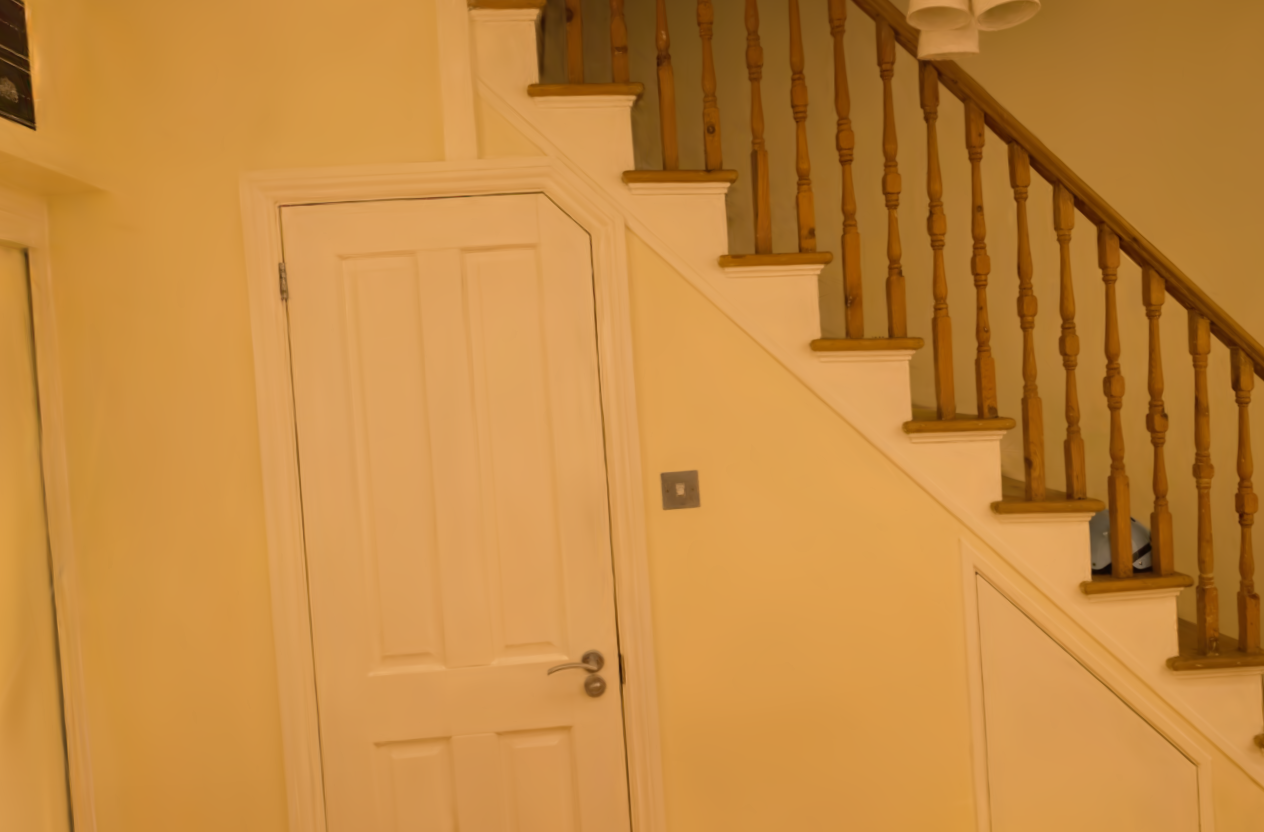} &
\zoominLego{3.22 MB}{161 k}{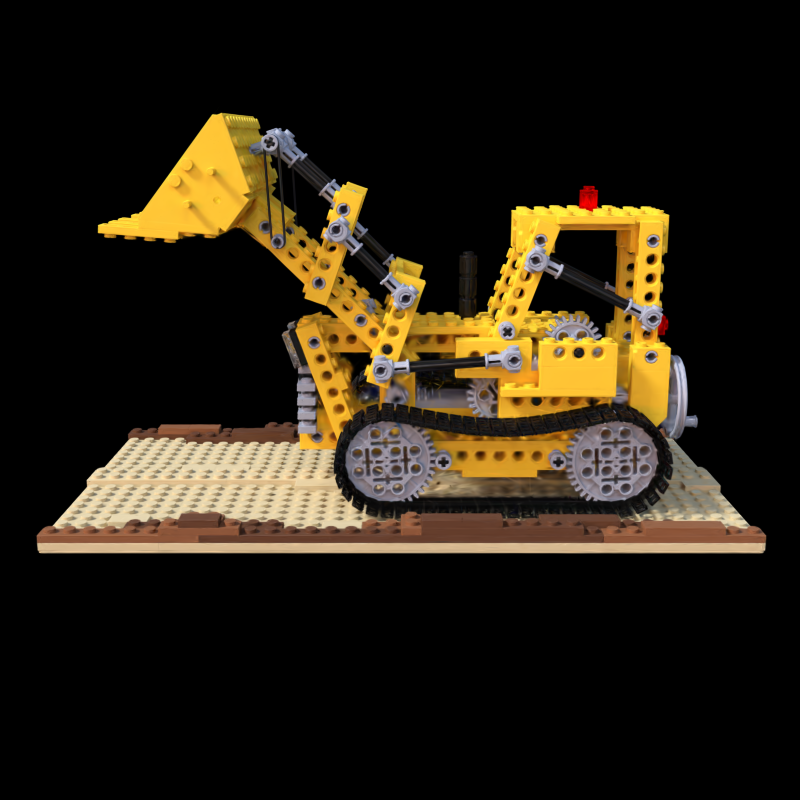} \\

&
\raisebox{0.25\height}{\rotatebox{90}{MesonGS-FT}} &
\zoominTruck{22.5 MB}{1.52 M}{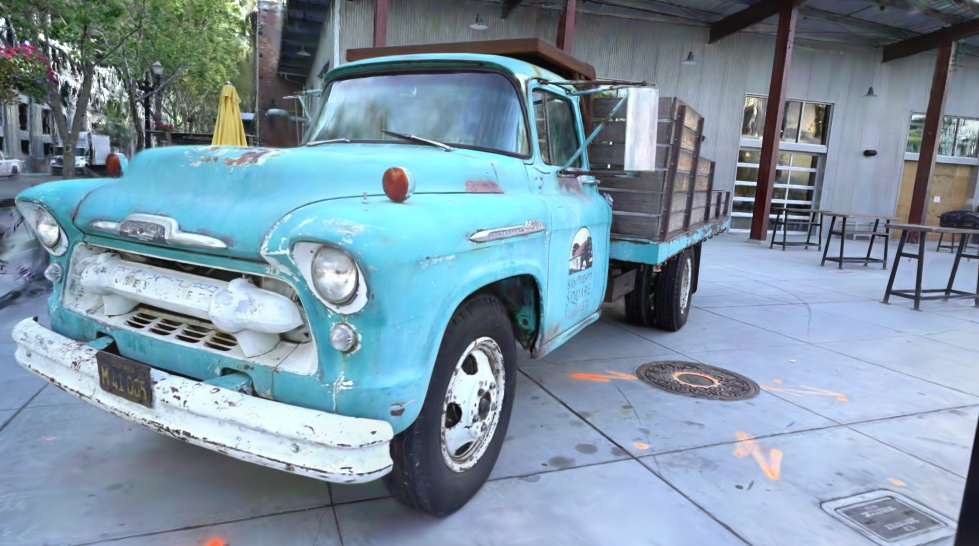} &
\zoominGarden{57.6 MB}{4.20 M}{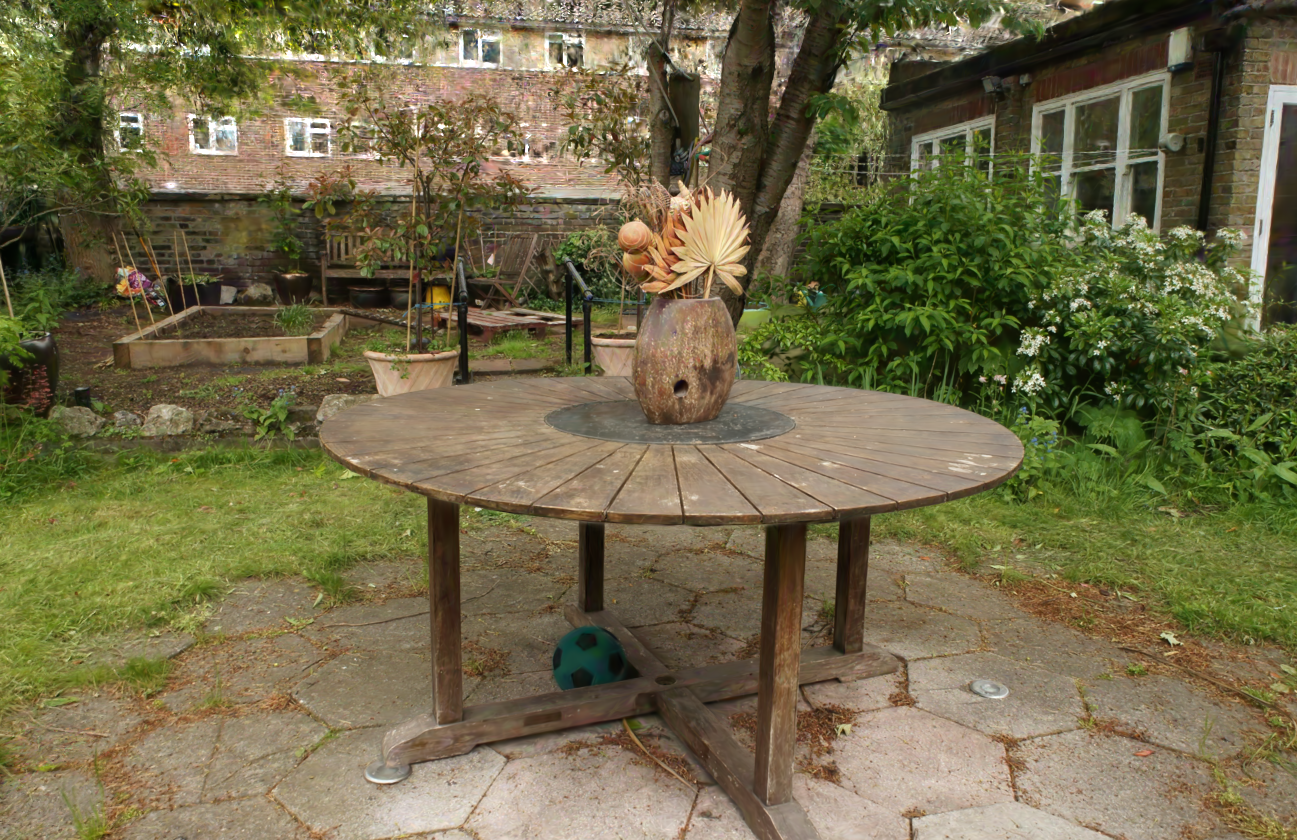} &
\zoominPlayroom{29.0 MB}{2.04 M}{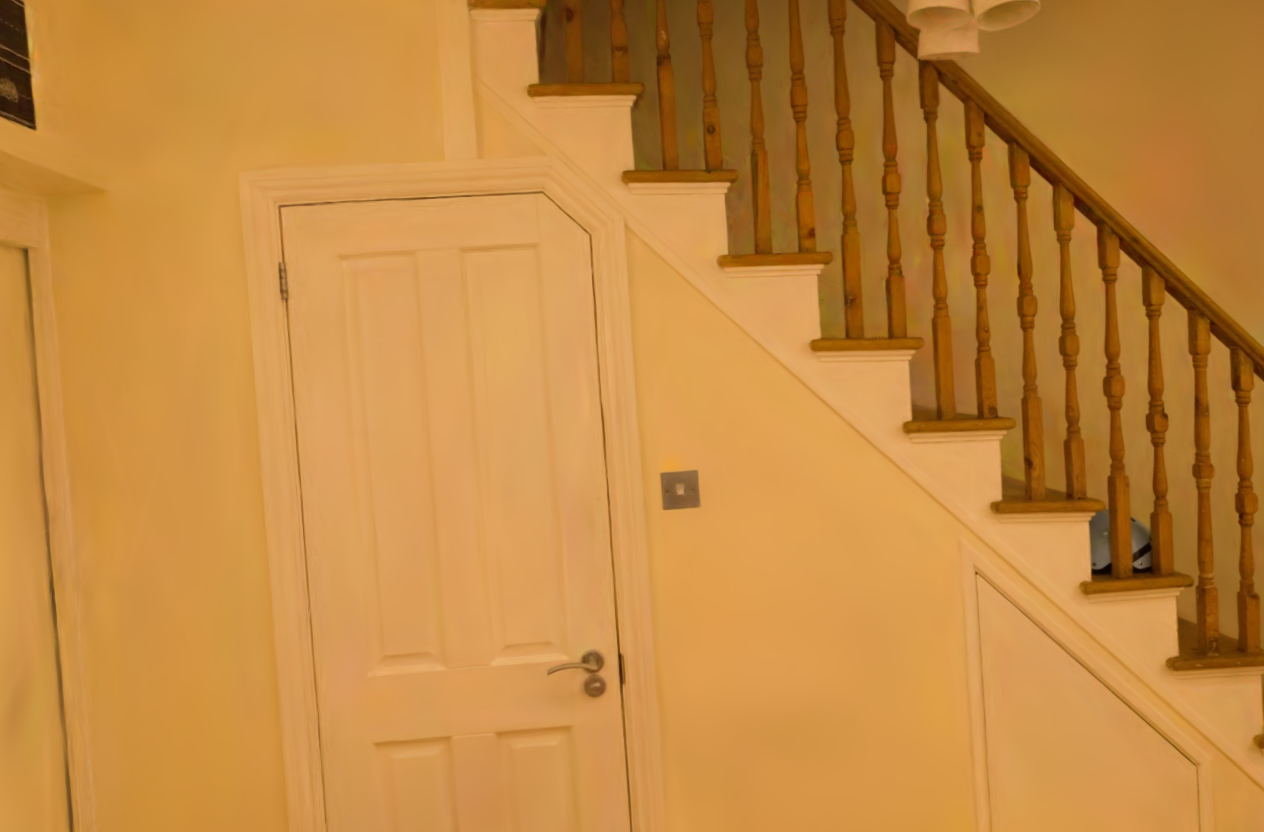} &
\zoominLego{4.42 MB}{274 k}{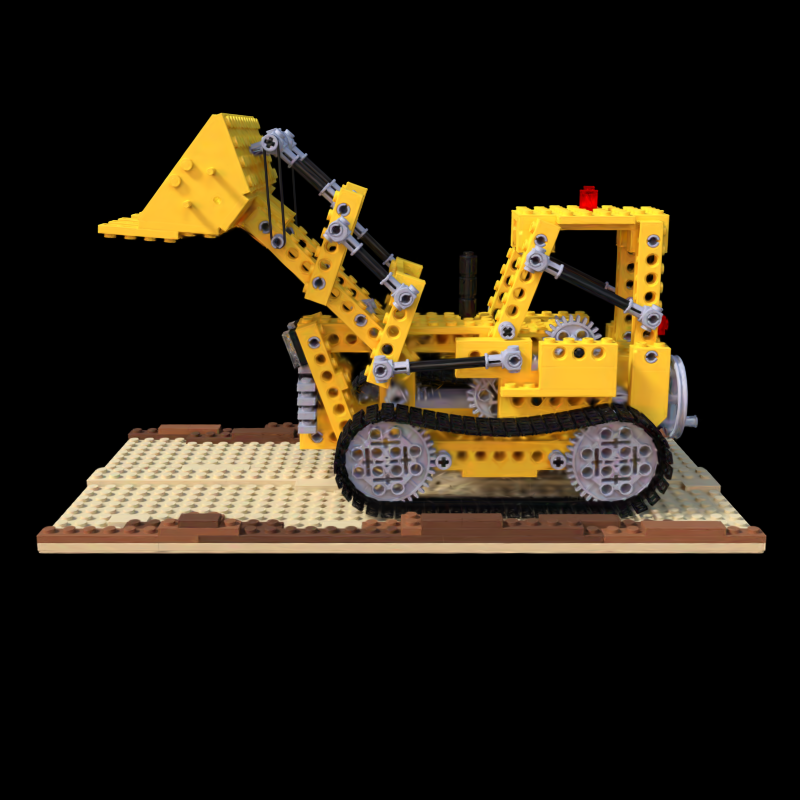} \\

&
\raisebox{0.8\height}{\rotatebox{90}{C3DGS}} &
\zoominTruck{22.4 MB}{2.05 M}{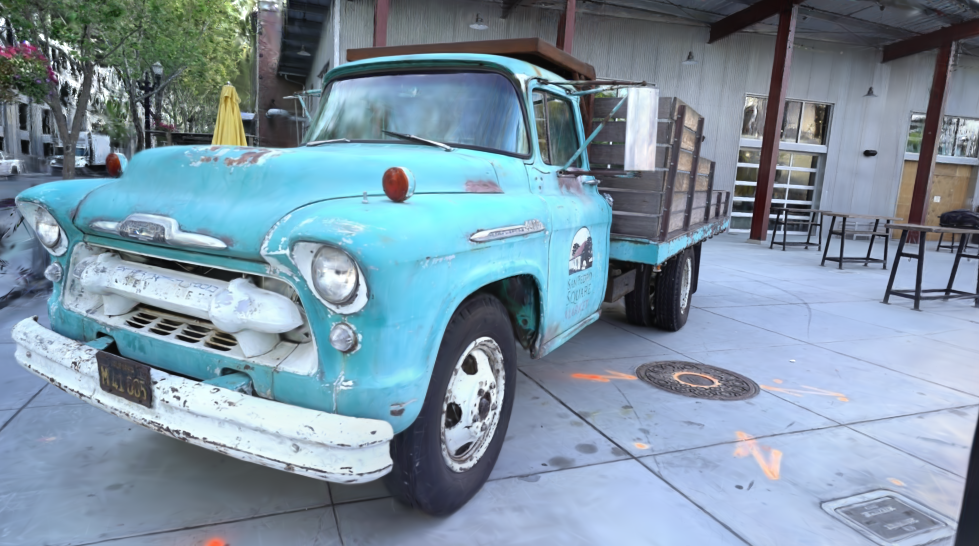} &
\zoominGarden{49.1 MB}{5.32 M}{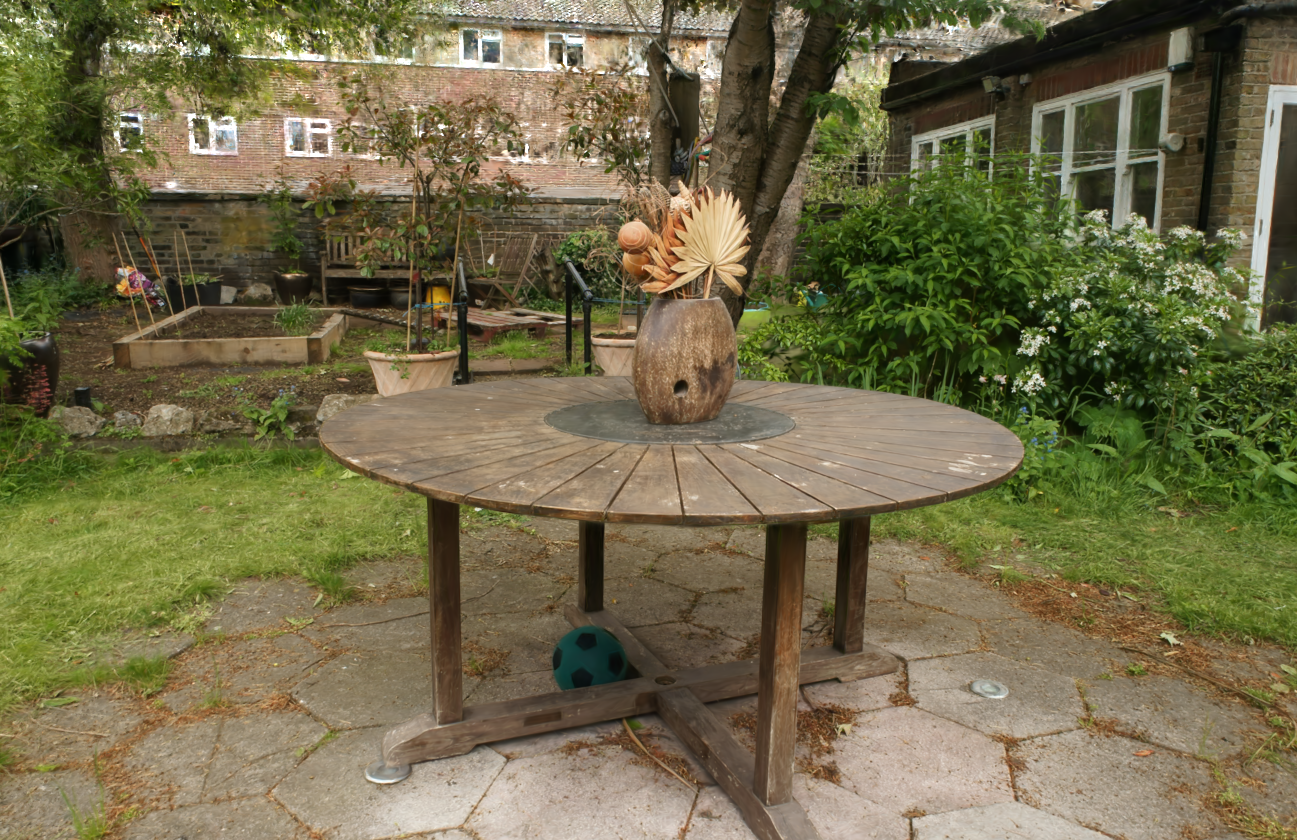} &
\zoominPlayroom{22.8 MB}{2.28 M}{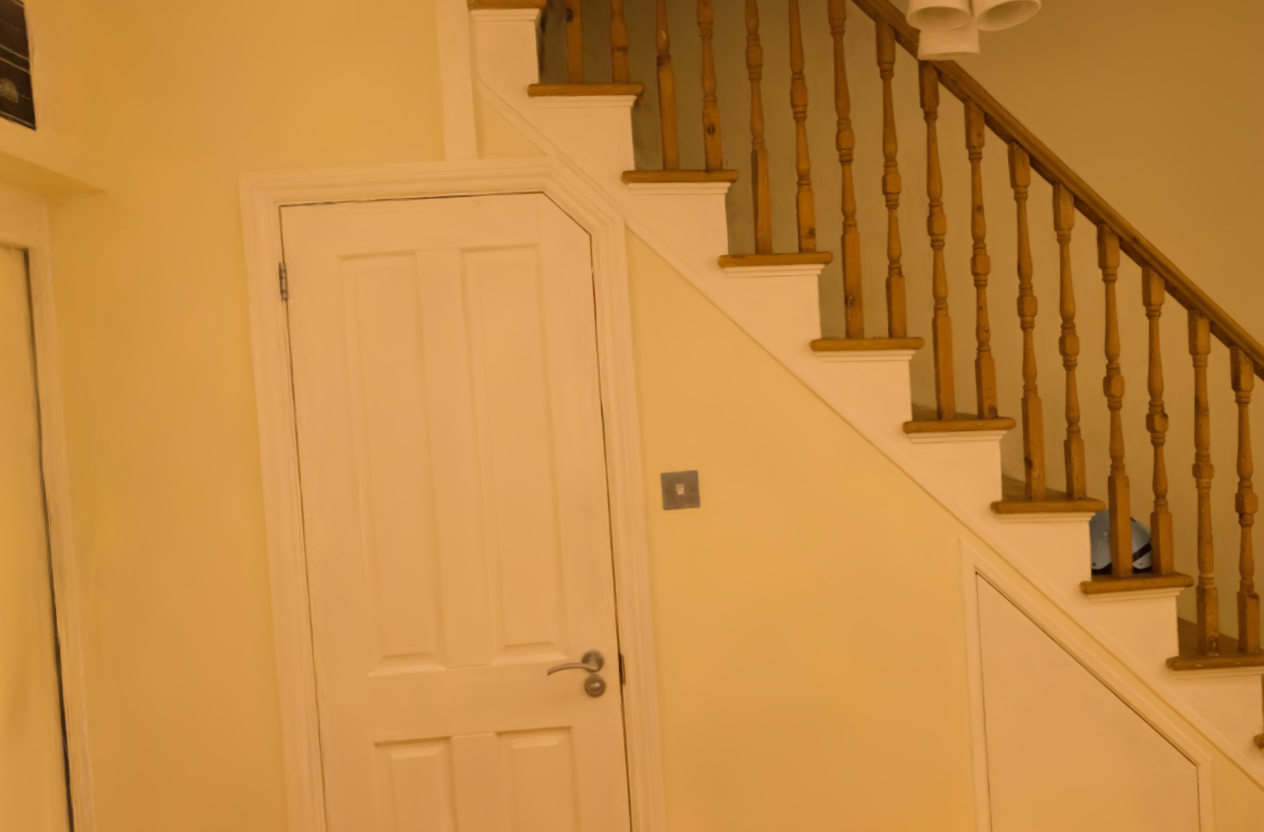} &
\zoominLego{4.99 MB}{333 k}{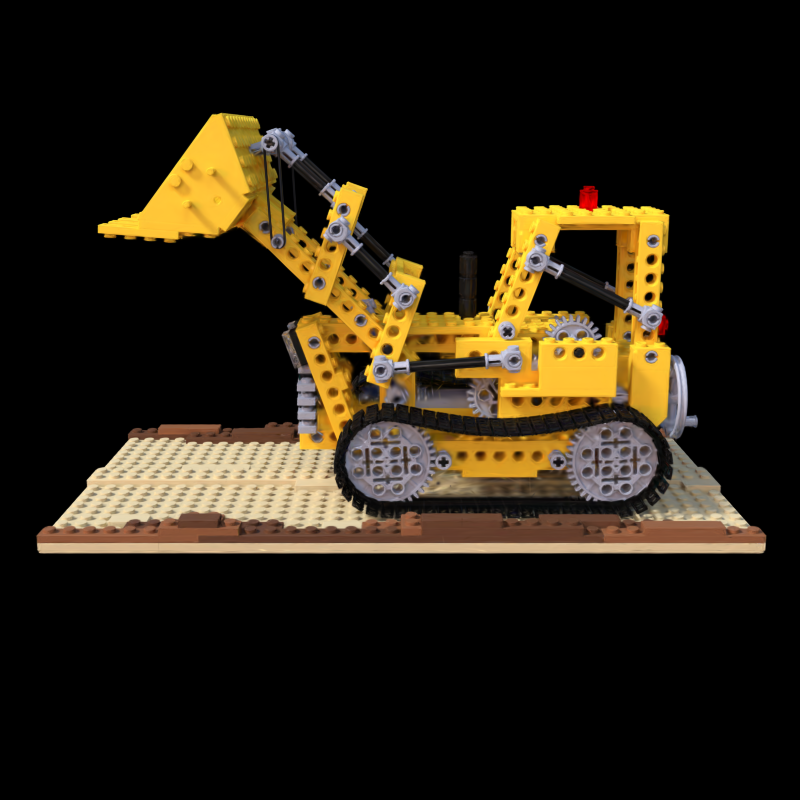} \\

\end{tabular}
    \caption{
        \label{fig:qualitative_results}
        Qualitative comparison of our proposed codec (POTR and POTR-FT) with the baseline (Kerbl et al.~\cite{kerbl3Dgaussians}) and the state-of-the-art in post-training 3DGS compression (MesonGS, MesonGS-FT~\cite{xie2024mesongs}, and C3DGS~\cite{Niedermayr_2024_CVPR}). One scene per discussed dataset is considered, namely (from left to right): Truck (Tanks And Temples), Garden (Mip-NeRF 360), Playroom (Deep Blending), and Lego (NeRF-Synthetic). Every model is annotated with its size and the number of splats.
    } 
\end{figure*}
\begin{figure}
    \centering
\pgfplotsset{
    every non boxed x axis/.style={}
}

\begin{tikzpicture}
\begin{groupplot}[
    group style={
        group size=2 by 1,
        yticklabels at=edge left,
        horizontal sep=0pt,
    },
    ymin=0.235, ymax=0.305,
    ytick={0.22, 0.24, 0.26, 0.28, 0.30, 0.32},
    enlarge y limits=0.05,
    height=5cm,
    width=5cm,
    legend style={
        cells={anchor=west},
        font=\small,
        at={(1.2,1.0)},
        anchor=north east
    },
]

\nextgroupplot[
    ylabel={LPIPS},
    xlabel={Size (MB)},
    xmin=0, xmax=48.5,
    xtick={0, 5, 10, 20, 30, 40, 50},
    axis y line=left, 
    axis x line=bottom, 
    width=7.5cm 
]

\draw[Stealth-Stealth] (8, 0.272) -- (28, 0.272);
\node[fill=white, text opacity=1, fill opacity=1.0, inner sep=1pt] at (axis cs:018,0.2725) {\footnotesize 4x smaller};

\addplot[color=blue, mark=triangle] table {
3.490625977 0.295534
4.610375977 0.285231
5.827505371 0.278425
7.242029297 0.272844
8.761384277 0.268070
10.443530762 0.264017
12.081468262 0.260071
13.701521973 0.257115
15.387286133 0.254339
17.184550293 0.252367
19.085396484 0.250517
21.115932129 0.249021
23.118461914 0.247908
25.382962891 0.246809
27.620252441 0.245983
29.917812012 0.245227
32.373335449 0.244685
34.713791504 0.244125
37.158008301 0.243717
};
\addlegendentry{POTR (ours)}

\addplot[color=blue, mark=o] table {
2.134489	0.2842679471
2.9537355	0.2735414803
3.850236	0.2676080316
4.849885	0.26341784
5.9686375	0.2599120736
7.230873	0.2567249686
8.540295	0.2540059686
9.995836	0.2519812733
11.4839385	0.2499682978
13.048576	0.2487811893
14.7182645	0.247428447
16.598715	0.246245876
18.480405	0.2454592884
20.489686	0.2446639538
22.534792	0.2441694587
24.6450505	0.2436332554
27.1074725	0.2431647331
29.4776415	0.2427982688
31.616126	0.2424660847
};
\addlegendentry{POTR-FT (ours)}

\addplot[color=red, mark=triangle] table {
28.9528150 0.27136527
};
\node[overlay, right] at (axis cs:28.95,0.272) {\footnotesize MesonGS};

\addplot[color=red, mark=o] table {
28.9514825 0.25527981
};
\node[overlay, right] at (axis cs:28.9514825,0.25527981) {\footnotesize MesonGS-FT};

\addplot[color=red, mark=o] table {
26.718243500 0.255876288
};
\node[overlay, left] at (axis cs:26.718,0.256) {\footnotesize C3DGS};

\addplot[color=red, mark=o] table {
47.9374075 0.2612154
};
\node[overlay, left] at (axis cs:47.9374075,0.2612154) {\footnotesize LightGaussian};

\nextgroupplot[
    xmin=728, xmax=745,
    xtick={738}, 
    axis y line=none,
    axis x line=bottom, 
    axis x discontinuity=parallel,
    width=2.9cm 
]
\addplot[color=black, mark=square] table {
738.0 0.246
};
\node[overlay, above] at (axis cs:738.0,0.246) {\footnotesize Baseline};

\end{groupplot}
\end{tikzpicture}
    \caption{
        \label{fig:rd_curve}
        RD performance comparison of various post-training methods on the Deep Blending dataset. Circle markers ({$\circ$}) indicate methods that utilize fine-tuning, while triangle markers ({\tiny$\triangle$}) denote methods without fine-tuning.
    }
\end{figure}
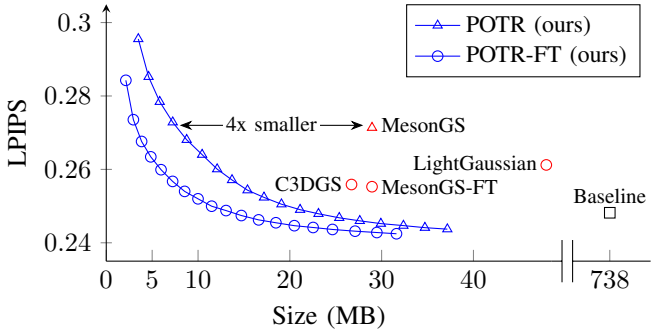

\Cref{tab:quantitative_results} and \Cref{fig:qualitative_results} respectively present the quantitative and qualitative performance of POTR(-FT). POTR achieves compression ratios of 20–45× while retaining the models' quality, surpassing all other post-training codecs. This highlights that no fine-tuning is required to achieve strong RD performance. Using fine-tuning, POTR-FT further improves POTR's already superior RD performance, reducing storage requirements by 35-50\% for the Deep Blending dataset, as shown in \Cref{fig:rd_curve}. Notably, the same figure shows that POTR(-FT) surpasses the baseline's quality at sufficiently high $q$, even without fine-tuning.

POTR(-FT)'s strong RD performance primarily results from our superior \changed{culling}{pruning} method as our compressed models typically use more bytes per splat than other post-training methods. We suspect that the latter results from our comparatively simple quantization, serialization, and entropy compression scheme. We believe future works could combine our ideas with existing literature to further improve RD performance.

\subsection{\changed{Culling}{Pruning}}
\label{section:culling_results}

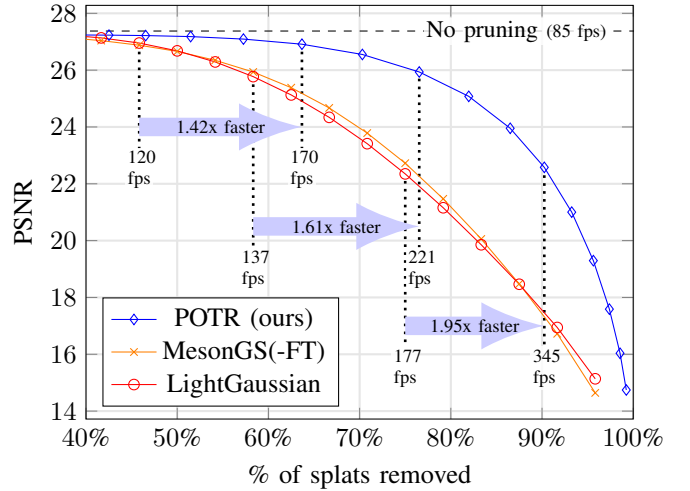
\begin{figure}
    \centering
\begin{tikzpicture}
    \begin{axis}[
        width=\linewidth,
        height=.8\linewidth,
        xlabel={\% of splats removed},
        ylabel={PSNR},
        legend pos=south west,
        grid=both,
        major grid style={line width=0.8pt, draw=gray!20},
        ymin=14.5, ymax=27.5,
        xmin=0.4, xmax=1.0,
        ytick={14,16,18,20,22,24,26,28},
        enlarge y limits=0.06,
        xticklabel={\pgfmathparse{\tick*100}\pgfmathprintnumber{\pgfmathresult}\%}
    ]

\addplot[dotted, line width=1pt, forget plot] coordinates {(0.75, 22.1) (0.75, 16.25)};
\addplot[dotted, line width=1pt, forget plot] coordinates {(0.90249, 22.3) (0.90249, 16.25)};
\draw[->, >=latex, blue!20!white, line width=7pt] (0.75,17) to node[black,sloped]{\scriptsize 1.95x faster} (0.90249,17);
\node[fill=white, text opacity=1, fill opacity=0.8, inner sep=1pt] at (axis cs:0.75,15.5) {\parbox[c]{.3cm}{\centering \scriptsize 177 \\ fps}};
\node[fill=white, text opacity=1, fill opacity=0.8, inner sep=1pt] at (axis cs:0.90249,15.5) {\parbox[c]{.3cm}{\centering \scriptsize 345 \\ fps}};

\addplot[dotted, line width=1pt, forget plot] coordinates {(0.58333, 25.5) (0.58333, 19.75)};
\addplot[dotted, line width=1pt, forget plot] coordinates {(0.76536, 25.7) (0.76536, 19.75)};
\draw[->, >=latex, blue!20!white, line width=7pt] (0.58333,20.5) to node[black,sloped]{\scriptsize 1.61x faster} (0.76536,20.5);
\node[fill=white, text opacity=1, fill opacity=0.8, inner sep=1pt] at (axis cs:0.58333,19) {\parbox[c]{.3cm}{\centering \scriptsize 137 \\ fps}};
\node[fill=white, text opacity=1, fill opacity=0.8, inner sep=1pt] at (axis cs:0.766,19) {\parbox[c]{.3cm}{\centering \scriptsize 221 \\ fps}};

\addplot[dotted, line width=1pt, forget plot] coordinates {(0.45833, 26.65) (0.45833, 23.25)};
\addplot[dotted, line width=1pt, forget plot] coordinates {(0.63675, 26.65) (0.63675, 23.25)};
\draw[->, >=latex, blue!20!white, line width=7pt] (0.45833,24) to node[black,sloped]{\scriptsize 1.42x faster} (0.63675,24);
\node[fill=white, text opacity=1, fill opacity=0.8, inner sep=1pt] at (axis cs:0.45833,22.5) {\parbox[c]{.3cm}{\centering \scriptsize 120 \\ fps}};
\node[fill=white, text opacity=1, fill opacity=0.8, inner sep=1pt] at (axis cs:0.63675,22.5) {\parbox[c]{.3cm}{\centering \scriptsize 170 \\ fps}};

     \addplot[color=blue, mark=diamond] table {
0 27.3764267
0.306756514 27.2375851
0.3082551128 27.2404766
0.3250593338 27.2404785
0.3346900245 27.2404079
0.3486183893 27.2402859
0.3677565442 27.239563
0.3930155769 27.2376862
0.4251968882 27.2319431
0.4654151722 27.2167034
0.5147328504 27.1808224
0.572595318 27.0934963
0.6367539912 26.9128647
0.7030109427 26.5516453
0.7653609799 25.9358501
0.8196632472 25.077713
0.8650942349 23.9535599
0.9024889353 22.5785885
0.9325548298 21.0020447
0.9562763592 19.2949963
0.9738350211 17.5886364
0.9855456517 16.0344086
0.9923848081 14.7427444
    };
    \addlegendentry{POTR (ours)}
    
   \addplot[color=orange, mark=x] table {
0 27.3764267
0.07588747073 27.3760166
0.08333333333 27.3758945
0.125 27.3744049
0.1666666667 27.3677883
0.2083333333 27.3534393
0.25 27.3285122
0.2916666667 27.2894802
0.3333333333 27.2335758
0.375 27.153162
0.4166666667 27.0394592
0.4583333333 26.8777866
0.5000001714 26.6591721
0.5416666667 26.3468914
0.5833333333 25.9340897
0.625 25.3827305
0.6666666667 24.6707172
0.7083333333 23.7850018
0.75 22.724968
0.7916666667 21.4706478
0.8333333333 20.0576649
0.875 18.4778652
0.9166666667 16.7220535
0.9583333333 14.6407442
    };
    \addlegendentry{MesonGS(-FT)}
    
    \addplot[color=red, mark=o] table {
0 27.3764267
0.07588747073 27.3760166
0.08333333333 27.3760204
0.125 27.3764191
0.1666666667 27.3768368
0.2083333333 27.3747196
0.25 27.3667507
0.2916666667 27.3477249
0.3333333333 27.3110371
0.375 27.2455959
0.4166666667 27.1352291
0.4583333333 26.9539242
0.5 26.6789646
0.5416666667 26.2887478
0.5833333333 25.7720623
0.625 25.1272755
0.6666666667 24.3381958
0.7083333333 23.4118195
0.75 22.3486633
0.7916666667 21.1563015
0.8333333333 19.8538456
0.875 18.4643097
0.9166666667 16.9426498
0.9583333333 15.1367006
    };
    \addlegendentry{LightGaussian}

\addplot[dashed] coordinates {(-0.1, 27.3764267) (1.1, 27.3764267)};
\node[fill=white, text opacity=1, fill opacity=0.8, inner sep=1pt] at (axis cs:0.77,27.3764267) [anchor=west] {\small No pruning \scriptsize (85 fps)};

    \end{axis}
\end{tikzpicture}
    \caption{
        \label{fig:cull_comparison}
        Quantitative comparison of post-training \changed{culling}{pruning} methods for the Garden model. Frame rate values represent rendering at 1080p on an RTX 2080 Ti. LightGaussian and MesonGS inference speeds are shown as one as they are quasi-equal.
    }
\end{figure}

Existing \changed{culling}{pruning} methods are typically just as capable as POTR at removing trivial and low-importance splats. It is primarily after these splats are removed that our \changed{culling}{pruning} method starts to meaningfully outperform other methods. As a result, our method keeps becoming comparatively better than other methods as more and more distortion is tolerated. To illustrate this, \Cref{fig:cull_comparison} shows the relation between the PSNR and the fraction of splats removed for various \changed{culling}{pruning} methods. Here, all methods remove the first 40\% of splats quasi-losslessly, but afterward, performances start to diverge significantly. For a PSNR of 27, POTR requires 29\% fewer splats than MesonGS and LightGaussian, but for a PSNR of 22, this has more than doubled to POTR requiring 64\% fewer splats.

\Cref{fig:cull_comparison} also shows that our \changed{culling}{pruning} method leads to markedly faster inference times. While this is primarily due to POTR simply removing more splats, there is a less obvious secondary phenomenon at play as models created by POTR can have far more splats, while still achieving the same inference speed as other methods. For example, \Cref{fig:cull_comparison} shows that LightGaussian and MesonGS reach 177 fps after removing 75\% of splats while POTR only has to remove a little over 65\% of splats to achieve the same inference speed. Upon closer inspection, we find that despite our model having more splats, it has fewer Gaussian instances during rendering for common camera poses. We suspect that this is due to our method removing more splats that are more frequently in view, but further research is required to verify this.

Beyond RD performance and inference speed, various other topics benefit from having a better removal order. Examples include level-of-detail rendering and progressive loading more broadly. Future works could look at how our proposed \changed{culling}{pruning} method can be adapted to these purposes.

\subsection{Spherical harmonics energy compaction}
\label{section:sh_results}

\begin{figure}
    \begin{tikzpicture}
\begin{groupplot}[
    group style={
        group size=1 by 3,
        vertical sep=.1cm,
        xlabels at=edge bottom,
        ylabels at=edge left,
        x descriptions at=edge bottom,
    },
    width=0.47\textwidth,
    height=0.26\textwidth,
    xlabel={$ \log_{10} \lambda $},
    xmin=-2,
    xmax=5,
    xtick={-4, -3, -2, -1, 0, 1, 2, 3, 4, 5, 6, 7},
    enlarge x limits=0.05,
    grid=major,
    legend style={
        at={(0.5,1.20)},
        anchor=north,
        column sep=5.9pt,
    },
    legend columns=3,
    cycle list name=exotic,
]

\nextgroupplot[
    ylabel={Sparsity AC coef.},
    ylabel style={align=center},
    ymin=0.7, ymax=1.,
    enlarge y limits=0.08,
]
\addplot[color=red, mark=o] table {
-5.00E+00 0.812450452
-4.50E+00 0.8179594752
-4.00E+00 0.8240272608
-3.50E+00 0.8309289748
-3.00E+00 0.839046055
-2.50E+00 0.8489372905
-2.00E+00 0.8610826565
-1.50E+00 0.8756848478
-1.00E+00 0.8925517673
-5.00E-01 0.9109943221
0.00E+00 0.9297872708
5.00E-01 0.9475340689
1.00E+00 0.9632210818
1.50E+00 0.9760348269
2.00E+00 0.9856035009
2.50E+00 0.9919499049
3.00E+00 0.9957139701
3.50E+00 0.9977236036
4.00E+00 0.9987555168
4.50E+00 0.9992906751
5.00E+00 0.9995851633
5.50E+00 0.9997590624
6.00E+00 0.9998603027
6.50E+00 0.9999240229
7.00E+00 0.9999639517
7.50E+00 0.9999860838
8.00E+00 0.9999960125
};
\addlegendentry{$\alpha = 1.0$}
\addplot[color=blue, mark=x] table {
-5.00E+00 0.8518676949
-4.50E+00 0.8568438614
-4.00E+00 0.8622582545
-3.50E+00 0.868245781
-3.00E+00 0.875042053
-2.50E+00 0.8829106755
-2.00E+00 0.8921823991
-1.50E+00 0.9029629603
-1.00E+00 0.9152393794
-5.00E-01 0.9285446609
0.00E+00 0.942251291
5.00E-01 0.9556280804
1.00E+00 0.9679765962
1.50E+00 0.978569021
2.00E+00 0.9868602946
2.50E+00 0.9925546174
3.00E+00 0.9959978339
3.50E+00 0.9978575203
4.00E+00 0.9988194778
4.50E+00 0.9993256797
5.00E+00 0.9996060377
5.50E+00 0.9997729518
6.00E+00 0.9998698032
6.50E+00 0.9999292682
7.00E+00 0.9999659588
7.50E+00 0.9999867528
8.00E+00 0.9999963336
};
\addlegendentry{$\alpha = 0.9$}
\addplot[color={rgb,255: red,119;green,172;blue,48}, mark=+] table {
-5.00E+00 0.8915650817
-4.50E+00 0.8937484001
-4.00E+00 0.8966332825
-3.50E+00 0.900362022
-3.00E+00 0.9050815148
-2.50E+00 0.9108811994
-2.00E+00 0.9179480345
-1.50E+00 0.9263228588
-1.00E+00 0.9357709971
-5.00E-01 0.9458757393
0.00E+00 0.956144987
5.00E-01 0.9660636803
1.00E+00 0.9752955019
1.50E+00 0.9833337415
2.00E+00 0.9897361849
2.50E+00 0.9941804581
3.00E+00 0.9968682924
3.50E+00 0.9983293331
4.00E+00 0.9990850097
4.50E+00 0.9994789453
5.00E+00 0.9997025947
5.50E+00 0.9998346648
6.00E+00 0.9999078587
6.50E+00 0.9999526849
7.00E+00 0.9999782961
7.50E+00 0.9999917841
8.00E+00 0.9999978323
};
\addlegendentry{$\alpha = 0.7$}

\addplot[dashed] coordinates {(-5, 1.0) (10, 1.0)};
\node[fill=white, text opacity=1, fill opacity=0.8, inner sep=1pt] at (axis cs:-0.9,1.0) [anchor=west] {\small DC only};

\addplot[dashed] coordinates {(-5, 0.701727121194) (10, 0.701727121194)};
\node[fill=white, text opacity=1, fill opacity=0.8, inner sep=1pt] at (axis cs:1.6,0.701727121194) [anchor=west] {\small No energy compaction};

\nextgroupplot[
    ylabel={Avg. L1-norm\\non-zero AC coef.},
    ylabel style={align=center},
    ymin=.0, ymax=6,
    enlarge y limits=0.08,
    scaled y ticks=false,
    yticklabel style={/pgf/number format/.cd, fixed, precision=2},
]
\addplot[color=red, mark=o] table {
-5.00E+00 3.981865701
-4.50E+00 3.720864966
-4.00E+00 3.50381213
-3.50E+00 3.328349547
-3.00E+00 3.183639942
-2.50E+00 3.067138629
-2.00E+00 2.968878567
-1.50E+00 2.881810513
-1.00E+00 2.802139351
-5.00E-01 2.724441481
0.00E+00 2.645461326
5.00E-01 2.563137351
1.00E+00 2.483299758
1.50E+00 2.412523546
2.00E+00 2.362011797
2.50E+00 2.33661402
3.00E+00 2.349216646
3.50E+00 2.391987716
4.00E+00 2.45521197
4.50E+00 2.516229988
5.00E+00 2.552492421
5.50E+00 2.56936386
6.00E+00 2.516713474
6.50E+00 2.387890113
7.00E+00 2.215071849
7.50E+00 2.04527449
8.00E+00 2.014948993
};
\addplot[color=blue, mark=x] table {
-5.00E+00 7.610471681
-4.50E+00 7.040644871
-4.00E+00 6.478867193
-3.50E+00 5.926817322
-3.00E+00 5.395096118
-2.50E+00 4.893163658
-2.00E+00 4.430516173
-1.50E+00 4.01506799
-1.00E+00 3.653383875
-5.00E-01 3.348630829
0.00E+00 3.097448886
5.00E-01 2.89277667
1.00E+00 2.728196284
1.50E+00 2.597845119
2.00E+00 2.509006912
2.50E+00 2.464262908
3.00E+00 2.471789752
3.50E+00 2.515570846
4.00E+00 2.578887732
4.50E+00 2.642429259
5.00E+00 2.68275361
5.50E+00 2.709781941
6.00E+00 2.654780531
6.50E+00 2.502504807
7.00E+00 2.293444847
7.50E+00 2.099959
8.00E+00 2.120826679
};
\addplot[color={rgb,255: red,119;green,172;blue,48}, mark=+] table {
-5.00E+00 11.71463227
-4.50E+00 10.99455956
-4.00E+00 10.18596263
-3.50E+00 9.308999239
-3.00E+00 8.397886537
-2.50E+00 7.488677226
-2.00E+00 6.623024633
-1.50E+00 5.837505713
-1.00E+00 5.152981166
-5.00E-01 4.575280888
0.00E+00 4.100287261
5.00E-01 3.706859088
1.00E+00 3.389177492
1.50E+00 3.141608533
2.00E+00 2.980715288
2.50E+00 2.90753023
3.00E+00 2.922123012
3.50E+00 3.005401645
4.00E+00 3.11314997
4.50E+00 3.203321475
5.00E+00 3.293530652
5.50E+00 3.365547939
6.00E+00 3.251063609
6.50E+00 3.007937138
7.00E+00 2.695440684
7.50E+00 2.377640961
8.00E+00 2.337702778
};

\addplot[dashed] coordinates {(-5, 3.37E+00) (10, 3.37E+00)};
\node[fill=white, text opacity=1, fill opacity=0.8, inner sep=1pt] at (axis cs:1.6,3.73E+00) [anchor=west] {\small No energy compaction};

\nextgroupplot[
    ylabel={PSNR},
    ymin=22.96, ymax=25,
    enlarge y limits=0.08,
]
\addplot[color=red, mark=o] table {
-5.00E+00 24.964573
-4.50E+00 24.963974
-4.00E+00 24.963608
-3.50E+00 24.963181
-3.00E+00 24.963878
-2.50E+00 24.962458
-2.00E+00 24.959977
-1.50E+00 24.957391
-1.00E+00 24.951775
-5.00E-01 24.94288
0.00E+00 24.929607
5.00E-01 24.902381
1.00E+00 24.859461
1.50E+00 24.787729
2.00E+00 24.679767
2.50E+00 24.534592
3.00E+00 24.369657
3.50E+00 24.215596
4.00E+00 24.080515
4.50E+00 23.966277
5.00E+00 23.867485
5.50E+00 23.779875
6.00E+00 23.697718
6.50E+00 23.620615
7.00E+00 23.55704
7.50E+00 23.506926
8.00E+00 23.471051
};
\addplot[color=blue, mark=x] table {
-5.00E+00 24.940783
-4.50E+00 24.940053
-4.00E+00 24.939195
-3.50E+00 24.93846
-3.00E+00 24.938851
-2.50E+00 24.937197
-2.00E+00 24.934029
-1.50E+00 24.929245
-1.00E+00 24.92195
-5.00E-01 24.911394
0.00E+00 24.895344
5.00E-01 24.870432
1.00E+00 24.831206
1.50E+00 24.764533
2.00E+00 24.660191
2.50E+00 24.519447
3.00E+00 24.358693
3.50E+00 24.207278
4.00E+00 24.072231
4.50E+00 23.958716
5.00E+00 23.861271
5.50E+00 23.772959
6.00E+00 23.689065
6.50E+00 23.611446
7.00E+00 23.547317
7.50E+00 23.499996
8.00E+00 23.466949
};
\addplot[color={rgb,255: red,119;green,172;blue,48}, mark=+] table {
-5.00E+00 24.845029
-4.50E+00 24.844635
-4.00E+00 24.844351
-3.50E+00 24.843856
-3.00E+00 24.842436
-2.50E+00 24.840515
-2.00E+00 24.837498
-1.50E+00 24.832724
-1.00E+00 24.824418
-5.00E-01 24.811018
0.00E+00 24.790519
5.00E-01 24.757693
1.00E+00 24.706845
1.50E+00 24.628473
2.00E+00 24.517837
2.50E+00 24.38328
3.00E+00 24.238268
3.50E+00 24.103692
4.00E+00 23.987624
4.50E+00 23.891653
5.00E+00 23.807413
5.50E+00 23.731215
6.00E+00 23.657816
6.50E+00 23.584764
7.00E+00 23.526808
7.50E+00 23.482022
8.00E+00 23.456207
};

\addplot[dashed] coordinates {(-5, 24.923665) (10, 24.923665)};
\node[fill=white, text opacity=1, fill opacity=0.8, inner sep=1pt] at (axis cs:1.6,24.923665) [anchor=west] {\small No energy compaction};

\addplot[dashed] coordinates {(-5, 23.428956) (10, 23.428956)};
\node[fill=white, text opacity=1, fill opacity=0.8, inner sep=1pt] at (axis cs:-0.9,23.428956) [anchor=west] {\small DC only ($\lambda = +\infty$)};

\addplot[dashed] coordinates {(-5, 22.962794) (10, 22.962794)};
\node[fill=white, text opacity=1, fill opacity=0.8, inner sep=1pt] at (axis cs:-0.9,22.962794) [anchor=west] {\small DC only (naive)};

\end{groupplot}
    \end{tikzpicture}
    \caption{
        \label{fig:reg_effect_on_sh}
        Effect of $\lambda$ and $\alpha$ on the magnitude and sparsity of the AC lighting coefficients and PSNR for the Truck model (post-\changed{culling}{pruning} and post-quantization).
        'Naive' refers to setting all AC lighting coefficients to zero without recomputing the DC lighting coefficients.
    }
\end{figure}
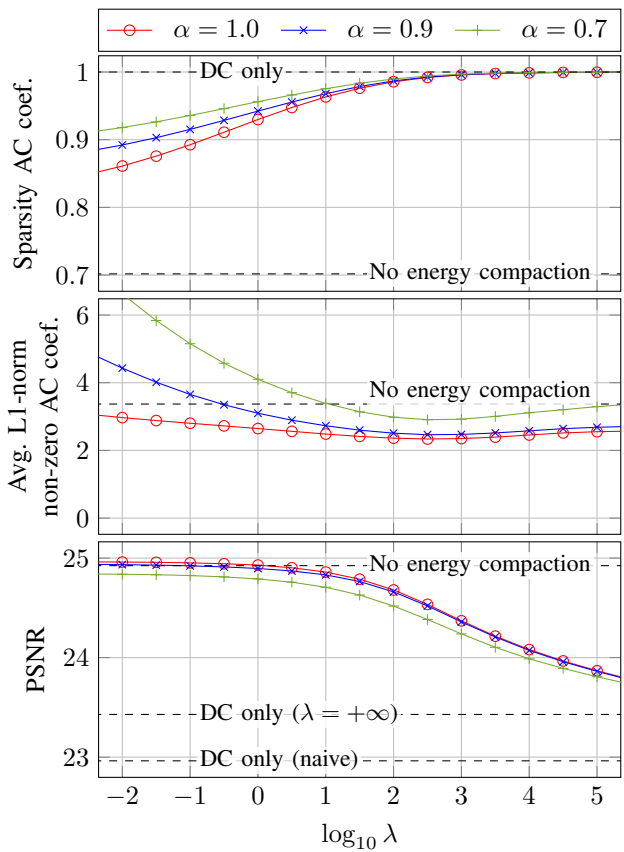

To better understand the effectiveness of spherical harmonics energy compaction at lowering the entropy of the AC lighting coefficients, we analyze how $\lambda$ and $\alpha$ influence the magnitude and sparsity of the AC lighting coefficients and quality of the model. \Cref{fig:reg_effect_on_sh} shows that the sparsity increases far more quickly than the PSNR decreases upon applying harsher regularization. This is despite the magnitude of non-zero AC lighting coefficients remaining relatively constant, indicating that by carefully choosing $\lambda$ and $\alpha$, the entropy of the AC lighting coefficients can be dramatically reduced while minimally impacting the PSNR.
For example in \Cref{fig:reg_effect_on_sh} for $\lambda=10^{1.1}$ and $\alpha=0.9$, the number of non-zero AC coefficients decreases 10-fold. Despite eliminating 90\% of non-zero AC coefficients, the PSNR is minorly impacted, going from 24.97 to 24.82. Further removing the remaining 10\% of AC coefficients would cause an 11-fold larger increase in the MSE, highlighting our method's effectiveness in identifying each lighting coefficient's importance to the quality of the model.


\begin{figure}
    \centering
    \includegraphics[width=0.84\linewidth]{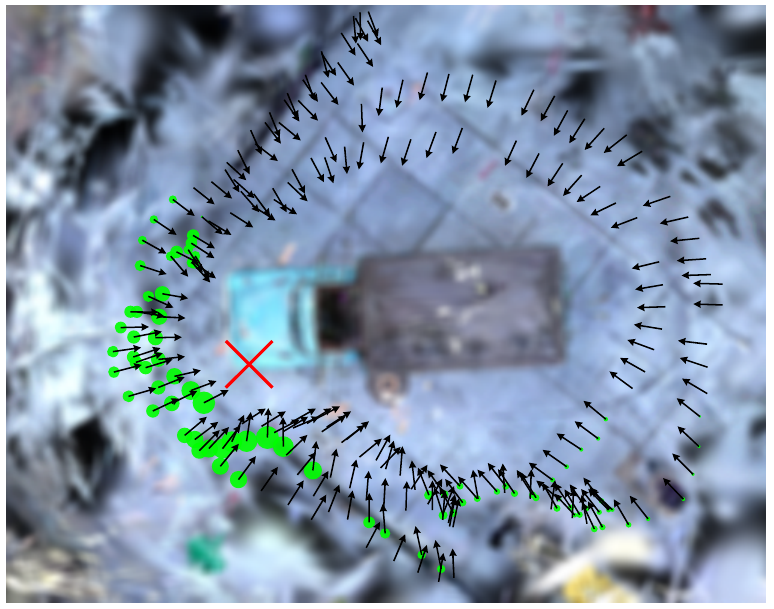} %
    \hspace*{-.2cm}
    \includegraphics[width=0.15\linewidth]{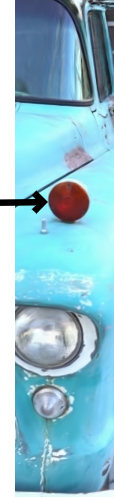} \\
    \begin{picture}(0,5)
        \put(-109, 4){
        \footnotesize \textbf{Before energy compaction}
        }

        \put(20, 4){
        \footnotesize \textbf{After energy compaction}
        }
    \end{picture}
    \includegraphics[trim={.3cm 0.8cm .3cm 1.7cm},clip, width=\linewidth]{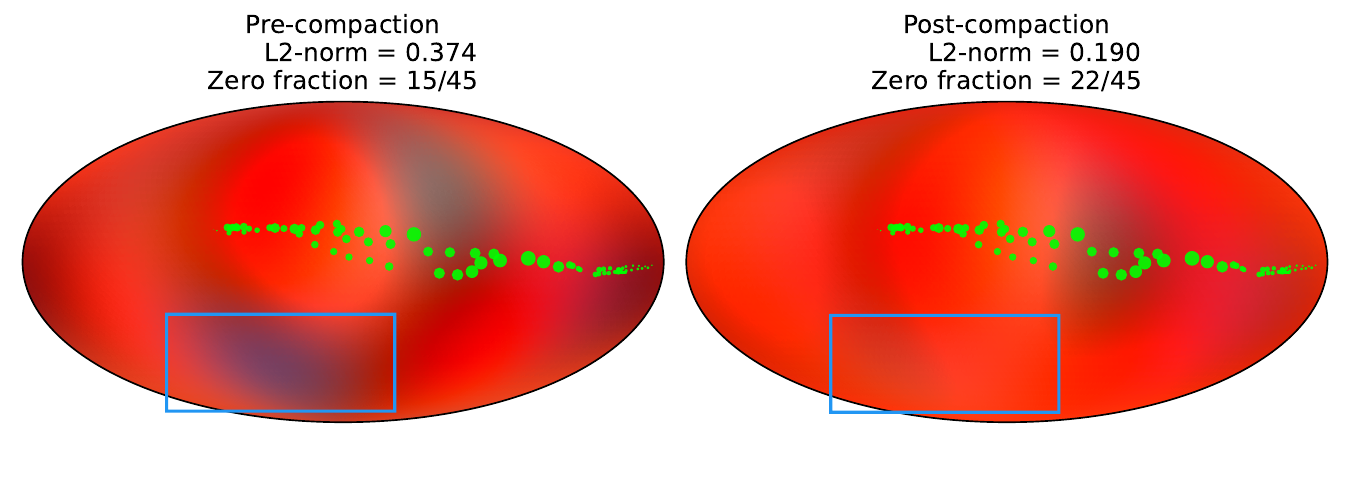}
    
    \caption{
    A blurred, top-down view of the Truck model shows a red cross marking a splat from the truck's red reflector. Each training camera $\vect{s}$ is represented by a black arrow, indicating its viewing direction, and a green dot whose size is proportional to $I_k(\vect{s})$. On the bottom, Mollweide projections depict the splat's view-dependent color before and after energy compaction, where every green dot again corresponds with a camera. Energy compaction results in a lower energy alias that removes a hallucinated blue color (highlighted by a blue rectangle) from an unsampled viewing area.
    }
    \label{fig:truck_scenario_1}
\end{figure}
To practically illustrate the impact of energy compaction, we examine a single splat in \Cref{fig:truck_scenario_1}. Here, training cameras with high importance preserve their directions' colors while other directions' colors change in favor of a low-energy alias. As low-energy aliases typically correspond to 'simpler' solutions, we indirectly follow Occam's razor which often leads to more sensible generalizations for unseen viewpoints. Originally the splat in the red reflector became blue when looked at from below, but with the simpler alias, the red reflector stays red. More broadly, we suspect low-energy aliases to be the driving force behind the PSNR increasing slightly, from 24.92 to 24.97 under minimal regularization in \Cref{fig:reg_effect_on_sh}. Notably, this increase in PSNR is achieved without using any ground-truth images.

A key advantage of our energy compaction approach is its adaptability to model- and scene-specific characteristics.
For example, the geometry of the scene and the camera trajectory are clearly reflected in the sparsity of the spherical harmonics bases. 
For the Truck model, splats are primarily observed from the side. Low-energy solutions prefer spherical harmonics bases that change significantly across samples as this leads to smaller coefficients.
As a result, the energy compacted Truck model has three times fewer non-zero $L_2$ than non-zero $L_4$ values as $Y_2$ remains relatively constant near the 'equator', while $Y_4$ changes substantially.
Such observations imply further predictability of the lighting coefficients and are therefore useful for compression. POTR exploits these observations through its serialization order, but future works could exploit these observations more explicitly through, for example, a custom entropy model.

\vspace{-3pt}
\subsection{Ablation Study}
\label{section:ablation}

\begin{table}
\addtolength{\tabcolsep}{-0.45em} 
\begin{tabular}{l|ccc|cc}
& \footnotesize PSNR$\uparrow$ & \footnotesize SSIM$\uparrow$ & \footnotesize LPIPS$\downarrow$ & \footnotesize \makecell{Size \\ MB}$\downarrow$ & \footnotesize \makecell{\# Splats \\ x1,000}$\downarrow$ \\
\hline
3DGS baseline & 25.46 & .775 & .210 & 1,521 & 6,132 \\
\hline
Ours & & & & & \\
+ \added{Serialization \& Quantization} & \added{25.10} & \added{.760} & \added{.213} & \added{715.5} & \added{6,132} \\
+ \added{Entropy compression} & \added{25.10} & \added{.760} & \added{.213} & \added{189.3} & \added{6,132} \\
+ \added{Pruning} & \added{25.05} & \added{.755} & \added{.220} & \added{91.44} & \added{2,656} \\
+ \added{SH compaction} & \added{25.03} & \added{.754} & \added{.221} & \added{57.40} & \added{2,626} \\
+ \added{RGB $\rightarrow$ YCoCg} & 25.01	& .753 & .224 & 41.53 & 2,624 \\
+ Fine-tuning & 25.19 & .758 & .219 & 33.86 & 2,221
\end{tabular}
\caption{
    \label{table:ablation_study}
    Ablation study for the different components of the POTR(-FT) encoder for the Bicycle model.
}
\end{table}

\begin{table}
\begin{center}
\addtolength{\tabcolsep}{-0.2em}
\begin{tabular}{ c | c c | c }
    \multirow{2}{*}{Splat property} & \multicolumn{2}{c|}{Bytes per splat (\%)} & \multirow{2}{*}{\makecell{Compression\\factor}}\\
    & Uncompressed & Compressed & \\
    \hline 
    Position     & 12 (5.1\%) & 3.58 (23 \%) & 3.35x \\
    Scale        & 12 (5.1\%) & 2.40 (16 \%) & 6.45x \\
    Opacity      & \hspace{.045cm} 4 (1.7\%) & 0.72 (4.7 \%) & 5.59x \\
    Rotation     & 16 (6.8\%) & 3.29 (22 \%) & 4.87x \\
    DC SH coefs. & 12 (5.1\%) & 2.24 (15 \%) & 5.37x \\
    AC SH coefs. & 180 (76\%) & 3.03 (20 \%) & 59.4x \\
    \hline
    Total & 236 (100\%) & 15.3 (100\%) & 15.5x \\
\end{tabular}

\caption{
    \label{table:size_contribution}
    Effect of the post-\changed{culling}{pruning} steps (no fine-tuning) on the size of the splat properties of the Bicycle model. A property's compressed size is approximated through its ablation from the bitstream.
}
\end{center}
\end{table}

\Cref{table:ablation_study} presents an ablation study that evaluates the incremental impact of the different components of our proposed codec. It reveals that quantization, combined with our serialization strategy and entropy compression, already compresses the Bicycle model \added{$8 \times$}. Incorporating \changed{culling}{pruning} further reduces the size by half, as more than half of the splats are removed. \added{Spherical harmonics energy compaction provides an additional 37\% size reduction while only modifying lighting coefficients and with virtually no quality loss. Similarly, switching from an RGB to a YCoCg color representation (including more harshly regularizing the chrominance channels) yields a further $28\%$ reduction, again highlighting the importance of lighting coeficients to 3DGS compression.}

\added{
Finally, fine-tuning consistently improves all metrics, reducing both distortion and the number of splats. This improvement arises because encoding and fine-tuning act in complementary ways: encoding is coarse, lossy, and  approximate (e.g., pruning, quantization, and SH compaction), while fine-tuning corrects the resulting errors through small, precise, global adjustments. Iteratively alternating between encoding and fine-tuning can hence yield further gains, as each process creates additional rate-distortion improvement opportunities for the other. For instance, SH energy compaction recomputes all SH coefficients in a coarse, non-gradual manner based on approximate statistics. Fine-tuning then refines these coefficients holistically, applying small, accurate, and globally consistent corrections. In effect, fine-tuning validates the updates introduced by energy compaction, allowing the subsequent compaction step to be even more aggressive relative to the original SH coefficients.}

To quantify the compression of lighting coefficients, we conduct a second ablation study, shown in \Cref{table:size_contribution}, to evaluate the contribution of individual splat properties to the final compressed model size. We estimate a property's contribution to the file size by measuring the reduction in the compressed model size when that property is removed from the bitstream\footnote{\label{footnote:disection}Zstd's compressed bitstream can, strictly speaking, not be decomposed into distinct parts due to the holistic nature of lossless compression. However, this analysis only aims to provide an approximate understanding of each property's contribution to the compressed file size.}. \Cref{table:size_contribution} reveals that the AC lighting coefficients are compressed by a factor of nearly 60×, whereas other properties achieve more modest compression factors of 3-6×. Consequently, AC lighting coefficients are no longer the primary contributors to the model size, with positional and rotational information each occupying a larger share.

\vspace{-3pt}
\subsection{Codec speed}
\label{section:speed}
\begin{figure}
    \centering
    \begin{tikzpicture}
\begin{axis}[
    ybar stacked,
    width=\columnwidth,
    height=6.2cm,
    bar width=7pt,
    enlarge x limits=0.05,
    ylabel={POTR Encoding Time (s)},
    symbolic x coords={ficus,mic,materials,hotdog,drums,lego,ship,chair,truck,train,playroom,stump,treehill,flowers,drjohnson,counter,bonsai,garden,room,bicycle,kitchen},
    xtick=data,
    xticklabel style={rotate=90, anchor=east},
    legend style={
        at={(0.02,0.98)},
        anchor=north west,
        font=\small,
    },
    legend cell align={left},
    ymin=0,
    ymajorgrids=true,
    grid style=solid,
]

\definecolor{prune}{RGB}{31,119,180}
\definecolor{sh}{RGB}{214,39,40}
\definecolor{misc}{RGB}{44,160,44}
\definecolor{zstd}{RGB}{255,127,14}

\addplot+[ybar, fill=prune] coordinates {
(ficus,19.654) (mic,20.49) (materials,20.308) (hotdog,22.846) (drums,21.893) (lego,22.208) (ship,24.217) (chair,24.106) (truck,51.308) (train,59.503) (playroom,67.94) (stump,44.696) (treehill,49.286) (flowers,60.173) (drjohnson,90.09899999999999) (counter,110.917) (bonsai,125.68599999999999) (garden,76.101) (room,136.656) (bicycle,77.378) (kitchen,138.102) };
\addlegendentry{Pruning}

\addplot+[ybar, fill=zstd] coordinates {
(ficus,1.85172) (mic,2.04601) (materials,3.01602) (hotdog,1.55732) (drums,3.26058) (lego,3.36805) (ship,3.5432) (chair,3.58639) (truck,11.20758) (train,10.97695) (playroom,13.99351) (stump,44.9302) (treehill,42.06229) (flowers,38.81351) (drjohnson,23.39464) (counter,12.25361) (bonsai,8.66276) (garden,48.56362) (room,6.38611) (bicycle,53.83516) (kitchen,22.40183) };
\addlegendentry{Entropy compression (zstd)}

\addplot+[ybar, fill=sh] coordinates {
(ficus,0.248) (mic,0.294) (materials,0.248) (hotdog,0.273) (drums,0.381) (lego,0.401) (ship,0.446) (chair,0.522) (truck,3.211) (train,2.68) (playroom,2.769) (stump,3.984) (treehill,4.09) (flowers,4.421) (drjohnson,4.392) (counter,2.24) (bonsai,2.755) (garden,6.329) (room,2.856) (bicycle,7.091) (kitchen,3.854) };
\addlegendentry{SH energy compaction}

\addplot+[ybar, fill=misc] coordinates {
(ficus,0.2122799999999998) (mic,0.2439900000000037) (materials,0.30697999999999936) (hotdog,0.2466799999999978) (drums,0.42841999999999913) (lego,0.44594999999999985) (ship,0.4817999999999998) (chair,0.51661) (truck,3.3484200000000044) (train,2.2950500000000034) (playroom,2.7704899999999952) (stump,8.060800000000015) (treehill,7.774709999999999) (flowers,6.706490000000002) (drjohnson,4.457360000000023) (counter,1.7493900000000053) (bonsai,1.7432400000000143) (garden,9.140379999999993) (room,1.7468900000000076) (bicycle,10.013840000000016) (kitchen,3.2001699999999857) };
\addlegendentry{Miscellaneous}

\end{axis}
\end{tikzpicture}
    \caption{
        \label{fig:encoding_times}
        \added{
        POTR's encoding time for various models. 'Miscellaneous' includes the remaining encoding operations, such as quantization and serialization.
        }
    }
\end{figure}
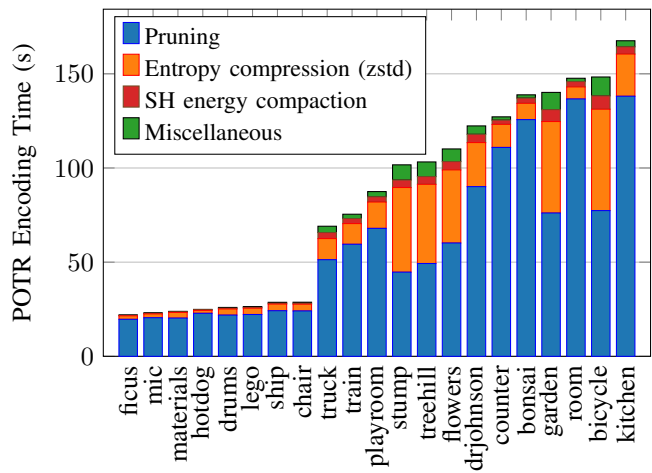

POTR and POTR-FT make trade-offs between their speed and RD performance.
For the results presented in \Cref{tab:quantitative_results} and \Cref{fig:qualitative_results}, RD performance is prioritized,
\added{
however, POTR can still be fast. In this subsection we discuss the speed of our codec using an NVIDIA GeForce RTX 5090 and AMD Ryzen 9 9950X.
}

\added{
Decoding is extremely fast, with all but the largest models loading in one second or less. A linear model that fits the total decoding time in seconds $y$ as a function of the number of splats post-pruning $x$ with high accuracy is
\[
y = 5.51 \times 10^{-7} x - 0.02 \quad (r > 0.999) \, .
\]
This corresponds to a decoding throughput of approximately 1.8 million splats per second, which is sufficient for most practical use cases. The total decoding time is dominated by three operations: zstd decompression, octree deserialization, and attribute deserialization, which on average correspond to $10\%$, $62\%$, and $26\%$ of total decode time respectively.
}

\added{
As shown in \Cref{fig:encoding_times}, encoding is considerably slower, with POTR requiring between 20 and 170, seconds depending on the model. Because POTR-FT repeatedly performs encoding and its fine-tuning stage is comparatively fast, the total encoding time of POTR-FT is closely tied to that of POTR. Each of the $R_{\text{FT}}$ fine-tuning cycles takes roughly $80\%$ of the original encoding time, although this can be reduced substantially by integrating fine-tuning directly into the training loop. \Cref{fig:encoding_times} also breaks down the contribution of each encoding step to the overall encoding time, we discuss the three most important steps in order of their importance:
}
\subsubsection{\changed{Culling}{Pruning}}
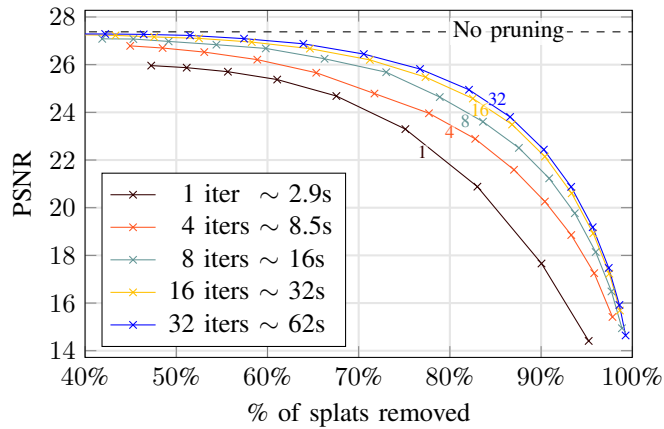
\begin{figure}
    \centering
\begin{tikzpicture}
    \begin{axis}[
        width=\linewidth,
        height=.7\linewidth,
        xlabel={\% of splats removed},
        ylabel={PSNR},
        legend pos=south west,
        grid=both,
        major grid style={line width=0.8pt, draw=gray!20},
        ytick={14,16,18,20,22,24,26,28},
        xmin=0.4, xmax=1.0,
        ymin=14.5, ymax=27.5,
        enlarge x limits=0.0,
        enlarge y limits=0.06,
        xticklabel={\pgfmathparse{\tick*100}\pgfmathprintnumber{\pgfmathresult}\%}
    ]

    \addplot[color={rgb,221:red,44;green,0;blue,0}, mark=x] table {
0.4725227189	25.9607
0.5114389839	25.8739
0.5566123785	25.7046
0.6106543104	25.3751
0.6755290342	24.6848
0.7511979878	23.2966
0.8301649213	20.8818
0.9004444723	17.6594
0.9521735509	14.4112
    };
    \addlegendentry{\hspace{0.15cm} 1 iter \hspace{0.012cm} $\sim$ 2.9s}
    \addplot[color={rgb,255:red,255;green,87;blue,34}, mark=x] table {
0.4494810776	26.7878
0.4849862480	26.6984
0.5306984800	26.5276
0.5888499043	26.2102
0.6534802317	25.6608
0.7172927053	24.788
0.7770510442	23.961
0.8277828280	22.8883
0.8702735183	21.5904
0.9041578917	20.2587
0.9330671024	18.8538
0.9582231322	17.2536
0.9782581497	15.417
    };
    \addlegendentry{\hspace{0.15cm} 4 iters $\sim$ 8.5s}

    \addplot[color={rgb,255:red,100;green,150;blue,150}, mark=x] table {
0.4188016214	27.0837
0.4529295343	27.075
0.4916267680	26.9748
0.5440203099	26.8378
0.5981467352	26.6809
0.6626063621	26.247
0.7301862417	25.6834
0.7889022113	24.6404
0.8362518647	23.6076
0.8756584648	22.5086
0.9088663779	21.2308
0.9369558496	19.7687
0.9598540409	18.1402
0.9771252886	16.4877
0.9886305988	14.9432
    };
    \addlegendentry{\hspace{0.05cm} 8 iters $\sim$ 16s}
    \addplot[color={rgb,255:red,255;green,192;blue,0}, mark=x] table {
0.3995422967	27.2537
0.4333730263	27.2267
0.4758225840	27.1624
0.5248667303	27.102
0.5827715302	26.947
0.6465459561	26.6721
0.7122424755	26.2001
0.7732488469	25.4785
0.8251667928	24.5794
0.8681918645	23.4897
0.9041196726	22.1442
0.9335378996	20.6052
0.9571728791	18.9222
0.9748326588	17.233
0.9864301746	15.6869
};
    \addlegendentry{16 iters $\sim$ 32s}

     \addplot[color=blue, mark=x] table {
0.3887048775	27.2868
0.4220550409	27.2874
0.4642567745	27.2695
0.5150007267	27.226
0.5743194607	27.0914
0.6393462037	26.8756
0.7056204651	26.4446
0.7672604504	25.8172
0.8209518981	24.948
0.8660010379	23.8071
0.9029837266	22.4375
0.9329560443	20.8756
0.9565598315	19.1819
0.9741447156	17.4733
0.9858851673	15.9189
0.9926295472	14.6411
    };
    \addlegendentry{32 iters $\sim$ 62s}

\node[fill=white, text={rgb,221:red,44;green,0;blue,0}, text opacity=1, fill opacity=0.5, inner sep=0] at (axis cs:0.77,22.35) {\scriptsize 1};
\node[fill=white, text={rgb,255:red,255;green,87;blue,34}, text opacity=1, fill opacity=0.5, inner sep=0] at (axis cs:0.8,23.2) {\scriptsize 4};
\node[fill=white, text={rgb,255:red,100;green,150;blue,150}, text opacity=1, fill opacity=0.5, inner sep=0] at (axis cs:0.817,23.65) {\scriptsize 8};
\node[fill=white, text={rgb,255:red,255;green,192;blue,0}, text opacity=1, fill opacity=0.5, inner sep=0] at (axis cs:0.833,24.1) {\scriptsize 16};
\node[fill=white, text=blue, text opacity=1, fill opacity=0.5, inner sep=0] at (axis cs:0.852,24.57) {\scriptsize 32};

\addplot[dashed] coordinates {(-0.1, 27.3764267) (1.1, 27.3764267)};
\node[fill=white, text opacity=1, fill opacity=0.8, inner sep=1pt] at (axis cs:0.8,27.3764267) [anchor=west] {\small No \added{pruning}};
    \end{axis}
\end{tikzpicture}
    \caption{
        \label{fig:cull_iterations}
        The impact of limiting the number of \changed{culling}{pruning} iterations on the Pareto front of POTR's \changed{culling}{pruning} method for the Garden model. Beyond 32 iterations ($\sim$ \added{62} seconds), the Pareto front \added{still improves, but minorly}.
    }
\end{figure}
\changed{Culling}{Pruning} is typically the most time-consuming step during encoding as every \changed{culling}{pruning} iteration starts by recalculating all $\triangle{\text{MSE}_j}$ values by executing a modified forward pass per training view. \changed{Culling}{Pruning} time is therefore roughly linear in
\begin{itemize}
    \item the number of \added{training} views,
    \item the time per modified forward pass, and 
    \item the number of \changed{culling}{pruning} iterations.
\end{itemize}
We expect that ideas from sparse scene reconstruction~\cite{liu2024georgs, li2024dngaussian} could be used to reduce the number of views rendered by an order of magnitude. Additionally, we expect that the time per modified forward pass can be further improved by carefully optimizing our modified forward pass.
We leave improving and studying these ideas to future works and focus on the effect of reducing the number of \changed{culling}{pruning} iterations. \Cref{fig:cull_iterations} shows that the number of \changed{culling}{pruning} iterations can be significantly reduced if a slightly lower RD performance is tolerated. Even using just one \changed{culling}{pruning} iteration, our \changed{culling}{pruning} performance is already on par with LightGaussian and MesonGS (also see \Cref{fig:cull_comparison}). Using only four \changed{culling}{pruning} iterations, POTR outperforms all other \changed{culling}{pruning} strategies. \added{We use 48 pruning iterations (see \Cref{tab:hyperparameters}), which effectively guarantees that the resulting RD performance is near-optimal.}

\subsubsection{Entropy compression \deleted{speed}}
\added{
At the highest compression level, entropy compression (zstd) can be a major contributor to total runtime, especially for the larger models, since zstd encode time scales roughly linearly with the final splat count. Lowering the compression level can reduce this overhead to the point of being negligible. For instance, using zstd at level 4 instead of level 22 yields approximately a $100\times$ speedup, at the cost of only about a $10\%$ increase in file size.
}

\subsubsection{Spherical harmonics energy compaction \deleted{speed}}
Our proposed spherical harmonics energy compaction method is embarrassingly parallel as each splat is processed independently. Furthermore, by using ridge regression, a closed-form solution is available that allows \added{one of the 32 cores of our test machine to energy compact upto 14,000 splats per second. Using a sufficient number of threads thus makes the time spherical harmonics energy compaction takes inconsequential.}

\section{Conclusion}
\label{section:conclusion}
This work introduced POTR, a post-training codec for 3D Gaussian Splatting that focuses on achieving strong rate-distortion performance without fine-tuning.
We first identified that current post-training \changed{culling}{pruning} methods rely heavily on heuristics which often leads to suboptimal splat removal decisions. To address this, we proposed an efficient method to precisely compute the impact of each splat's removal on an objective quality metric. POTR leverages this knowledge to remove splats in a manner that far outperforms existing \changed{culling}{pruning} methods, especially at higher distortion levels.
Additionally, we proposed the first method to non-trivially reduce the entropy of 3DGS lighting coefficients without training. Our approach is fast, embarrassingly parallel, highly adaptable, and shown to generalize lighting information more sensibly.
Combined with a simple quantization, serialization, and entropy compression scheme, these innovations allow POTR to significantly outperform existing methods in both rate-distortion performance and inference speed, despite not using fine-tuning.
Moreover, we demonstrated that incorporating a simple fine-tuning scheme further enhances POTR’s rate-distortion performance and inference speed. 
We believe that our codec's smaller file sizes and faster inference speeds could help make 3D Gaussian Splatting models more accessible, especially for on-demand and virtual reality applications.

\ifCLASSOPTIONcaptionsoff
  \newpage
\fi

\bibliographystyle{IEEEtran}
\bibliography{IEEEabrv, bibtex/bib/IEEEself}

\begin{IEEEbiography}[{\includegraphics[width=1in,height=1.25in,clip,keepaspectratio]{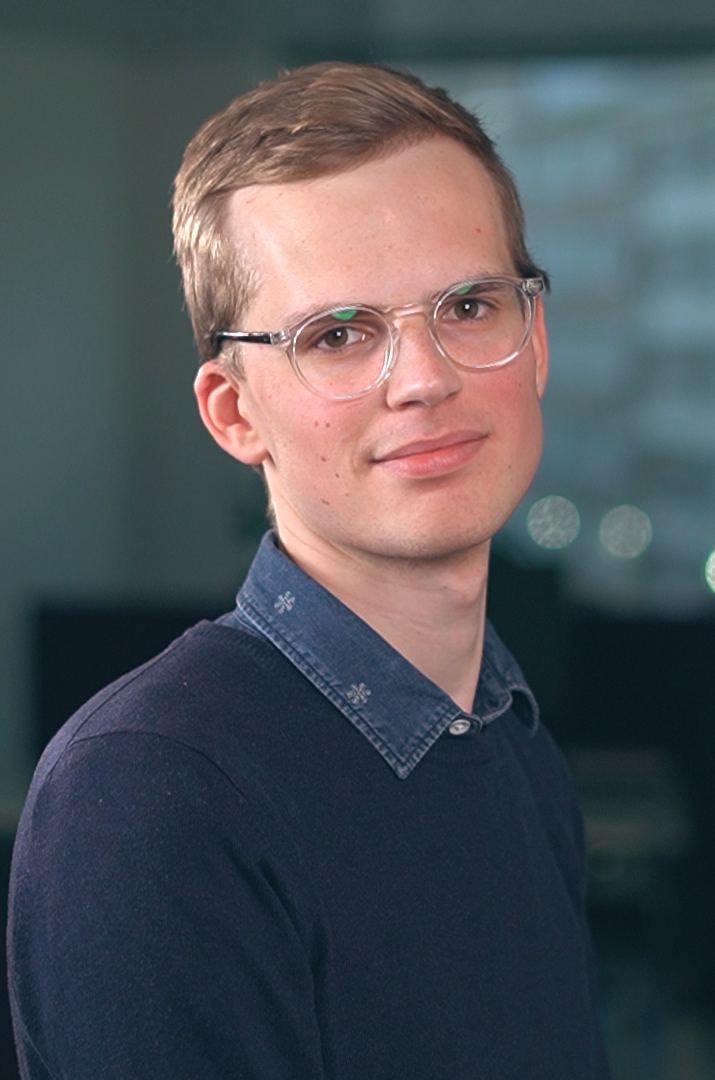}}]{Bert Ramlot}
 received the M.Sc. degree in computer science engineering from Ghent University, Belgium, in 2024, where he is currently pursuing the Ph.D. degree with IDLab, imec. His current research interests are 3D Gaussian Splatting and compression thereof.
\end{IEEEbiography}
\vspace{-3em}
\begin{IEEEbiography}[{\includegraphics[width=1in,height=1.25in,clip,keepaspectratio]{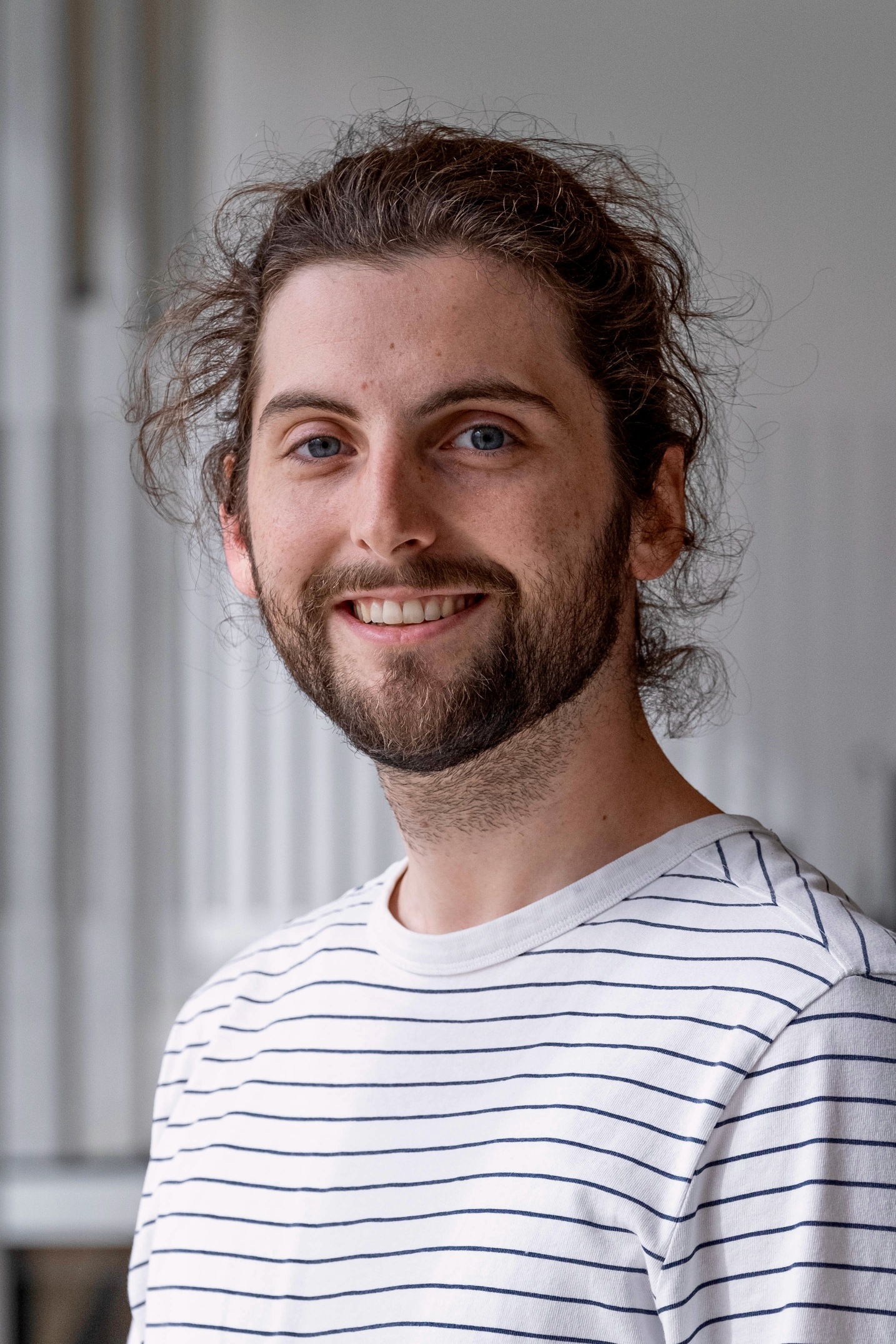}}]{Martijn Courteaux}
  received the M.Sc. degree in computer science engineering from Ghent University, Belgium, in 2018, where he is currently pursuing the Ph.D. degree with IDLab, imec, through the financial support of the Research Foundation–Flanders (FWO).
  His current research interests focus on the modeling and compression of light fields and light-field videos, and are set in the context of statistics, signal processing, and compression.
\end{IEEEbiography}
\vspace{-3em}
\begin{IEEEbiography}[{\includegraphics[width=1in,height=1.25in,clip,keepaspectratio]{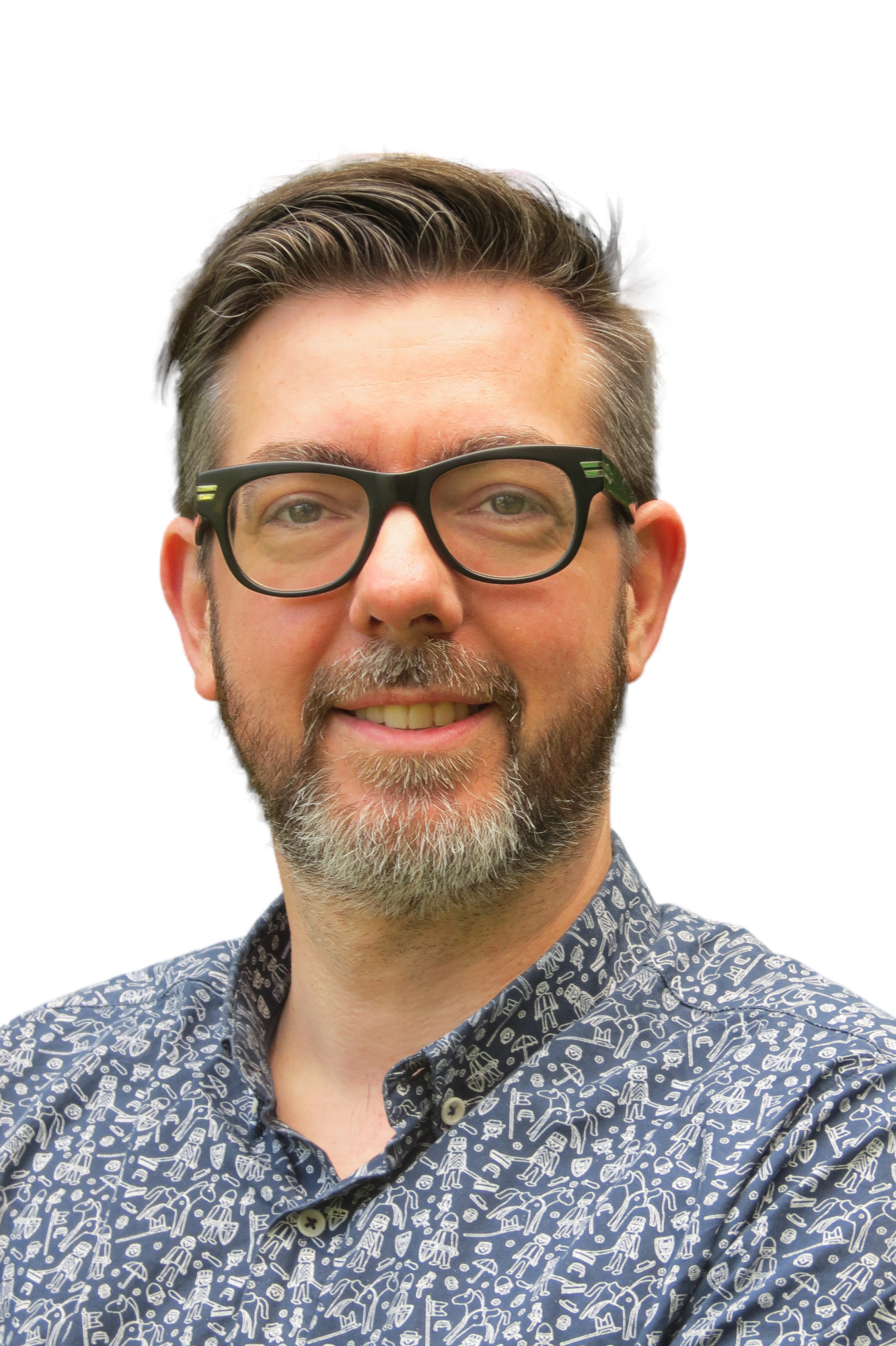}}]{Peter Lambert}
  (Senior Member, IEEE) is Full Professor at the IDLab of Ghent University – imec (Belgium). He received his Master's degree in science (mathematics) and in applied informatics from Ghent University in 2001 and 2002, respectively, and he obtained the Ph.D. degree in computer science in 2007 at the same university. His research interests include multimedia signal processing, data compression, computer graphics, XR, and visual communications.
\end{IEEEbiography}
\vspace{-3em}
\begin{IEEEbiography}[{\includegraphics[width=1in,height=1.25in,clip,keepaspectratio]{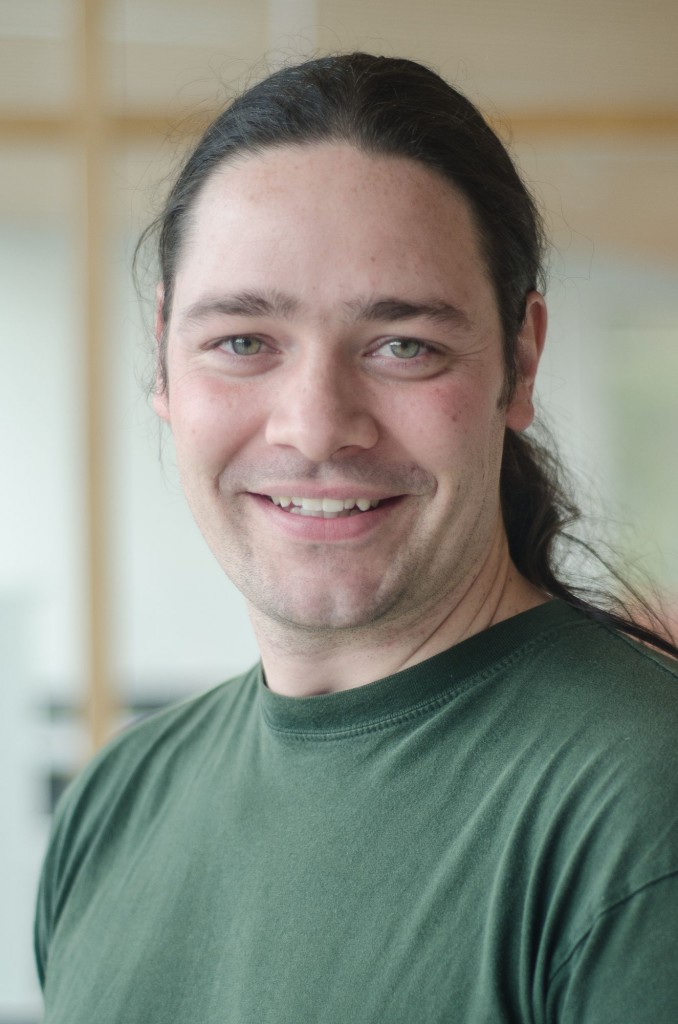}}]{Glenn Van Wallendael} (Member, IEEE)
  obtained the M.Sc. degree in Computer Science Engineering from Ghent University, Belgium in 2008. Afterwards, he obtained the Ph.D. at IDLab, Ghent University, with the financial support of the Research Foundation - Flanders (FWO). Since 2019, he works as a Professor for both Ghent University and imec on topics such as the efficient representation and compression of visual information, including immersive media, 360 degree video, light fields, \added{and} virtual reality\deleted{ and the different operations on these modalities such as (scalable) compression, computer vision, watermarking, personalized delivery, and quality estimation}.
\end{IEEEbiography}


\end{document}